%% file: _main.tex
\documentclass[pmlr,twocolumn,10pt]{jmlr} 

\usepackage{booktabs}
\usepackage{hyperref}
\usepackage{url}
\usepackage{amsmath}
\usepackage{bbm}
\usepackage{mismath}
\usepackage{multirow}
\usepackage{multicol}
\usepackage{makecell}
\usepackage{colortbl}
\usepackage{arydshln}
\usepackage{tcolorbox}
\usepackage{enumitem}
\usepackage{adjustbox}
\usepackage{graphicx}
\usepackage{subcaption}

\usepackage{include/titletoc}
\usepackage{appendix}

\usepackage[capitalize,noabbrev]{cleveref}

\definecolor{lightred}{rgb}{1,0.8,0.8}
\definecolor{lightgreen}{rgb}{0.8,1,0.8}
\definecolor{linen}{rgb}{0.980,0.941,0.902}
\definecolor{nyu}{RGB}{87, 6, 140}

\usepackage{siunitx}

\usepackage[switch]{lineno}

\theorembodyfont{\upshape}
\theoremheaderfont{\scshape}
\theorempostheader{:}
\theoremsep{\newline}

\jmlrvolume{333}
\jmlryear{2026}
\jmlrsubmitted{LEAVE UNSET}
\jmlrpublished{LEAVE UNSET}
\jmlrworkshop{Conference on Health, Inference, and Learning (CHIL) 2026} 

\title[]{An Empirical Analysis of Calibration and Selective Prediction \\ in Multimodal Clinical Condition Classification}

\author{%
 \Name{L. Julián Lechuga López} \Email{leopoldo.lechuga@nyu.edu}\\
 \Name{Farah E. Shamout} \Email{farah.shamout@nyu.edu}\\
 \addr New York University, New York University Abu Dhabi \\
 \Name{Tim G. J. Rudner} \Email{tim.rudner@utoronto.ca}\\
 \addr University of Toronto
}

\begin{document}

\maketitle

\input{0_abstract}

\paragraph*{Data and Code Availability}
This study used the publicly available MIMIC-CXR and MIMIC-IV datasets, which can be accessed via PhysioNet subject to completion of the required training and approval of the corresponding data use agreement.
Code developed for our study is available \href{https://github.com/jlaitue/medcertain}{here}.

\paragraph*{Institutional Review Board (IRB)}
This study does not involve human subjects, so IRB approval was not required.

\input{1_introduction}

\input{2_related_work}

\input{3_background}

\input{4_empirical_setup}

\input{5_results}

\input{discussion_conclusion}

\clearpage
\section*{Acknowledgments} 

This work was supported by the NYUAD Center for Artificial Intelligence and Robotics, funded by Tamkeen under the NYUAD Research Institute Award CG010. 
The research was carried out on the High Performance Computing resources at New York University Abu Dhabi.

\bibliography{references}

\clearpage
\onecolumn

\input{appendix}

\end{document}

%% file: 0_abstract.tex
\begin{abstract}
As artificial intelligence systems move toward clinical deployment, ensuring reliable prediction behavior is fundamental for safety-critical decision-making tasks.
One proposed safeguard is selective prediction, where models can defer uncertain predictions to human experts for review.
In this work, we empirically evaluate the reliability of uncertainty-based selective prediction in multilabel clinical condition classification using multimodal ICU data.
Across a range of state-of-the-art unimodal and multimodal models, we find that selective prediction can substantially \textit{degrade} performance despite strong standard evaluation metrics.
This failure is driven by severe class-dependent miscalibration, whereby models assign high uncertainty to correct predictions and low uncertainty to incorrect ones, particularly for underrepresented clinical conditions.
Our results show that commonly used aggregate metrics can obscure these effects, limiting their ability to assess selective prediction behavior in this setting.
Taken together, our findings characterize a task-specific failure mode of selective prediction in multimodal clinical condition classification and highlight the need for calibration-aware evaluation to provide strong guarantees of safety and robustness in clinical AI.
\end{abstract}

%% file: 1_introduction.tex
\section{Introduction}

Machine learning is increasingly embedded into the healthcare sector to support clinical decision-making, improve diagnostic performance, accelerate drug discovery, and optimize patient management \citep{paul2025harnessing, hanna2025future}. 
From enhancing disease detection \citep{mall2022implementation} to personalizing treatment plans \citep{agarwal2024machine}, machine learning-driven systems have demonstrated transformative potential. 
However, in high-stakes clinical environments exhibiting good performance, such as high accuracy, is not sufficient.
Models in healthcare must provide fail-safe mechanisms to ensure they are trustworthy and interpretable, allowing clinicians to recognize when predictions are unreliable so they can intervene accordingly \citep{javed2024robustness}.

\begin{figure*}[t!]
    \centering
     \includegraphics[width=\linewidth]{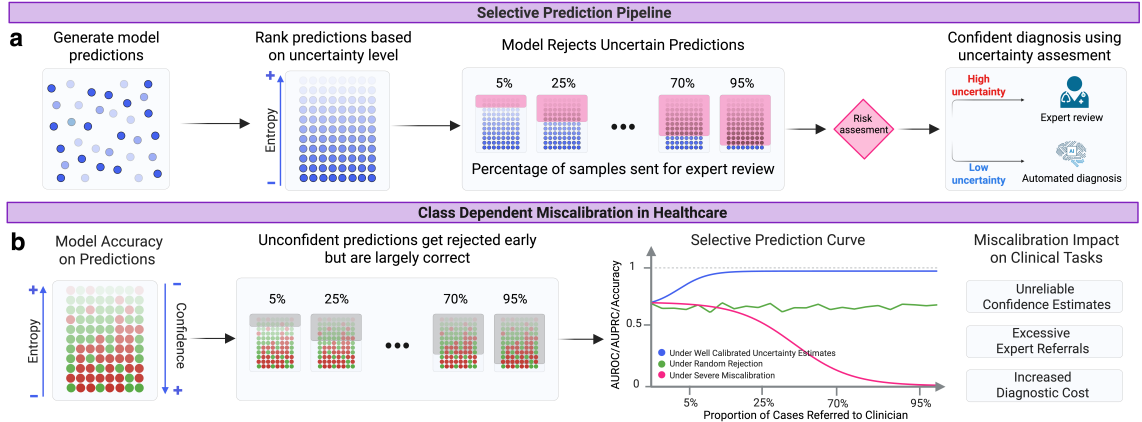}
    \caption{
     \textbf{Selective Prediction Can Serve as a Fail-Safe Mechanism in Safety-Critical Healthcare Settings.} 
    \textbf{a)} A model outputs prediction confidence/uncertainty scores; low-confidence cases are rejected and referred to an expert, yielding coverage-controlled evaluation and diagnosis.
    \textbf{b)} Miscalibration causes predictions that are mostly correct to have high uncertainty (low confidence) and incorrect predictions to have high confidence.
    Theoretical selective prediction curves show the difference between a well-calibrated model, random selection, and a severely miscalibrated model, showing improving, flat, and degrading performance with increasing coverage.
    Clinically, overconfidence risks missed or delayed diagnoses, while underconfidence triggers unnecessary rejections and shifts excess workload to clinicians.
    }
    \label{fig:main_figure}
\end{figure*}

One such mechanism is selective prediction, where a model can abstain from making a prediction and request a review by a human expert if its predictive uncertainty is high \citep{geifman2017selective}. 
By abstaining from cases of high uncertainty, selective prediction provides a fail-safe mechanism against critical errors that can endanger patients (\Cref{fig:main_figure}.a).
However, for selective prediction to improve model safety and robustness, models must be able to provide reliable uncertainty estimates. 
If uncertainty estimates are systematically miscalibrated, selective prediction may fail to identify risky cases, undermining its value in clinical decision-making (\Cref{fig:main_figure}.b).

Unfortunately, evidence from unimodal domains such as computer vision and time-series analysis shows that many models misestimate uncertainty, often being overconfident when wrong \citep{guo2017calibration,morey2025empirically, deng2025so}. 
In healthcare settings, these challenges may be further compounded when having to deal with multimodal data, where heterogeneous sources such as electronic health records (EHR), imaging, and text are integrated to provide a holistic view of a patient's state \citep{simon2025future}. 

In this paper, we study whether state-of-the-art multimodal models provide reliable uncertainty estimates that enable effective selective prediction in multilabel clinical condition classification.
We find that although multimodal models outperform unimodal baselines on standard evaluation metrics, multimodal fusion substantially degrades selective prediction, calling into question its suitability as a fail-safe mechanism in safety-critical healthcare settings (\Cref{table:summary_models}).
To understand the roots of this failure, we perform a careful analysis across unimodal and multimodal models trained on one of the largest publicly-available multimodal datasets, consisting of EHR and chest X-ray data. 
This comparison reveals that multimodal fusion does not consistently improve calibration and, in many cases, exacerbates class-dependent miscalibration. 
In particular, we consider class-dependent expected calibration errors and find that predictions are systematically overconfident, especially for conditions underrepresented in the data, undermining the effectiveness of selective prediction.

Finally, we investigate whether a simple calibration training strategy can mitigate these issues and find that it only offers modest improvements, falling short of resolving the underlying calibration shortcomings. 
Our findings underscore calibration-aware selective prediction as a key open challenge for safe deployment of clinical AI systems.
To summarize, our key contributions are:

\begin{enumerate}[topsep=0pt, align=left, leftmargin=15pt, labelindent=1pt,
listparindent=\parindent, labelwidth=0pt, itemindent=!, itemsep=3pt, parsep=0pt]
    \item
    We show that selective prediction in multilabel clinical condition classification degrades across a wide range of clinical conditions, and quantify how class-dependent miscalibration correlates with changes in selective AUC.
    
    \item
    We demonstrate that this degradation persists across state-of-the-art multimodal architectures, indicating that class-dependent miscalibration is not mitigated by architectural choice alone.
    
    \item
    We evaluate a simple loss up-weighting strategy to correct class-dependent miscalibration, showing limited but measurable improvements in selective prediction performance.

    \item
    We provide robustness analyses across alternative ECE binning strategies, Adaptive ECE, and Brier score, showing that the observed calibration--selective prediction misalignment persists across calibration metrics.
\end{enumerate}

%% file: 2_related_work.tex
\input{tables/table_summary_all_models}
\section{Related Work}
\label{sec:related_work}

Multimodal learning is increasingly used in clinical machine learning for integrating heterogeneous data sources such as structured EHR data, medical imaging, and free-text clinical reports \citep{warner2024multimodal, simon2025future}.
By combining complementary modalities, these models aim to emulate how clinicians synthesize diverse information for diagnosis and treatment planning \citep{ng2024traditional, lin2025has}.

Multimodal learning may refer to multiple imaging views of the same modality (e.g., frontal and lateral chest X-rays) \citep{warner2024multimodal} or fusion across fundamentally different data types such as imaging and tabular or textual inputs \citep{niu2024ehr, lee2024multimodal}.
While earlier work often focused on combining multiple imaging modalities for classification or segmentation \citep{sangeetha2024enhanced, adebiyi2024accurate}, recent advances in large language models and vision–language models have broadened multimodal clinical systems to include combinations of radiology images, structured vitals, free-text notes, and even patient–provider conversations \citep{mahesh2024advancing, zambrano2025clinically}. 

Despite this progress, prior work in multimodal clinical ML has paid little attention to calibration, leaving open questions about whether improvements in predictive performance translate into robust and reliable probability estimates for safe deployment \citep{zhao2024survey, deng2025so}. 
Calibration and selective prediction are closely connected in safety-critical prediction settings.
Calibration methods aim to align predicted probabilities with empirical correctness, with common post-hoc approaches including temperature scaling, Platt scaling, and isotonic regression \citep{guo2017calibration}.
Selective prediction instead allows a model to abstain from uncertain cases, but its effectiveness depends on the reliability of the confidence or uncertainty scores used to determine when to abstain \citep{geifman2017selective}.
Thus, even when a model achieves strong discrimination, miscalibrated confidence estimates can cause selective prediction to reject cases that are relatively safe while retaining cases where additional human review would be needed.

To investigate these challenges in a concrete and clinically meaningful setting, we focus on multimodal fusion of EHR time-series and chest X-ray (CXR) imaging, which is particularly relevant for critical care tasks such as diagnosis, mortality prediction, and intervention planning \citep{schilcher2024fusion}.  
Fusion architectures, such as MedFuse \citep{medfuse}, DrFuse \citep{yao2024drfuse}, and MeTra \citep{khader2023medical}, have shown that even relatively simple strategies like early or late concatenation can yield performance gains on benchmarks such as MIMIC \citep{mimiciv}. 
This paired EHR–CXR setting provides a practical and widely used benchmark for investigating how multimodal learning impacts selective prediction, calibration, and reliability in clinical settings.

Our work complements prior studies by focusing specifically on the interaction between multimodal fusion, class-dependent calibration, and selective prediction in multilabel clinical condition classification.
Rather than proposing a new calibration method, we examine whether existing multimodal architectures provide uncertainty estimates that are reliable enough to support abstention-based safety mechanisms.
This perspective addresses a gap in existing evaluations of multimodal clinical models, which often emphasize aggregate discrimination while providing less insight into whether probability estimates remain reliable across labels, classes, and clinically imbalanced conditions.


%% file: tables/table_summary_all_models.tex
\setlength{\tabcolsep}{2pt}
\begin{table*}[t!]
    \caption{
    \textbf{Performance Across Conditions.}
    Aggregate metrics suggest that multimodal learning (MedFuse) outperforms unimodal baselines. 
    However, these results mask substantial condition-specific variability. Particularly, standard calibration measures (ECE) are deceptively low, concealing severe and consistent miscalibration across individual clinical conditions. 
    This condition-specific variability, motivates the class-level and selective prediction analysis presented in our study.
    Other multimodal architectures (DrFuse, MeTra) achieve similar aggregate performance, yet, we show that they exhibit comparable calibration failures despite more complex fusion mechanisms.
    We report improvement gains over EHR given its widespread use as the foundation for most clinical prediction models.
    \small{(\textbf{Best}, \underline{Second Best}).}
    }
    \centering
    \resizebox{\linewidth}{!}{
    \begin{tabular}{l|ccccc}
    \toprule
    \multirow{2}{*}[0.5em]{\textbf{Model}}
    & \textbf{AUROC} 
    & \textbf{AUPRC} 
    & \textbf{\makecell{Selective \\ AUROC}}
    & \textbf{\makecell{Selective \\ AUPRC}}
     & \textbf{$\widehat{\mathbf{ECE}} \downarrow$}\\
    \midrule

    EHR (LSTM) \citep{lstm} & 0.681\tiny$\pm$0.007 & 0.371\tiny$\pm$0.010 & 0.625\tiny$\pm$0.115 & 0.292\tiny$\pm$0.091 & \cellcolor[gray]{0.9}\textbf{1.42\tiny$\pm$0.335}\\[1pt]

    CXR (ResNet-34) \citep{resnet50} & 0.654\tiny$\pm$0.013 & 0.350\tiny$\pm$0.012 & 0.643\tiny$\pm$0.095 & 0.342\tiny$\pm$0.077 & 9.41\tiny$\pm$0.826\\[1pt]
    \midrule

    DrFuse \citep{yao2024drfuse} & 0.726\tiny$\pm$0.007 & 0.418\tiny$\pm$0.010 & 0.677\tiny$\pm$0.139 & 0.352\tiny$\pm$0.116 & 2.54\tiny$\pm$1.15\\[1pt]

    MeTra \citep{khader2023medical} & 0.707\tiny$\pm$0.010 & 0.399\tiny$\pm$0.015 & 0.661\tiny$\pm$0.125 & 0.333\tiny$\pm$0.104 & 2.32\tiny$\pm$0.83\\[1pt]

    MedFuse \citep{medfuse} & 0.739\tiny$\pm$0.005 & \underline{0.449\tiny$\pm$0.008} & 0.705\tiny$\pm$0.136 & 0.396\tiny$\pm$0.112 & \underline{2.08\tiny$\pm$0.725}\\[1pt]

    \midrule
    
    MedFuse + Temp. Scaling \citep{guo2017calibration} & 0.739\tiny$\pm$0.005 & \underline{0.449\tiny$\pm$0.008} & 0.705\tiny$\pm$0.135 & 0.398\tiny$\pm$0.110 & 1.97\tiny$\pm$0.60\\[1pt]
       
    MedFuse + Loss-Upweighting \citep{king2001logistic} & \underline{0.742\tiny$\pm$0.006} & 0.447\tiny$\pm$0.010 & \cellcolor[gray]{0.9}\textbf{0.764\tiny$\pm$0.098} & \cellcolor[gray]{0.9}\textbf{0.532\tiny$\pm$0.093} & 4.88\tiny$\pm$1.470\\[1pt]
    
    MedFuse + Group Aware Priors \citep{rudner2024mind} & \cellcolor[gray]{0.9}\textbf{0.743\tiny$\pm$0.006} & \cellcolor[gray]{0.9}\textbf{0.451\tiny$\pm$0.008} & \underline{0.729\tiny$\pm$0.111} & \underline{0.474\tiny$\pm$0.096} & 8.31\tiny$\pm$1.061\\[1pt]
    
    \midrule

    $\Delta$ MedFuse vs EHR & \cellcolor{lightgreen}+0.058 (8.5\%) & \cellcolor{lightgreen}+0.078 (21.0\%) & \cellcolor{lightgreen}+0.080 (12.8\%) & \cellcolor{lightgreen}+0.104 (35.6\%) & \cellcolor{lightred}+0.66 (46\%)\\

    $\Delta$ MedFuse + Loss-Upweighting vs EHR
    & \cellcolor{lightgreen}+0.061 (9.0\%) 
    & \cellcolor{lightgreen}+0.076 (20.5\%) 
    & \cellcolor{lightgreen}+0.139 (22.2\%) 
    & \cellcolor{lightgreen}+0.240 (82.2\%) 
    & \cellcolor{lightred}+3.46 (243.7\%)\\

    
    \bottomrule
    \end{tabular}
    }
    \label{table:summary_models}
\end{table*}

%% file: 3_background.tex
\section{Preliminaries}
\label{sec:preliminaries}
\paragraph{Calibration.}

A core requirement for selective prediction is that model confidence scores are well-calibrated, (i.e., predicted probabilities reflect true likelihoods of correctness), which is commonly measured using the Expected Calibration Error (ECE) \citep{guo2017calibration}. 
Formally, we can measure miscalibration as the expected gap between predicted confidence and empirical accuracy:
\begin{equation}
    \mathrm{ECE}
    =
    \mathbb{E}_{\hat{P}} \big[ \, \big| \mathbb{P}(\hat{Y} =Y \mid \hat{P}=p) - p \big| \, \big] .
\end{equation}
To approximate ECE in practice, we can use a finite partition of the probability space into $M$ bins $\{B_m\}_{m=1}^M$:
\begin{equation}
    \widehat{\mathrm{ECE}} = \sum_{m=1}^M \frac{|B_m|}{n} \, \big| \mathrm{acc}(B_m) - \mathrm{conf}(B_m) \big|,
\end{equation}
where $\mathrm{acc}(B_m)$ and $\mathrm{conf}(B_m)$ denote the average accuracy and confidence of samples in bin $B_m$, respectively, and $n$ is the number of samples.
Additionally, we may wish to compute the conditional calibration of a predictor.
To do so, we can define a conditional expected calibration error,
\begin{equation}
    \mathrm{ECE}_c
    =
    \mathbb{E}_{\hat{P} \mid \hat{Y} = c} \big[ \, \big| \mathbb{P}(\hat{Y} = c \mid \hat{P}=p) - p \big| \, \big] .
    \label{eq:class_ece}
\end{equation}
which captures class-dependent calibration.
We can approximate the conditional calibration error by only considering the subset of points with label $c$ and, again, specifying a finite partition of the probability space into $M$ bins $\{B_m^c\}_{m=1}^M$:
\begin{equation}
    \widehat{\mathrm{ECE}}_{c} = \sum_{m=1}^M \frac{|B_m^c|}{n_c} \, \big| \mathrm{acc}(B_m^c) - \mathrm{conf}(B_m^c) \big|,
\end{equation}
where $\mathrm{acc}(B_m^c)$ and $\mathrm{conf}(B_m^c)$ denote the average accuracy and confidence of samples in bin $B_m^c$, respectively, and $n_c$ is the number of samples whose label is $c$.

\paragraph{Selective Prediction.}
Selective prediction modifies the standard prediction pipeline by introducing a ``reject option'' denoted by $\bot$, using a gating mechanism defined by a selection function $s:\mathcal{X}\to\mathbb{R}$ that determines whether a prediction should be made for a given input point $\mathbf{x}\in\mathcal{X}$  \citep{el2010foundations}.
For a rejection threshold $\tau$, with $s$ representing the entropy of $\mathbf{x}$, the prediction model is given by:
\begin{equation}
    (p(\mathbf{y}\,|\,\cdot;f),s)(\mathbf{x}) =  
    \begin{cases}
          p(\mathbf{y}\,|\,\mathbf{x};f), & \text{if}\ s\le \tau \\
          \bot, & \text{otherwise,}
    \end{cases}
\end{equation}
with $p(\mathbf{y} \mid \cdot; f)$ being a model's predictive distribution.
A variety of methods have been proposed to find a selection function $s$ \citep{rabanser2022selective, rabanser2025does}, with model's predictive uncertainty being a common choice: $s = \mathcal{H}(p(\mathbf{y} \mid \mathbf{x}; f))$.
That is, a point $\mathbf{x} \in \mathcal{X}$ will be placed into the rejection class if the model's predictive uncertainty is above a certain threshold.

To evaluate the predictive performance of a prediction model $(p(\mathbf{y} \mid \cdot; f), s)(\mathbf{x})$, we compute the predictive performance of the classifier $p(\mathbf{y} \mid \mathbf{x}; f)$ over a range of thresholds $\tau$ (\Cref{fig:main_figure}.a). 
Performance is then assessed with standard metrics (e.g., accuracy, AUROC, AUPRC, or ECE), yielding selective prediction curves that capture the trade-off between coverage, the proportion of cases for which predictions are made, and performance.

\paragraph{Predictive Uncertainty Quantification.}
Furthermore, understanding calibration requires examining how models quantify predictive uncertainty, which is typically decomposed into aleatoric uncertainty, arising from inherent noise in the data, and epistemic uncertainty, arising from limited knowledge about the model or parameters \citep{depeweg2018decomposition}. 
A common way to quantify predictive uncertainty is via the Shannon entropy of the predictive distribution:
\begin{equation}
    \mathcal{H}(p(\mathbf{y} \mid \mathbf{x})) = - \sum_{c} p(\mathbf{y} = c \mid \mathbf{x}) \log p(\mathbf{y} = c \mid \mathbf{x}),
\end{equation}
where higher entropy indicates greater uncertainty \citep{rudner2023fseb, rudner2023fsmap}.

%% file: 4_empirical_setup.tex
\section{Empirical Setup}
\label{sec:methods}


\subsection{Clinical Task and Dataset}

\paragraph{Clinical Condition Classification.}
We study multilabel clinical condition classification, where the goal is to predict whether a patient presents any of 25 chronic, acute, or mixed clinical conditions during an ICU stay \citep{medfuse}.
This task is particularly relevant for analyzing selective prediction and calibration because each condition corresponds to a distinct patient subpopulation with varying prevalence, severity, and clinical consequences.
As a result, strong average performance across labels may obscure systematic failures for underrepresented or high-risk conditions, leading to unreliable predictions for specific patient groups.

In safety-critical clinical settings, accurately identifying uncertainty at the condition level is essential: miscalibration for even a small subset of conditions can result in missed diagnoses or inappropriate deferral behavior, despite seemingly robust aggregate performance.
The multilabel nature of this task therefore provides a natural and clinically meaningful setting to study class-dependent miscalibration and its impact on selective prediction.
The full list of clinical conditions and their prevalence is provided in \Cref{app:clinical_condition_distribution} and \Cref{table:clinical_info}.

\paragraph{Multimodal Dataset.}
We construct paired multimodal samples from MIMIC-IV \citep{mimiciv} (structured EHR time-series) and MIMIC-CXR \citep{mimiccxr} (frontal-view chest X-rays), such that each patient sample contains both modalities, ${x}^{ehr}$ and ${x}^{cxr}$, along with multilabel ground truth ${y}$.
We follow standard preprocessing procedures and ensure patient-level separation between training and test splits, consistent with prior work \citep{medfuse}.

\subsection{Model Variants}

Throughout this study, we use \emph{unimodal} to refer to models that operate on a single clinical data source, such as structured EHR time-series or chest radiographs alone.
We use \emph{multimodal} to refer to models that combine two or more clinical data sources before prediction, allowing the information to be integrated within a single classifier.

\paragraph{Base Architecture.} We adopt the MedFuse architecture \citep{medfuse} to train a multimodal classifier $f(\cdot)$ on paired EHR and CXR data 
$\mathcal{D}=\{(x^{ehr}_n, x^{cxr}_n, y_n)\}_{n=1}^{N}$. 
EHR time-series are encoded using a two-layer LSTM ($\Phi_{\textrm{ehr}}$) \citep{lstm}, and CXR images using a ResNet-34 ($\Phi_{\textrm{cxr}}$) \citep{resnet50}, with latent representations concatenated and passed to a classification head $g(\cdot)$ with sigmoid outputs. 
Models are trained with the standard binary cross-entropy loss.

\paragraph{Unimodal Baselines.}
We train unimodal models using only $\Phi_{\textrm{ehr}}$ (unimodal EHR) and $\Phi_{\textrm{cxr}}$ (unimodal CXR) to establish reference points for performance and calibration. 
The CXR encoder is initialized with ImageNet weights, and the EHR encoder is trained from scratch.

\paragraph{Multimodal Architectures.}
The original MedFuse fusion model serves as our deterministic multimodal baseline, allowing us to measure the calibration impact of modality fusion and providing a reference for uncertainty-aware variants.
To assess the robustness of our findings to the choice of architecture, we additionally evaluate two alternative multimodal backbones that integrate EHR and CXR modalities for clinical condition classification: DrFuse \citep{yao2024drfuse}, which learns representations with divergence-based alignment, and MeTra \citep{khader2023medical}, which encodes images and clinical variables using a transformer-based cross-modal fusion encoder. 

\paragraph{Label-Dependent Loss-Upweighting.}
To probe class-dependent miscalibration and assess whether it can be mitigated, we apply a simple label-dependent loss-upweighting scheme: for each condition $c$, a weight $w_c$ upweights low-prevalence positive labels during training. 
This is not intended as a complete solution, but as a lightweight intervention to quantify its effect on class-level calibration and selective prediction across multimodal backbones (MedFuse, DrFuse, MeTra).

\subsection{Evaluation Metrics}
We report discrimination metrics (AUROC, AUPRC), standard calibration (ECE), and class-stratified calibration metrics (ECE$_{c=0}$, ECE$_{c=1}$).
Additional robustness metrics, including ECE across different bins, Adaptive ECE and Brier score, are reported in \Cref{sec:extended_calibration_metrics}.
Selective prediction performance is evaluated using selective AUROC and selective AUPRC.
We summarize selective prediction performance by the area under the resulting selective prediction curves.

%% file: 5_results.tex
\section{Results}

We present empirical results analyzing selective prediction and calibration behavior in multilabel clinical condition classification using multimodal ICU data.
Our findings are organized around four guiding questions:

\begin{enumerate}\setlength\itemsep{-0.3em}
    \item Do multimodal models improve or degrade selective prediction performance in this task?
    \item How does calibration influence selective prediction behavior?
    \item Are class-dependent calibration effects consistent across multimodal architectures?
    \item Can loss-upweighting mitigate observed calibration failures?
\end{enumerate}

Our experiments are conducted on a commonly used ICU benchmark pairing structured EHR time-series with chest X-ray images.
This multimodal setting has been widely adopted in prior work on clinical risk prediction and patient deterioration modeling, serving as a practical testbed for studying multimodal fusion in critical care contexts \citep{yang2021multimodal,khader2023medical,wang2024multimodal,insalata2024multimodal,yao2024drfuse,zheng2025multimodal}.
Within this setting, our results characterize how calibration and selective prediction behave for multilabel clinical condition classification, and should be interpreted as task- and data-specific rather than as general claims about multimodal clinical models.
Additional qualitative results and analysis are provided in \Cref{sec:further_empirical_results,sec:appendix_evaluation_across_selected_condition,sec:regression_analysis}, offering complementary views of the findings discussed below.

\input{tables/table_ece_unimodals}
\subsection{Do Multimodal Models Improve or Degrade Selective Prediction Performance in This Task?}

\paragraph{Overall Performance.}
Across the 25 clinical conditions, the multimodal baseline MedFuse generally achieves higher AUROC, AUPRC, and selective AUROC than both unimodal variants, often with statistically significant improvements.
These gains confirm that multimodal fusion improves discrimination in this task.
However, improvements in discrimination are not accompanied by consistent gains in calibration, revealing a recurrent mismatch between predictive accuracy and probability alignment.

For example, in \textit{Acute and unspecified renal failure (1)}, MedFuse achieves the highest AUROC (0.761) and AUPRC (0.589), and reduces calibration error relative to CXR (ECE: 12.77 vs.\ 2.31), yet EHR remains the best calibrated model overall (ECE: 1.62).
In \textit{Coronary atherosclerosis (10)}, MedFuse nearly doubles EHR’s calibration error (3.52 vs.\ 1.80) while still achieving the strongest discrimination across all other metrics.
The only condition for which MedFuse significantly outperforms both unimodal variants in calibration is \textit{Other upper respiratory disease (20)} (MedFuse: 0.68 vs.\ CXR: 3.03, EHR: 0.98); however, this improvement does not translate into statistically superior AUROC, AUPRC, or selective prediction performance.
Overall, these results indicate that multimodal fusion reliably improves discrimination but does not consistently improve, and can in some cases worsen, calibration.
Detailed per-condition results for unimodal and multimodal models are reported in \Cref{table:eval_unimodals} in \Cref{app:uni_multi_sel_pred}.

\begin{figure*}[t!]
    \centering
    \includegraphics[width=\textwidth]{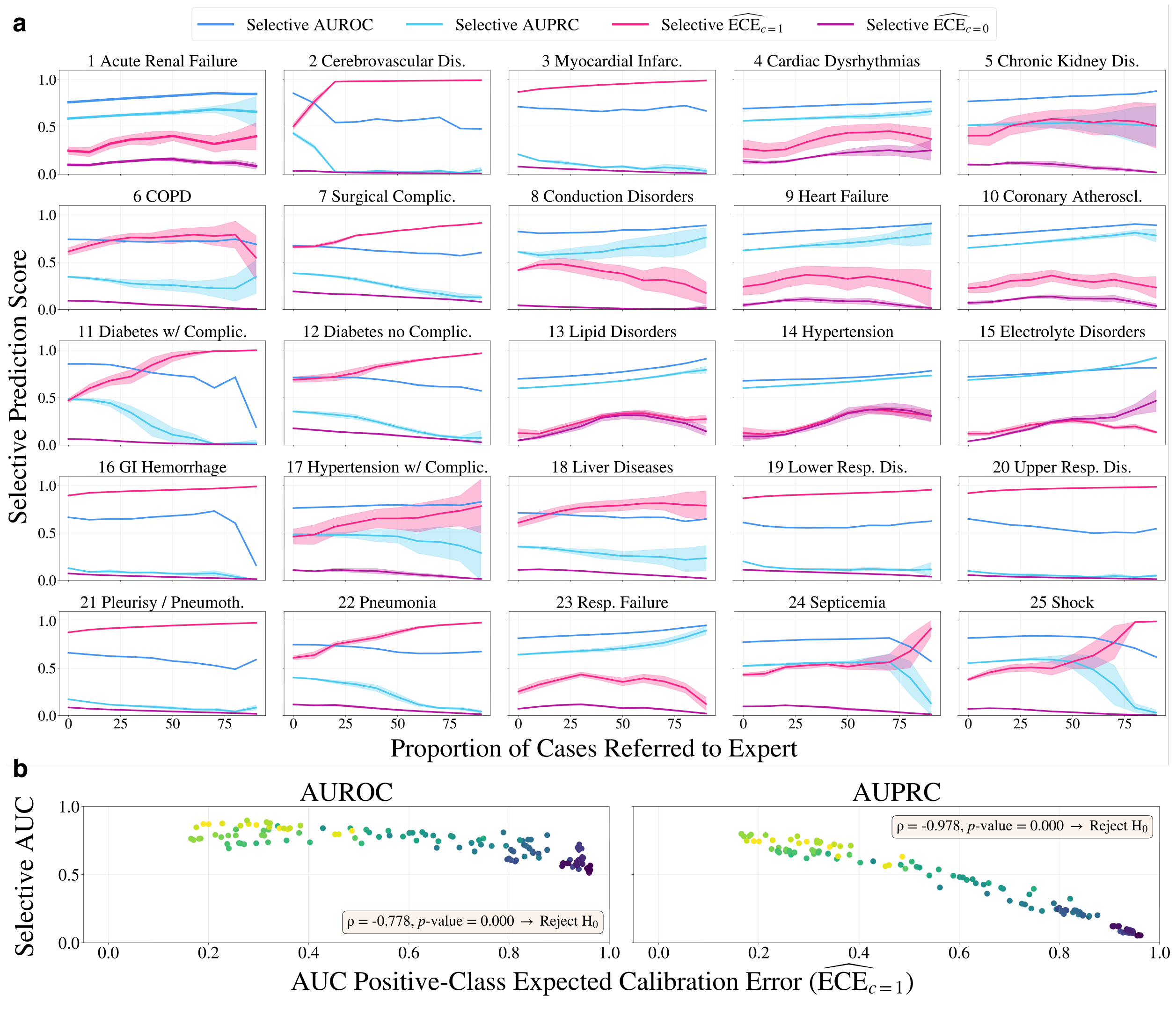}
    \caption{
     \textbf{How Does Calibration Influence Selective Prediction Behavior?}
    \textbf{(a)} Selective AUROC, selective AUPRC, and class-stratified ECE across 25 clinical conditions for MedFuse as low-confidence predictions are rejected.
    \textbf{(b)} Relationship between positive-class calibration error (ECE$_{c=1}$ AUC) and selective AUROC/AUPRC across conditions and random seeds, showing worse selective performance with higher ECE$_{c=1}$ \footnotesize{(Spearman's rank, $p<0.05$. Shaded region: variance across seeds.})
    }
    \label{fig:det_class_level_ece}
\end{figure*}

\paragraph{Class-Level Calibration Stratification.}
To better understand these effects, we decompose expected calibration error into positive- and negative-class components, ECE$_{c=1}$ and ECE$_{c=0}$, as defined in \Cref{eq:class_ece}.
As shown in \Cref{table:ece_unimodals}, positive-class calibration error is consistently larger than negative-class error across conditions, indicating that overconfidence in positive predictions is the dominant source of miscalibration.
This imbalance is obscured when reporting only standard ECE (\Cref{table:summary_models}), which can therefore lead to misleading conclusions about model robustness.

We further observe that no single model dominates across all calibration components.
EHR and MedFuse tend to achieve lower positive-class ECE than CXR, whereas CXR often exhibits the lowest negative-class ECE.
This inversion illustrates how aggregate calibration metrics can mask complementary strengths and weaknesses across modalities, and why positive-class miscalibration is particularly detrimental to selective prediction in this task.
These findings suggest that any observed improvements in positive-class calibration for the multimodal model are primarily inherited from the EHR modality rather than arising from multimodal fusion itself.

\begin{tcolorbox}[colback=nyu!11, colframe=nyu, arc=1mm, outer arc=1mm, colbacktitle=nyu, coltitle=white, fonttitle=\bfseries, title={Finding 1}]
Multimodal fusion improves discrimination in multilabel clinical condition classification, but does not reliably improve calibration; overconfidence in underrepresented class is the dominant source of error.
\end{tcolorbox}

\begin{figure*}[t!]
    \centering
    \includegraphics[width=0.98\textwidth]{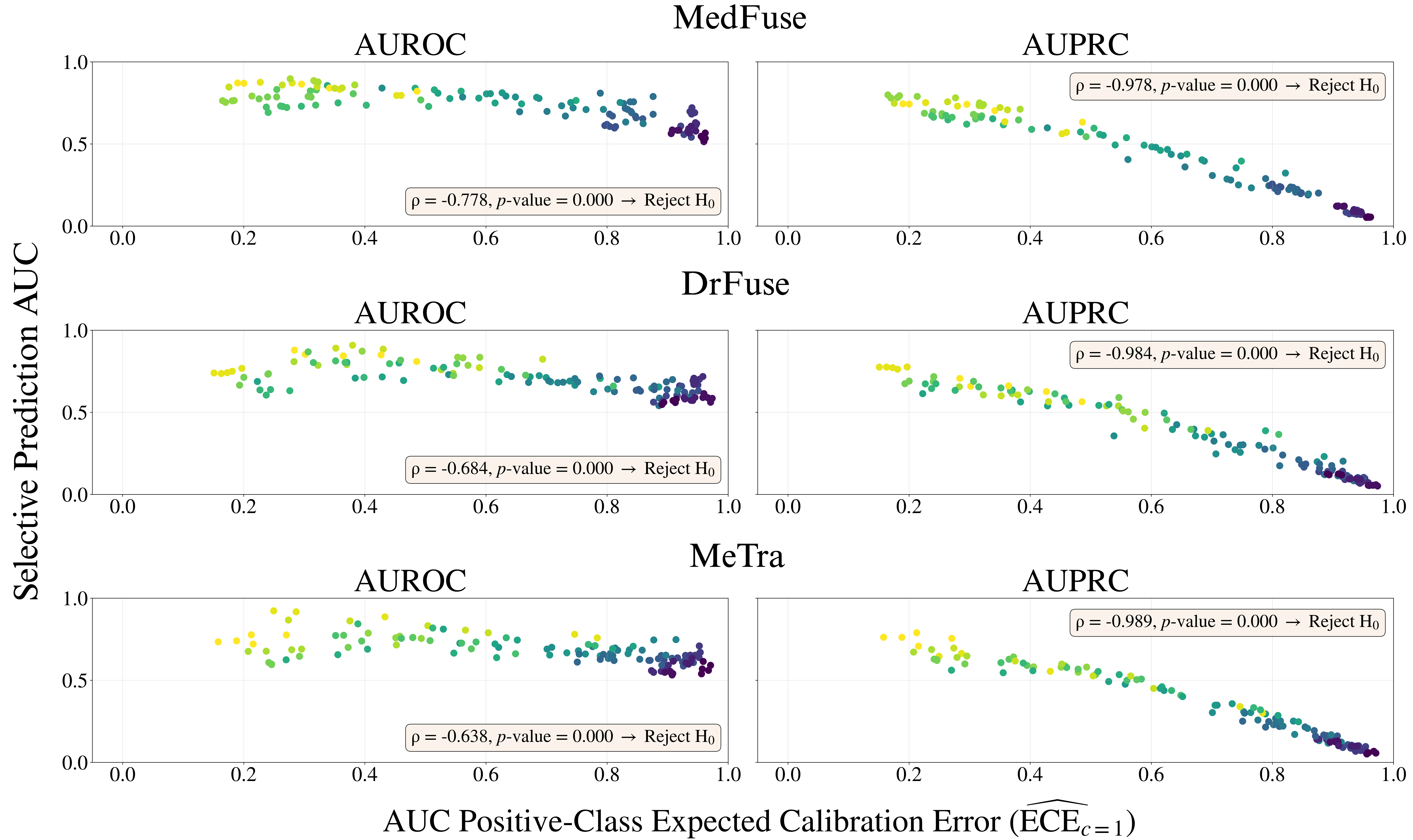}
    \caption{
    \textbf{Are Class-dependent Calibration Effects Consistent Across Multimodal Architectures?}
    Across MedFuse, DrFuse, and MeTRA, higher calibration error in the underrepresented class (i.e., positive class in this task) is consistently associated with lower selective AUROC and AUPRC across conditions.
    For all architectures and metrics, the null hypothesis of no association between stratified ECE and selective AUC is rejected, indicating that architectural complexity alone does not mitigate calibration-driven selective prediction failures.
    \footnotesize{(Spearman's rank $\mathrm{H}_0$, $p<0.05$)}
    }
    \label{fig:new_multimodals_scatter}
\end{figure*} 

\subsection{How Does Calibration Influence Selective Prediction Behavior?}
\label{subsec:sel_pred}

\paragraph{Calibration Drives Selective Performance.}
We first inspect selective prediction thresholds for all 25 conditions using MedFuse, plotting selective AUROC and AUPRC alongside stratified ECE (\Cref{fig:det_class_level_ece}.a). 
While selective metrics should monotonically improve as uncertain cases are rejected, we observe systematic departures from this expected behavior across many conditions.  
At extreme thresholds, selective AUROC and AUPRC collapse when the class distribution becomes highly imbalanced, and in mid-threshold regimes metrics often stagnate or degrade, suggesting that the model does not consistently reject the ``right'' cases.

Linking these failures to calibration reveals clear patterns.  
Conditions with low, balanced positive and negative ECE exhibit stable and improving selective AUROC/AUPRC (e.g., Labels 13 to 15).  
By contrast, high ECE in the underrepresented class, corresponding to the positive class in this task, produces noisy, unstable selective curves: AUROC hovers near chance and AUPRC declines sharply (e.g., Labels 2, 22, 24 and 25).
Variance across random seeds follows the same trend, with poor calibration associated with wide, unstable spreads.  

\paragraph{Calibration as a Predictor of Evaluation Stability.}
Motivated by these trends, we quantify the relationship between calibration and selective behavior across conditions (\Cref{fig:det_class_level_ece}.b). 
Positive-class ECE AUC and Selective AUC are consistently negatively correlated: higher ECE$_{c=1}$ predicts smaller or even negative changes in selective prediction metrics.  
The effect is strongest for AUPRC, while AUROC shows greater variability but follows the same direction.  
We test the null hypothesis of no association between ECE and Selective AUC and reject it for both metrics (Spearman's rank correlation, $p<0.05$).

Overall, these findings establish calibration of the underrepresented class as a reliable predictor of selective prediction stability.
In this clinical task, miscalibration is dominated by the positive class due to class imbalance; however, high ECE in any minority class serves as an early warning that selective prediction will fail to yield meaningful improvements over standard evaluation.

\begin{tcolorbox}[colback=nyu!11, colframe=nyu, arc=1mm, outer arc=1mm, colbacktitle=nyu, coltitle=white, fonttitle=\bfseries, title={Finding 2}
]
Calibration of the underrepresented class in multilabel clinical condition classification is a leading indicator of selective reliability: when ECE is high for the minority class, selective AUROC/AUPRC fail to improve and can deteriorate.
\end{tcolorbox}

\subsection{Are Class-Dependent Calibration Effects Consistent Across Multimodal Architectures?}
\label{subsec:drfuse_metra}

\paragraph{Extending to DrFuse and MeTRA.}
To assess whether the observed relationship between calibration and selective prediction is specific to MedFuse or persists across architectures, we extend our analysis to DrFuse \citep{yao2024drfuse} and MeTRA \citep{khader2023medical}, two multimodal models with substantially different fusion mechanisms.

To facilitate comparison, we adopt the same analysis relating class-dependent calibration to selective prediction performance, focusing on the underrepresented class, which corresponds to the positive class in this task.
As shown in \Cref{fig:new_multimodals_scatter}, all three multimodal architectures exhibit a consistent downward trend for both AUROC and AUPRC: higher calibration error in the minority class reliably predicts degraded selective performance.
This shared behavior indicates that the observed selective prediction failures are not an artifact of a specific fusion strategy or architectural complexity, but instead reflect a common consequence of class imbalance in this multilabel clinical condition classification setting.

Additional results supporting this comparison across architectures are provided in \Cref{app:takeaway_2}, including aggregate discrimination and selective prediction metrics, class-stratified ECE tables, and per-condition selective prediction curves for DrFuse and MeTRA.

\begin{tcolorbox}[colback=nyu!11, colframe=nyu, arc=1mm, outer arc=1mm, colbacktitle=nyu, coltitle=white, fonttitle=\bfseries, title={Finding 3}]
Across multiple multimodal fusion architectures, class-dependent miscalibration in the underrepresented class persists and is consistently associated with degraded selective prediction performance in multilabel clinical condition classification.
\end{tcolorbox}

\begin{figure*}[t!]
    \centering
    \includegraphics[width=0.98\textwidth]{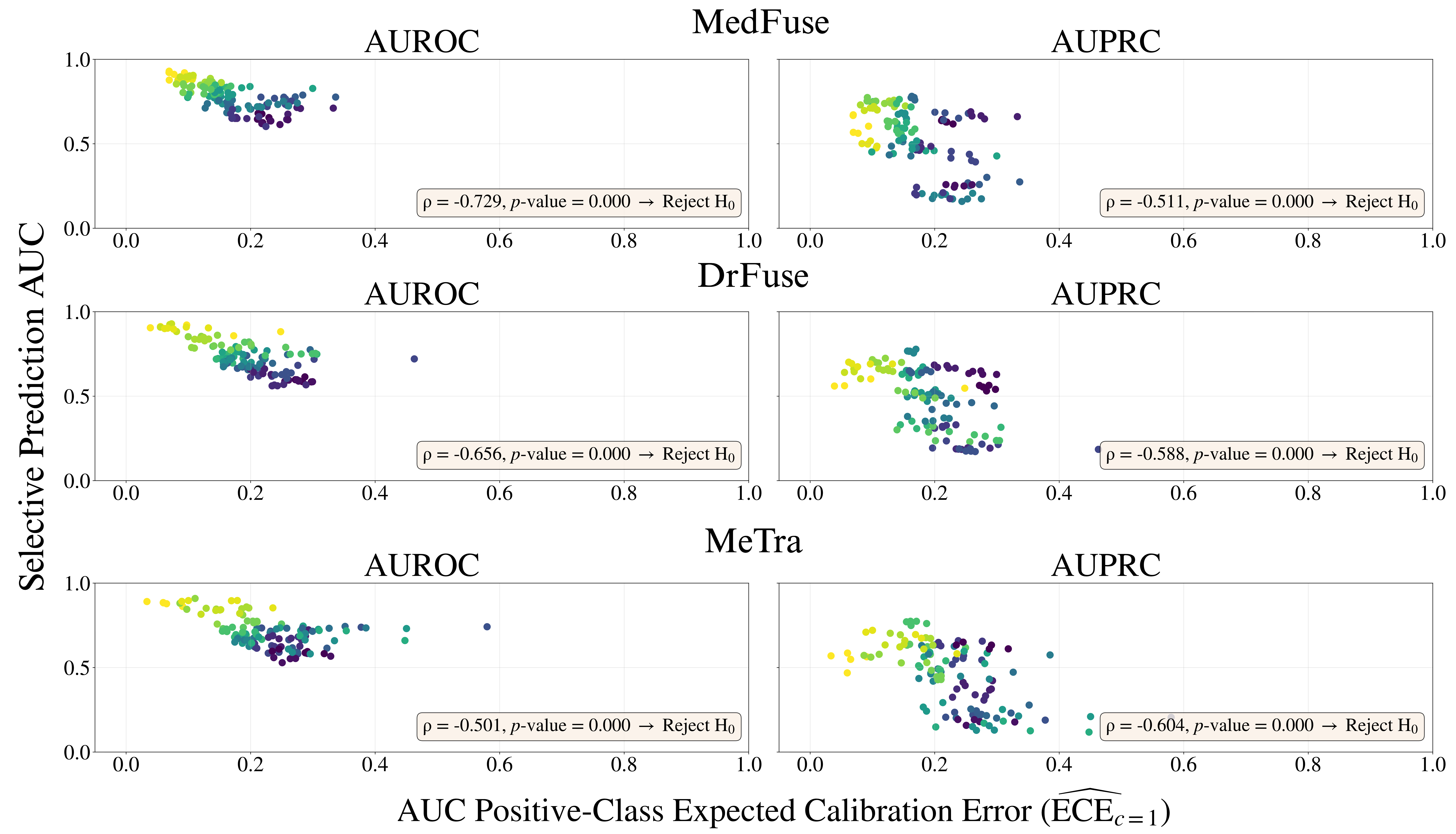}
    \caption{
    \textbf{Can Loss-Upweighting Mitigate Observed Calibration Failures?}
    Across architectures loss-upweighting reduces calibration error for the underrepresented class, but the negative relationship between stratified ECE and selective AUROC/AUPRC persists across conditions.
    For all architectures and metrics, the null hypothesis of no association between ECE and selective AUC is rejected \footnotesize{(Spearman’s rank, $p<0.05$)}.
    }   \label{fig:new_multimodals_ups_scatter}
\end{figure*}

\subsection{Can Loss-Upweighting Mitigate Observed Calibration Failures?}
\label{sec:loss_upweight}

\paragraph{Persistent Reliability Gaps.}
To assess whether class-dependent miscalibration can be mitigated, we apply a loss-upweighting strategy that emphasizes low-prevalence labels during training.
This intervention is evaluated consistently across all multimodal architectures considered in this study, including MedFuse, DrFuse, and MeTRA.
As shown in \Cref{fig:new_multimodals_ups_scatter}, loss upweighting systematically reduces calibration error for the underrepresented class across conditions and architectures, indicating decreased overconfidence for rare outcomes.
However, selective prediction behavior remains largely unchanged.

Across models, selective AUROC and AUPRC exhibit only small average increases that are highly condition-dependent and rarely statistically significant.
Moreover, the negative relationship between class-dependent calibration error and selective performance persists, indicating that improved calibration alone is insufficient to ensure reliable selective prediction in this setting.

Detailed per-condition and architecture-specific analyses are provided in \Cref{sec:takeaway_3}.
In \Cref{sec:extended_calibration_metrics}, we further assess whether our findings are sensitive to the calibration metric or binning strategy by reporting results across alternative ECE bin counts, Adaptive ECE, and Brier score.

These robustness analyses show that the observed calibration--selective prediction misalignment persists across evaluation choices: while aggregate ECE shows no consistent global improvement, class-stratified metrics confirm that loss-upweighting primarily affects calibration of the underrepresented class, with weaker and less consistent effects for the majority class.

\begin{tcolorbox}[colback=nyu!11, colframe=nyu, arc=1mm, outer arc=1mm, colbacktitle=nyu, coltitle=white, fonttitle=\bfseries, title={Finding 4}]
Loss upweighting for underrepresented labels yields measurable improvements in class-dependent calibration, but does not consistently translate into improved selective prediction reliability in multilabel clinical condition classification.
\end{tcolorbox}

%% file: tables/table_ece_unimodals.tex
\setlength{\tabcolsep}{6pt}
\begin{table*}[t!]
\caption{\textbf{Class-level Calibration Stratification.}
Aggregate ECE (standard metric) appears low but obscures severe class-specific miscalibration. 
Across conditions and modality settings, positive-class ECE significantly dominates, revealing systematic overconfidence for patients who present the condition and poor calibration across modalities.
\footnotesize{(Dark-bold: Wilcoxon $p<0.05$, 5 seeds; Light-bold: highest mean, not significant)}}
    \centering
    \scriptsize
    \begin{adjustbox}{max width=\linewidth}
    \begin{tabular}{c|ccc|ccc|cccc}
    \multirow{2}{*}[-1.5em]{\textbf{\makecell{Clinical \\ Condition}}} 
    & \multicolumn{3}{c}{\textbf{EHR}}
    & \multicolumn{3}{c}{\textbf{CXR}}
    & \multicolumn{3}{c}{\textbf{MedFuse}}\\
    \midrule
    & \textbf{$\widehat{\mathbf{ECE}} \downarrow$}
    & \textbf{$\widehat{\mathbf{ECE}}_{c=1} \downarrow $}
    & \textbf{$\widehat{\mathbf{ECE}}_{c=0} \downarrow$}
    
     & \textbf{$\widehat{\mathbf{ECE}} \downarrow$}
    & \textbf{$\widehat{\mathbf{ECE}}_{c=1} \downarrow$}
    & \textbf{$\widehat{\mathbf{ECE}}_{c=0} \downarrow$}
    
     & \textbf{$\widehat{\mathbf{ECE}} \downarrow$}
    & \textbf{$\widehat{\mathbf{ECE}}_{c=1} \downarrow$}
    & \textbf{$\widehat{\mathbf{ECE}}_{c=0} \downarrow$}
    \\[2pt]
    \midrule

    1 & \cellcolor[gray]{0.95}\textbf{1.62\tiny$\pm$0.52} & 30.76\tiny$\pm$1.82 & 14.32\tiny$\pm$1.14 & 12.77\tiny$\pm$0.98 & 35.26\tiny$\pm$5.07 & \cellcolor[gray]{0.85}\textbf{3.63\tiny$\pm$1.74} & 2.31\tiny$\pm$0.97 & \cellcolor[gray]{0.95}\textbf{24.85\tiny$\pm$3.96} & 9.97\tiny$\pm$1.14 \\[3pt]

    2 & 1.08\tiny$\pm$0.14 & \cellcolor[gray]{0.95}\textbf{47.76\tiny$\pm$3.00} & 4.68\tiny$\pm$0.25 & 3.99\tiny$\pm$0.31 & 83.21\tiny$\pm$3.28 & 4.13\tiny$\pm$0.75 & \cellcolor[gray]{0.95}\textbf{0.88\tiny$\pm$0.24} & 50.56\tiny$\pm$3.07 & \cellcolor[gray]{0.95}\textbf{3.56\tiny$\pm$0.19} \\[3pt]
    
    3 & \cellcolor[gray]{0.95}\textbf{0.73\tiny$\pm$0.07} & 88.27\tiny$\pm$0.33 & 8.22\tiny$\pm$0.06 & 4.35\tiny$\pm$0.42 & \cellcolor[gray]{0.95}\textbf{80.59\tiny$\pm$5.85} & \cellcolor[gray]{0.85}\textbf{4.91\tiny$\pm$0.70} & 0.76\tiny$\pm$0.23 & 86.93\tiny$\pm$0.14 & 8.05\tiny$\pm$0.25 \\[3pt]
    
    4 & \cellcolor[gray]{0.85}\textbf{1.15\tiny$\pm$0.67} & 49.64\tiny$\pm$1.14 & 28.93\tiny$\pm$0.65 & 12.50\tiny$\pm$0.48 & 29.02\tiny$\pm$5.95 & \cellcolor[gray]{0.85}\textbf{4.34\tiny$\pm$3.01} & 2.61\tiny$\pm$1.65 & \cellcolor[gray]{0.95}\textbf{26.90\tiny$\pm$9.52} & 13.57\tiny$\pm$2.63 \\[3pt]
    
    5 & \cellcolor[gray]{0.95}\textbf{2.13\tiny$\pm$0.45} & 67.12\tiny$\pm$0.76 & 19.71\tiny$\pm$0.41 & 15.37\tiny$\pm$0.37 & 45.19\tiny$\pm$2.51 & \cellcolor[gray]{0.85}\textbf{5.92\tiny$\pm$0.92} & 2.34\tiny$\pm$2.04 & \cellcolor[gray]{0.95}\textbf{40.71\tiny$\pm$8.56} & 10.21\tiny$\pm$1.06 \\[3pt]
    
    6 & \cellcolor[gray]{0.95}\textbf{1.87\tiny$\pm$0.23} & 81.50\tiny$\pm$0.22 & 16.00\tiny$\pm$0.07 & 7.15\tiny$\pm$1.26 & 64.69\tiny$\pm$14.12 & \cellcolor[gray]{0.85}\textbf{3.32\tiny$\pm$1.04} & 2.17\tiny$\pm$0.45 & \cellcolor[gray]{0.95}\textbf{61.34\tiny$\pm$3.97} & 9.27\tiny$\pm$0.68 \\[3pt]
    
    7 & \cellcolor[gray]{0.95}\textbf{1.16\tiny$\pm$0.45} & 70.71\tiny$\pm$1.80 & 20.18\tiny$\pm$0.20 & 9.15\tiny$\pm$0.58 & \cellcolor[gray]{0.95}\textbf{66.07\tiny$\pm$3.97} & \cellcolor[gray]{0.85}\textbf{7.68\tiny$\pm$1.09} & 1.46\tiny$\pm$0.31 & 66.13\tiny$\pm$1.23 & 19.12\tiny$\pm$0.40 \\[3pt]
    
    8 & \cellcolor[gray]{0.85}\textbf{1.09\tiny$\pm$0.19} & 87.20\tiny$\pm$0.28 & 10.42\tiny$\pm$0.04 & 3.42\tiny$\pm$0.45 & \cellcolor[gray]{0.95}\textbf{38.99\tiny$\pm$6.15} & \cellcolor[gray]{0.85}\textbf{2.77\tiny$\pm$0.79} & 2.32\tiny$\pm$0.45 & 41.64\tiny$\pm$0.88 & 4.38\tiny$\pm$0.89 \\[3pt]
    
    9 & \cellcolor[gray]{0.85}\textbf{1.81\tiny$\pm$0.56} & 41.15\tiny$\pm$1.10 & 17.36\tiny$\pm$0.89 & 11.92\tiny$\pm$1.95 & \cellcolor[gray]{0.95}\textbf{20.38\tiny$\pm$5.59} & 8.40\tiny$\pm$4.88 & 4.50\tiny$\pm$1.71 & 24.13\tiny$\pm$9.07 & \cellcolor[gray]{0.95}\textbf{4.69\tiny$\pm$1.96} \\[3pt]
    
    10 & \cellcolor[gray]{0.85}\textbf{1.80\tiny$\pm$0.43} & 38.09\tiny$\pm$0.54 & 17.50\tiny$\pm$0.76 & 11.15\tiny$\pm$1.04 & 30.34\tiny$\pm$5.39 & \cellcolor[gray]{0.95}\textbf{3.44\tiny$\pm$2.02} & 3.52\tiny$\pm$1.29 & \cellcolor[gray]{0.95}\textbf{22.44\tiny$\pm$4.96} & 7.08\tiny$\pm$1.88 \\[3pt]
    
    11 & 1.41\tiny$\pm$0.26 & 80.37\tiny$\pm$1.11 & 10.92\tiny$\pm$0.15 & 5.42\tiny$\pm$0.83 & 74.92\tiny$\pm$6.19 & \cellcolor[gray]{0.95}\textbf{5.25\tiny$\pm$0.72} & \cellcolor[gray]{0.95}\textbf{1.15\tiny$\pm$0.22} & \cellcolor[gray]{0.85}\textbf{47.13\tiny$\pm$2.20} & 6.30\tiny$\pm$0.34 \\[3pt]
    
    12 & \cellcolor[gray]{0.95}\textbf{1.38\tiny$\pm$0.28} & 77.14\tiny$\pm$0.38 & 19.78\tiny$\pm$0.07 & 8.45\tiny$\pm$1.35 & \cellcolor[gray]{0.95}\textbf{62.23\tiny$\pm$7.64} & \cellcolor[gray]{0.85}\textbf{6.34\tiny$\pm$2.18} & 1.74\tiny$\pm$0.56 & 68.84\tiny$\pm$2.85 & 17.64\tiny$\pm$0.79 \\[3pt]
    
    13 & \cellcolor[gray]{0.85}\textbf{0.99\tiny$\pm$0.42} & 22.03\tiny$\pm$1.04 & 15.95\tiny$\pm$0.73 & 15.71\tiny$\pm$0.34 & \cellcolor[gray]{0.95}\textbf{12.50\tiny$\pm$9.56} & 18.16\tiny$\pm$6.99 & 3.61\tiny$\pm$0.64 & 12.51\tiny$\pm$4.83 & \cellcolor[gray]{0.85}\textbf{4.88\tiny$\pm$0.85} \\[3pt]
    
    14 & 2.77\tiny$\pm$0.48 & 26.80\tiny$\pm$2.53 & 25.28\tiny$\pm$1.99 & 18.90\tiny$\pm$1.04 & 24.94\tiny$\pm$8.12 & 14.45\tiny$\pm$7.33 & \cellcolor[gray]{0.95}\textbf{1.86\tiny$\pm$0.62} & \cellcolor[gray]{0.95}\textbf{12.64\tiny$\pm$5.92} & \cellcolor[gray]{0.95}\textbf{9.09\tiny$\pm$4.73} \\[3pt]
    
    15 & \cellcolor[gray]{0.85}\textbf{1.41\tiny$\pm$0.42} & \cellcolor[gray]{0.95}\textbf{10.36\tiny$\pm$0.73} & 8.89\tiny$\pm$0.74 & 16.93\tiny$\pm$1.17 & 26.86\tiny$\pm$3.30 & 8.69\tiny$\pm$1.27 & 3.39\tiny$\pm$0.92 & 12.05\tiny$\pm$2.66 & \cellcolor[gray]{0.85}\textbf{3.95\tiny$\pm$0.52} \\[3pt]
    
    16 & \cellcolor[gray]{0.85}\textbf{0.58\tiny$\pm$0.14} & 91.64\tiny$\pm$0.15 & 6.89\tiny$\pm$0.09 & 4.36\tiny$\pm$0.27 & \cellcolor[gray]{0.95}\textbf{87.04\tiny$\pm$3.51} & \cellcolor[gray]{0.85}\textbf{5.26\tiny$\pm$0.63} & 1.05\tiny$\pm$0.36 & 89.64\tiny$\pm$0.80 & 7.21\tiny$\pm$0.42 \\[3pt]
    
    17 & \cellcolor[gray]{0.95}\textbf{1.91\tiny$\pm$0.27} & 72.75\tiny$\pm$0.85 & 18.83\tiny$\pm$0.48 & 14.12\tiny$\pm$0.48 & 50.40\tiny$\pm$1.45 & \cellcolor[gray]{0.85}\textbf{4.05\tiny$\pm$0.56} & 2.70\tiny$\pm$1.70 & \cellcolor[gray]{0.95}\textbf{46.22\tiny$\pm$7.91} & 10.70\tiny$\pm$0.96 \\[3pt]
    
    18 & \cellcolor[gray]{0.85}\textbf{0.98\tiny$\pm$0.19} & 80.92\tiny$\pm$0.15 & 15.41\tiny$\pm$0.14 & 6.42\tiny$\pm$0.82 & \cellcolor[gray]{0.95}\textbf{59.33\tiny$\pm$5.58} & \cellcolor[gray]{0.85}\textbf{5.30\tiny$\pm$0.43} & 1.75\tiny$\pm$0.19 & 60.84\tiny$\pm$4.55 & 11.05\tiny$\pm$0.92 \\[3pt]
    
    19 & \cellcolor[gray]{0.85}\textbf{0.55\tiny$\pm$0.15} & 86.08\tiny$\pm$0.27 & 12.65\tiny$\pm$0.12 & 5.55\tiny$\pm$1.19 & \cellcolor[gray]{0.85}\textbf{78.99\tiny$\pm$4.96} & \cellcolor[gray]{0.85}\textbf{7.68\tiny$\pm$0.63} & 1.50\tiny$\pm$0.29 & 86.71\tiny$\pm$0.43 & 11.15\tiny$\pm$0.47 \\[3pt]
    
    20 & 0.98\tiny$\pm$0.15 & 91.94\tiny$\pm$0.23 & 6.23\tiny$\pm$0.09 & 3.03\tiny$\pm$0.34 & \cellcolor[gray]{0.85}\textbf{86.07\tiny$\pm$1.83} & \cellcolor[gray]{0.85}\textbf{3.03\tiny$\pm$0.70} & \cellcolor[gray]{0.85}\textbf{0.68\tiny$\pm$0.10} & 92.06\tiny$\pm$0.39 & 5.53\tiny$\pm$0.33 \\[3pt]
    
    21 & 1.91\tiny$\pm$0.38 & 88.61\tiny$\pm$0.18 & 9.83\tiny$\pm$0.14 & 5.11\tiny$\pm$0.98 & \cellcolor[gray]{0.85}\textbf{81.78\tiny$\pm$4.69} & \cellcolor[gray]{0.85}\textbf{3.49\tiny$\pm$0.52} & \cellcolor[gray]{0.95}\textbf{1.43\tiny$\pm$0.48} & 87.93\tiny$\pm$0.48 & 8.56\tiny$\pm$0.60 \\[3pt]
    
    22 & \cellcolor[gray]{0.85}\textbf{1.10\tiny$\pm$0.27} & 63.95\tiny$\pm$0.75 & 14.87\tiny$\pm$0.30 & 8.30\tiny$\pm$1.29 & 64.87\tiny$\pm$8.07 & \cellcolor[gray]{0.85}\textbf{4.72\tiny$\pm$0.81} & 2.04\tiny$\pm$0.83 & \cellcolor[gray]{0.95}\textbf{61.13\tiny$\pm$2.02} & 11.72\tiny$\pm$0.84 \\[3pt]
    
    23 & \cellcolor[gray]{0.95}\textbf{1.57\tiny$\pm$0.36} & \cellcolor[gray]{0.95}\textbf{24.30\tiny$\pm$1.74} & 10.96\tiny$\pm$0.99 & 11.41\tiny$\pm$1.37 & 43.39\tiny$\pm$7.84 & \cellcolor[gray]{0.85}\textbf{2.39\tiny$\pm$0.73} & 2.09\tiny$\pm$0.85 & 25.38\tiny$\pm$3.17 & 7.17\tiny$\pm$0.15 \\[3pt]
    
    24 & \cellcolor[gray]{0.95}\textbf{1.88\tiny$\pm$0.40} & 47.22\tiny$\pm$1.96 & 13.14\tiny$\pm$0.80 & 11.07\tiny$\pm$1.03 & 59.06\tiny$\pm$5.83 & \cellcolor[gray]{0.85}\textbf{3.53\tiny$\pm$1.01} & 2.38\tiny$\pm$0.68 & \cellcolor[gray]{0.95}\textbf{43.16\tiny$\pm$1.67} & 9.60\tiny$\pm$0.84 \\[3pt]
    
    25 & \cellcolor[gray]{0.95}\textbf{1.55\tiny$\pm$0.48} & 42.32\tiny$\pm$3.98 & 9.78\tiny$\pm$0.65 & 8.71\tiny$\pm$0.31 & 60.56\tiny$\pm$4.27 & \cellcolor[gray]{0.85}\textbf{2.58\tiny$\pm$1.21} & 1.65\tiny$\pm$0.35 & \cellcolor[gray]{0.95}\textbf{38.14\tiny$\pm$1.48} & 7.12\tiny$\pm$0.45 \\[3pt]
    
    \midrule
    
    \textbf{Average} & \cellcolor[gray]{0.95}\textbf{1.42\tiny$\pm$0.34} & 60.35\tiny$\pm$1.08 & 14.27\tiny$\pm$0.48 & 9.41\tiny$\pm$0.83 & 54.67\tiny$\pm$5.63 & \cellcolor[gray]{0.95}\textbf{5.74\tiny$\pm$1.71} & 2.08\tiny$\pm$0.73 & \cellcolor[gray]{0.95}\textbf{49.20\tiny$\pm$3.47} & 8.86\tiny$\pm$0.97 \\[3pt]

    \bottomrule
    \end{tabular}
    \end{adjustbox}
    \label{table:ece_unimodals}
\end{table*}

%% file: discussion_conclusion.tex
\section{Discussion}
\label{sec:discussion}

This study shows that aggregate performance metrics can overstate reliability by masking condition-specific calibration and selective prediction failures in multilabel clinical condition classification.
By moving beyond global averages, we reveal failure modes that are especially relevant in safety-critical settings, where errors affecting specific conditions or patient subpopulations may have disproportionate consequences.
These findings underscore the need for calibration-aware evaluation when selective prediction is used as a safeguard in multimodal clinical models.

\paragraph{Why Can Selective Prediction Fail Despite Improved Discrimination?}
Our results suggest that improvements in aggregate discrimination do not necessarily translate into reliable uncertainty estimates.
Although multimodal models can improve AUROC and AUPRC by combining complementary clinical data sources, their confidence estimates may remain poorly aligned with empirical correctness.
Since abstention policies often rely on confidence or predictive uncertainty, class-dependent miscalibration can cause models to retain harder or underrepresented cases while rejecting easier ones.
Thus, selective prediction can fail not because discrimination is poor overall, but because the confidence scores used for abstention are unreliable for specific labels or classes.

\paragraph{Potential Drivers of Class-Dependent Miscalibration.}
Although our experiments do not establish a causal mechanism by which multimodal fusion induces miscalibration, the results suggest plausible contributors.
Calibration failures are concentrated in the minority class and are most pronounced for lower-prevalence conditions, suggesting that label imbalance is a major driver.
Moreover, multimodal fusion may improve representation quality and ranking performance while still producing poorly calibrated class-conditional probabilities.
Simple mitigation strategies such as loss-upweighting can reduce positive-class calibration error, but may redistribute error toward the negative class, indicating that reweighting alone does not fully resolve class-dependent reliability failures.

\paragraph{Clinical Implications in Deployment.}
Minority-class overconfidence has important implications for clinical deployment.
Low-prevalence findings may correspond to rare but clinically important conditions, where confident false predictions can be especially harmful.
If a model is overconfident on these cases, selective prediction may fail to abstain precisely when additional human review is needed.
Consequently, strong aggregate performance may create a misleading impression of safety while reliability failures remain concentrated in clinically meaningful conditions.
For clinical AI systems intended to support triage, monitoring, or decision support, evaluation should therefore characterize both class-conditional calibration and selective prediction behavior across clinically relevant labels.

\paragraph{Limitations \& Future Work.}
This study has several limitations.
First, the analysis is conducted on a single ICU benchmark, and it remains open whether similar patterns hold across other datasets, modality combinations, or clinical tasks.
Second, ECE is sensitive to binning strategy: coarse bins may obscure localized miscalibration, while finer bins may increase estimator variance.
We mitigate this concern through robustness analyses using multiple ECE bin counts, Adaptive ECE, and Brier score, which support the same qualitative calibration--selective prediction misalignment.
Third, although loss-upweighting improves class-dependent calibration in some settings, it does not consistently improve selective prediction reliability across architectures.
Finally, our analysis is retrospective and in-distribution; future work should examine out-of-distribution clinical settings, prospective validation, and interactions with clinical workflows and human--AI decision-making.



\section{Conclusion}
\label{sec:conclusion}

This work provides an empirical characterization of the interaction between calibration and selective prediction in multimodal multilabel clinical condition classification.
We show that gains in discrimination from multimodal fusion do not reliably translate into improved selective prediction due to persistent class-dependent miscalibration.
These findings highlight the need for evaluation and training approaches that explicitly account for calibration when selective prediction is used as a safety mechanism.
More broadly, our results suggest that reliable deployment of multimodal clinical models requires moving beyond aggregate performance metrics toward task- and class-aware assessments of uncertainty.
Future work should develop calibration-aware selective prediction methods that remain reliable across labels, prevalence regimes, and clinical deployment settings.

%% file: appendix.tex
\appendix
\begin{appendices}

\section*{\LARGE Appendix}
\label{sec:appendix}

\setcounter{table}{0}
\setcounter{figure}{0}
\setcounter{equation}{0}
\renewcommand{\thetable}{\thesection.\arabic{table}}
\renewcommand{\thefigure}{\thesection.\arabic{figure}}
\renewcommand{\theequation}{\thesection.\arabic{equation}}

\vspace*{10pt}
\section*{Table of Contents}
\startcontents[sections]
\printcontents[sections]{l}{1}{\setcounter{tocdepth}{2}}

\section*{Reproducibility}

We have made a significant effort to ensure the reproducibility of our results.
An anonymized implementation of our method is provided at \href{https://github.com/jlaitue/medcertain}{https://github.com/jlaitue/medcertain}, which includes training, evaluation, and analysis scripts.
The experimental setup, including hyperparameters, model configurations, and sampling parameters are described in \Cref{app:training_details}.
All datasets used in our experiments are publicly available, and we additionally provide scripts for data preparation.
We provide model checkpoints for all models used in this paper upon request, with instructions for reproducing results in the \texttt{README.md} file in the released code.

\clearpage
\section{Clinical Condition Dataset Distribution}
\label{app:clinical_condition_distribution}
\input{tables/table_clinical_info}

\clearpage
\section{Training Details}
\label{app:training_details}

\subsection*{Hyperparameter Optimization and Model Selection}

We conducted 50 experiments corresponding to randomly sampled hyperparameter sets, each evaluated across five runs with different random seeds. 
The learning rate was sampled uniformly from $[10^{-5}, 10^{-2}]$, and the number of training epochs from $\{5, 10, 15, 20, 30\}$. 
We used a fixed batch size of 16 and a cosine decay scheduler with $\alpha = 0$. 
Model checkpoints were saved at the final training epoch, and the optimal hyperparameter set was selected based on the highest mean validation AUROC across seeds, yielding a total of 250 runs per task.

Using the selected configuration, we retrained the models by combining the training and validation splits and ran five seeds with different initializations to obtain the final models for test evaluation. 
This procedure was applied to unimodal CXR, EHR, and MedFuse models, following the original MedFuse protocol for model selection based on AUROC.

For DrFuse, we used the authors’ publicly released code and evaluation pipeline without modification, training and evaluating the model according to the specifications provided in their scripts.
For MeTra, we adapted the authors’ implementation originally designed for in-hospital mortality prediction by modifying only the final output layer to support clinical condition classification; all other training and optimization settings were kept consistent with the original implementation.

We report test performance as means and standard errors over five random seeds. 
All models were trained using the Adam optimizer.
Experiments were executed on NVIDIA A100 and V100 Tensor Core GPUs.

\subsection*{Data Availability} 

The MIMIC (Medical Information Mart for Intensive Care) dataset is publicly available for research purposes. 
Access to the data requires completion of a data use agreement, which ensures compliance with the Health Insurance Portability and Accountability Act (HIPAA). 
Researchers can request access through the PhysioNet platform at \url{https://physionet.org}, where detailed instructions and requirements are provided.
In our study, we used versions MIMIC-IV and MIMIC-CXR which include de-identified health data, vital signs, laboratory test results, medication records and chest X-ray images from patients admitted to the intensive care units at the Beth Israel Deaconess Medical Center.

\clearpage
\section{Extended Empirical Analysis}
\label{sec:further_empirical_results}

This section provides extended quantitative and qualitative evidence supporting the cross-architecture trends summarized in the main text.
We report detailed evaluation metrics, class-stratified calibration statistics, and full selective prediction curves for all architectures.

\subsection{Do Multimodal Models Improve or Degrade Selective Prediction Performance in This Task?}
\label{app:uni_multi_sel_pred}
\input{tables/table_eval_metrics_unimodals}

\subsection{Are Class-Dependent Calibration Effects Consistent Across Multimodal Architectures?}
\label{app:takeaway_2}

\Cref{table:new_eval_multimodals} reports aggregate discrimination and selective prediction metrics across all clinical conditions.
While MedFuse often achieves the strongest discrimination performance, no single multimodal architecture consistently attains lower calibration error across conditions.
Class-stratified ECE results in \Cref{table:new_ece_multimodals} further show that increased architectural complexity does not resolve class-dependent miscalibration, particularly for underrepresented positive outcomes.

Additional stratified calibration plots and selective prediction analyses for unimodal and multimodal models (EHR, CXR, MedFuse, DrFuse, and MeTRA) are provided in \Cref{sec:stratified_plots_models}, offering a complete view of per-condition behavior across model families.
Specifically, selective prediction curves for DrFuse and MeTRA are shown in \Cref{fig:drfuse_25_grid,fig:metra_25_grid}.
Across conditions, these curves closely mirror the MedFuse behavior observed in the main analysis, with moderate gains at intermediate rejection thresholds followed by degradation at extreme coverage levels.
This consistency indicates that the observed selective prediction failures are driven by shared calibration dynamics rather than architecture-specific fusion mechanisms.

\input{tables/table_eval_metrics_new_multimodals}

\input{tables/table_ece_new_multimodals}

\clearpage
\subsection{Can Loss-Upweighting Mitigate Observed Calibration Failures?}
\label{sec:takeaway_3}

This section provides extended quantitative evidence supporting the loss-upweighting results summarized in the main text.
We report per-condition class-stratified calibration statistics, and cross-architecture analyses to assess whether simple loss reweighting can mitigate calibration-driven selective prediction failures.



\paragraph{Cross-Architecture Consistency.}
\Cref{table:new_eval_multimodals_ups} reports the quantitative evaluation metrics for each multimodal architecture under loss upweighting, confirming that while selective AUROC/AUPRC exhibit modest gains, these improvements are inconsistent across conditions and do not fundamentally alter the observed reliability trends.

\paragraph{Calibration Effects.}
Using the standard aggregate ECE, we observe no meaningful global improvement under loss upweighting.
In contrast, class-stratified analysis reveals statistically significant reductions in positive-class ECE for 23 of 25 conditions, while negative-class calibration improves in 11 of 25.
These effects are quantified in \Cref{table:new_ece_multimodals_ups}, which reports class-stratified calibration metrics for each architecture under loss upweighting.
Together, these results confirm that loss reweighting reduces overconfidence for rare positive labels, but that these calibration gains do not reliably translate into improved selective prediction.

\input{tables/table_eval_metrics_upscaled_multimodals}

\input{tables/table_ece_upscaled_multimodals}

\clearpage
\subsection{Extended Calibration Metrics and Binning Strategies}
\label{sec:extended_calibration_metrics}

This section provides additional robustness analyses for the calibration results reported in the main text.
Since Expected Calibration Error (ECE) depends on the choice of confidence bins, we first evaluate whether the observed class-dependent miscalibration is sensitive to the number of bins used for estimation.
We then complement ECE with Adaptive Expected Calibration Error (AECE) and Brier score, which provide alternative views of calibration quality.
Across these analyses, the central trend remains consistent: calibration errors are substantially larger for the positive class than for the negative class, and this class-dependent miscalibration persists across unimodal and multimodal architectures.

\subsubsection{Variable Binning for ECE}

Binning-based ECE estimates involve an inherent bias--variance tradeoff.
Using fewer bins can reduce estimator variance but may obscure localized miscalibration, while using more bins reveals sharper calibration failures at the cost of increased variance, particularly for low-prevalence labels.
To assess whether our conclusions depend on a specific binning choice, we recompute aggregate and class-stratified ECE using $n \in \{5, 15, 20, 50\}$ bins (\Cref{table:ece_unimodals_binning}, \Cref{table:ece_multimodals_binning}).

Across bin counts, the same qualitative pattern is observed.
Although aggregate ECE changes moderately with the number of bins, positive-class calibration error remains substantially higher than negative-class calibration error across model families.
This indicates that the observed class-dependent miscalibration is not an artifact of a single ECE binning configuration.
\input{tables/table_ece_unimodals_binning}

\input{tables/table_ece_multimodals_binning}

\clearpage
\subsubsection{Adaptive Expected Calibration Error}

We further report Adaptive Expected Calibration Error (AECE), which uses adaptive bins with approximately equal sample counts rather than fixed-width confidence intervals.
This provides a complementary estimate of calibration error that is less directly tied to the distribution of confidence scores across fixed bin boundaries.
As with ECE, we report aggregate AECE as well as class-stratified AECE for the positive and negative classes for the unimodal, multimodal and loss-upweighted multimodal models.

Overall, the AECE results are consistent with the fixed-bin ECE analysis.
For standard unimodal and multimodal models (\Cref{table:aece_unimodals}, \Cref{table:aece_multimodals}), positive-class AECE remains substantially higher than negative-class AECE, confirming that calibration failures are concentrated in the minority class.
Under loss upweighting (\Cref{table:aece_multimodals_upweighted}), positive-class AECE decreases, but this improvement equally accompanied by a redistribution of calibration error toward the negative class.
Thus, while loss upweighting partially mitigates positive-class overconfidence, it does not eliminate class-dependent calibration imbalance.

\input{tables/table_aece_unimodals}

\input{tables/table_aece_new_multimodals}

\input{tables/table_aece_upscaled_multimodals}

\clearpage
\subsubsection{Brier Calibration Loss}

Finally, we report Brier score as an additional proper scoring rule for probabilistic predictions.
For binary labels, the Brier score measures the squared error between the predicted probability and the observed outcome.
We compute both aggregate Brier score and class-stratified Brier scores to evaluate whether the calibration trends observed with ECE and AECE persist under a metric that does not rely on binning.

The Brier score results reinforce the calibration trends observed with ECE and AECE.
For standard unimodal and multimodal models (\Cref{table:brier_unimodals}, \Cref{table:brier_multimodals}), the positive class exhibits substantially higher Brier loss than the negative class, indicating that errors in predicted confidence are concentrated among positive labels.
Loss upweighting reduces this gap by lowering positive-class Brier loss, but it also increases negative-class and aggregate Brier loss (\Cref{table:brier_multimodals_upweighted}).
Together, these results suggest that simple reweighting shifts calibration error across classes rather than fully resolving the class-dependent miscalibration that undermines selective prediction reliability.

\input{tables/table_brier_unimodals}

\input{tables/table_brier_new_multimodals}

\input{tables/table_brier_upscaled_multimodals}


\clearpage
\subsection{Stratified Calibration Behavior Across Modalities}
\label{sec:stratified_plots_models}
\begin{figure}[h!]
    \centering

    \includegraphics[width=\columnwidth]{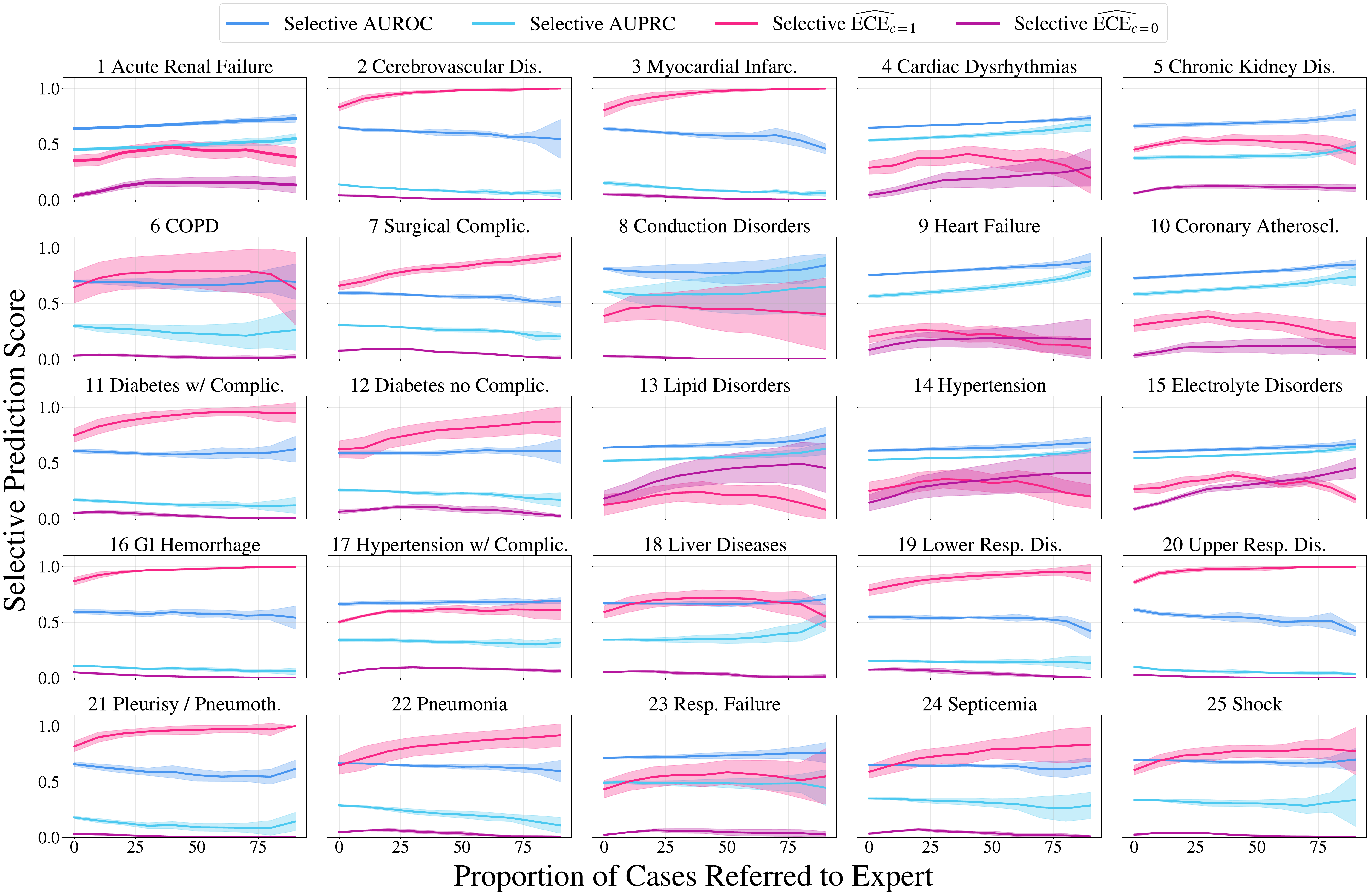}
    \vspace{1mm}

    \includegraphics[width=\columnwidth]{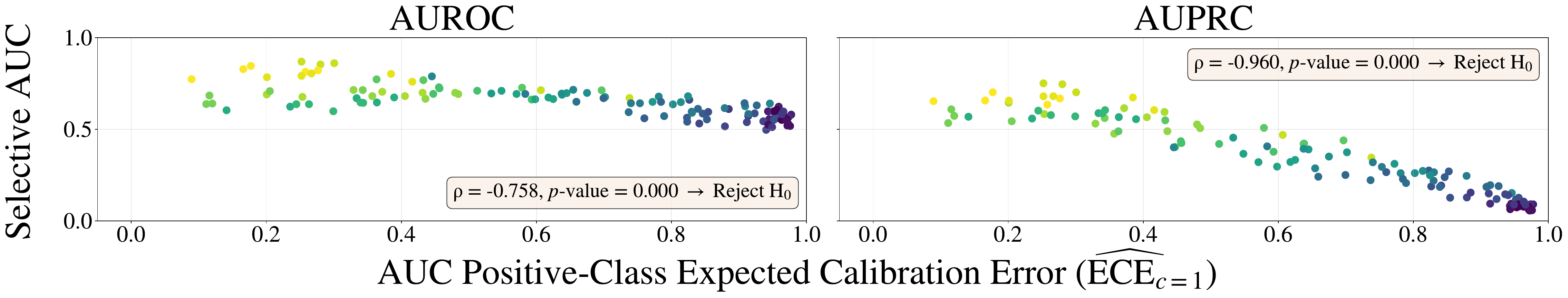}
    \vspace{0.5mm} 

    \includegraphics[width=\columnwidth]{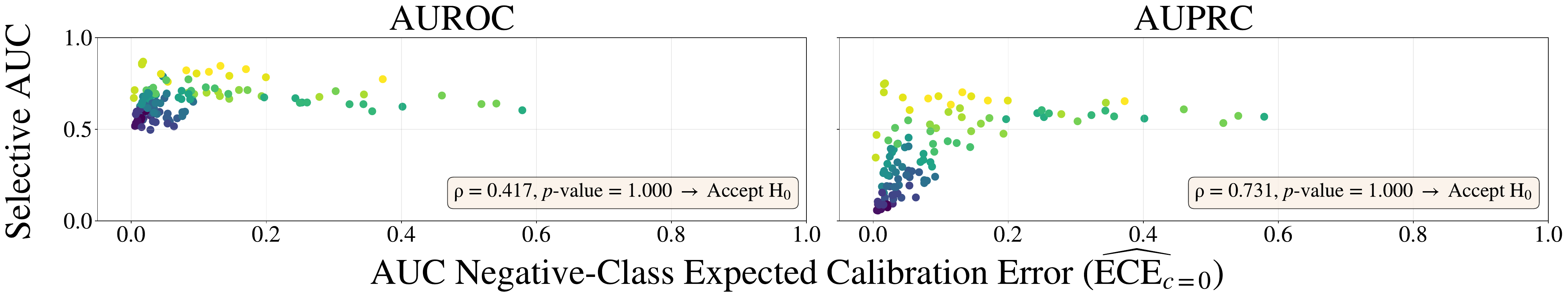}

    \caption{
    \textbf{Unimodal CXR Stratified Calibration and Selective Prediction Across Clinical Conditions.}
    For unimodal CXR, stratified positive-class ECE exhibits substantial variability across seeds and conditions (top), and AUROC/AUPRC remain below multimodal MedFuse performance.
    Across conditions, higher positive-class ECE AUC is associated with lower selective AUC (middle; Spearman’s rank $p<0.05$).
    Negative-class ECE AUC shows no clear monotonic relationship with selective AUC (bottom).
    }
    \label{fig:cxr_25_grid}
\end{figure}

\begin{figure}[h!]
    \centering

    \includegraphics[width=\columnwidth]{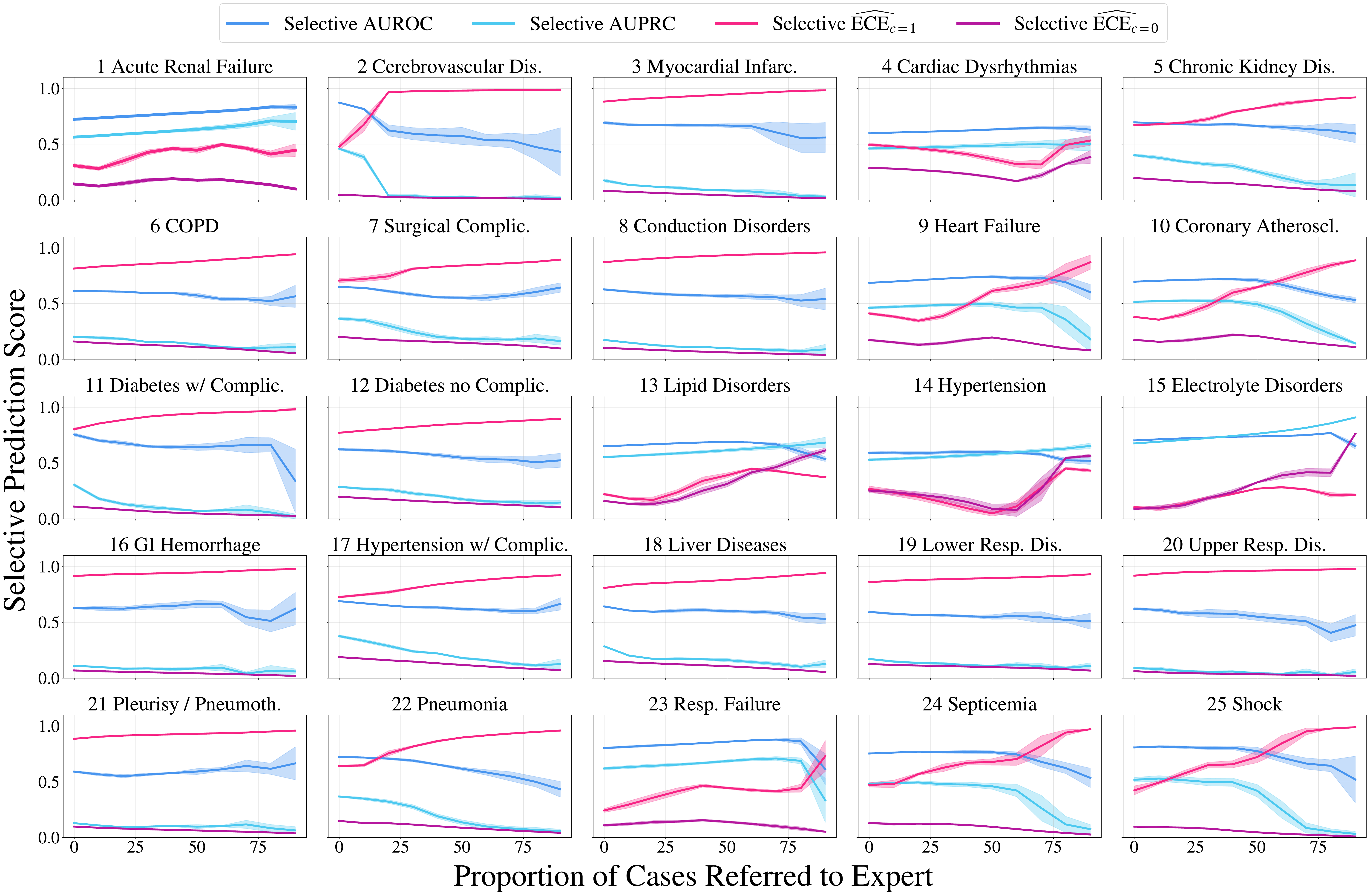}

    \includegraphics[width=\columnwidth]{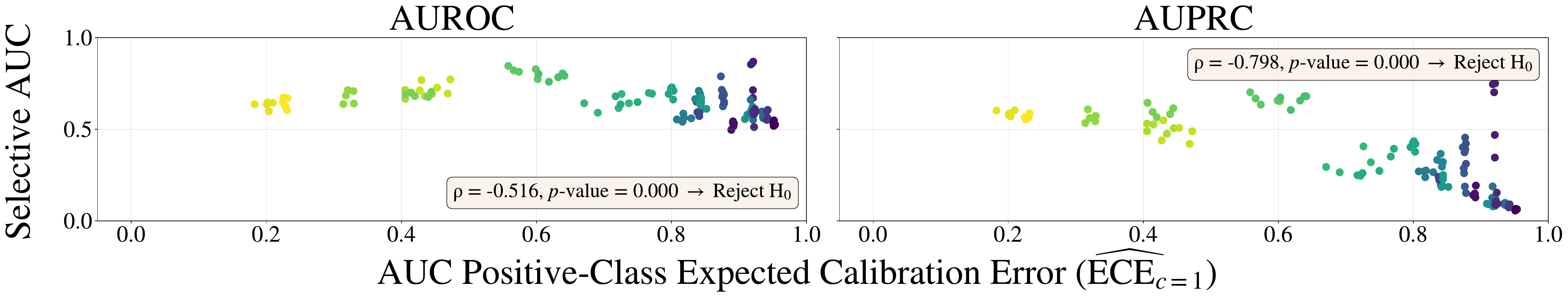}

    \includegraphics[width=\columnwidth]{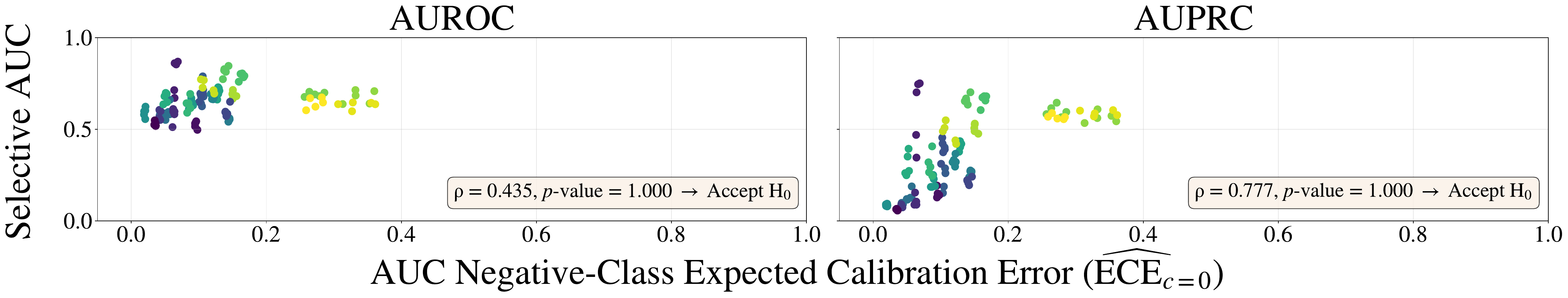}

    \caption{
   \textbf{Unimodal EHR Stratified Calibration and Selective Prediction Across Clinical Conditions.}
    Positive-class ECE is elevated for several conditions, indicating overconfidence on underrepresented positive cases.
    Compared to CXR, stratified ECE and AUROC/AUPRC show lower variability across seeds and conditions, yielding more stable predictions that closely mirror MedFuse behavior.
    Positive-class ECE AUC remains negatively associated with selective AUC, while negative-class ECE shows no clear monotonic relationship with selective performance.
    }
    \label{fig:ehr_25_grid}
\end{figure}

\begin{figure}[h!]
    \centering

    \includegraphics[width=\columnwidth]{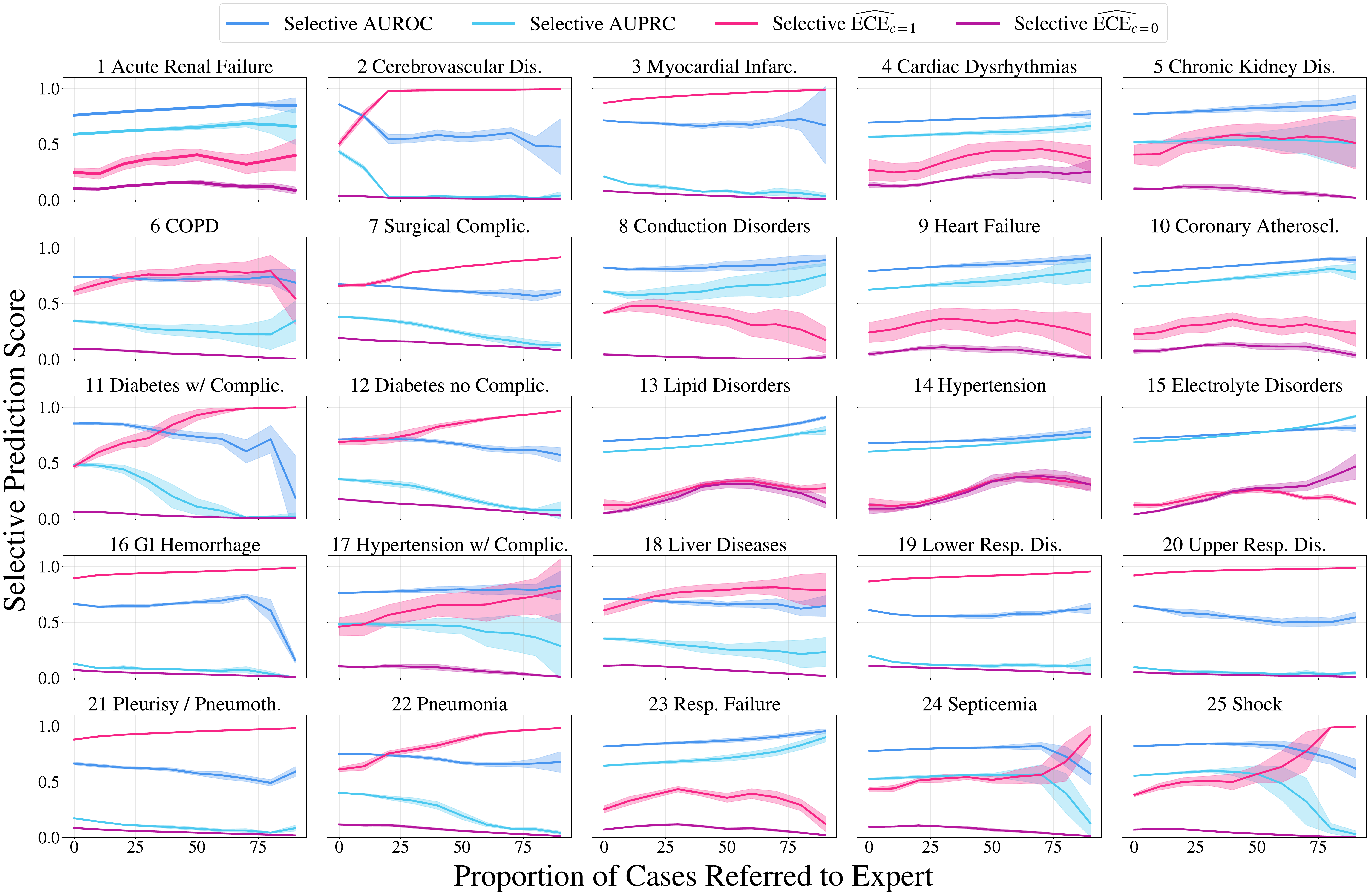}

    \includegraphics[width=\columnwidth]{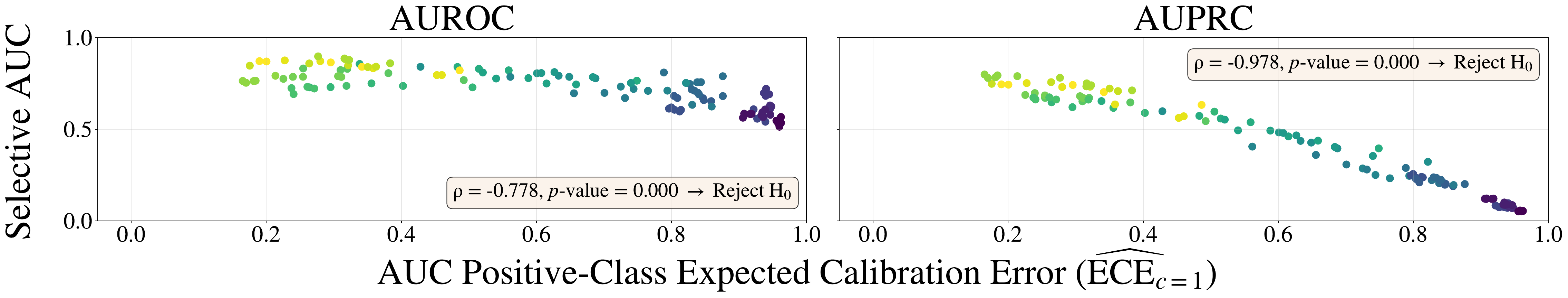}

    \includegraphics[width=\columnwidth]{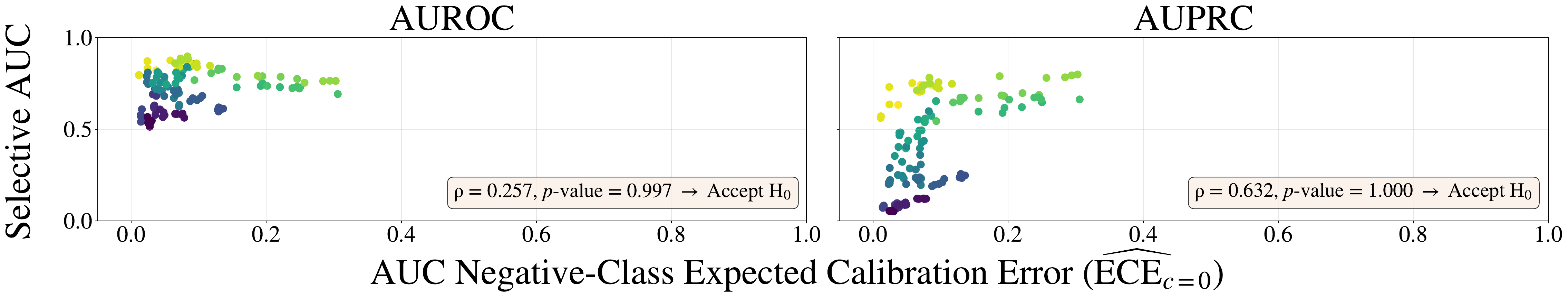}

    \caption{
    \textbf{MedFuse Stratified Calibration and Selective Prediction Across Clinical Conditions.}
    Across conditions, higher positive-class ECE AUC consistently predicts lower selective AUC, indicating that overconfidence in positive predictions drives instability in selective evaluation.
    In contrast, negative-class ECE AUC remains low for most conditions and shows no clear monotonic relationship with selective AUC, suggesting that miscalibration is primarily driven by overconfident positive predictions rather than negative-class errors.
    }
    \label{fig:medfuse_25_grid}
\end{figure}

To complement the results in \Cref{fig:det_class_level_ece}, we visualize class-stratified ECE and selective AUC for unimodal (CXR, EHR) and multimodal models (MedFuse, DrFuse, and MeTRA) across all 25 clinical conditions.  
These modality-level grids highlight how calibration dynamics evolve with selective rejection and emphasize that miscalibration, particularly in positive predictions, varies across architectures but follows the same underlying trend.
For the unimodal CXR model (\Cref{fig:cxr_25_grid}), positive-class ECE exhibits substantial variance across seeds, reflecting unstable confidence estimates despite selective AUC patterns that parallel those of MedFuse.  
The unimodal EHR model (\Cref{fig:ehr_25_grid}) shows stratified calibration curves closely aligned with the multimodal results from MedFuse (\Cref{fig:medfuse_25_grid}), though with consistently higher positive-class ECE for several low-prevalence conditions.  

\begin{figure}[h!]
    \centering

    \includegraphics[width=\columnwidth]{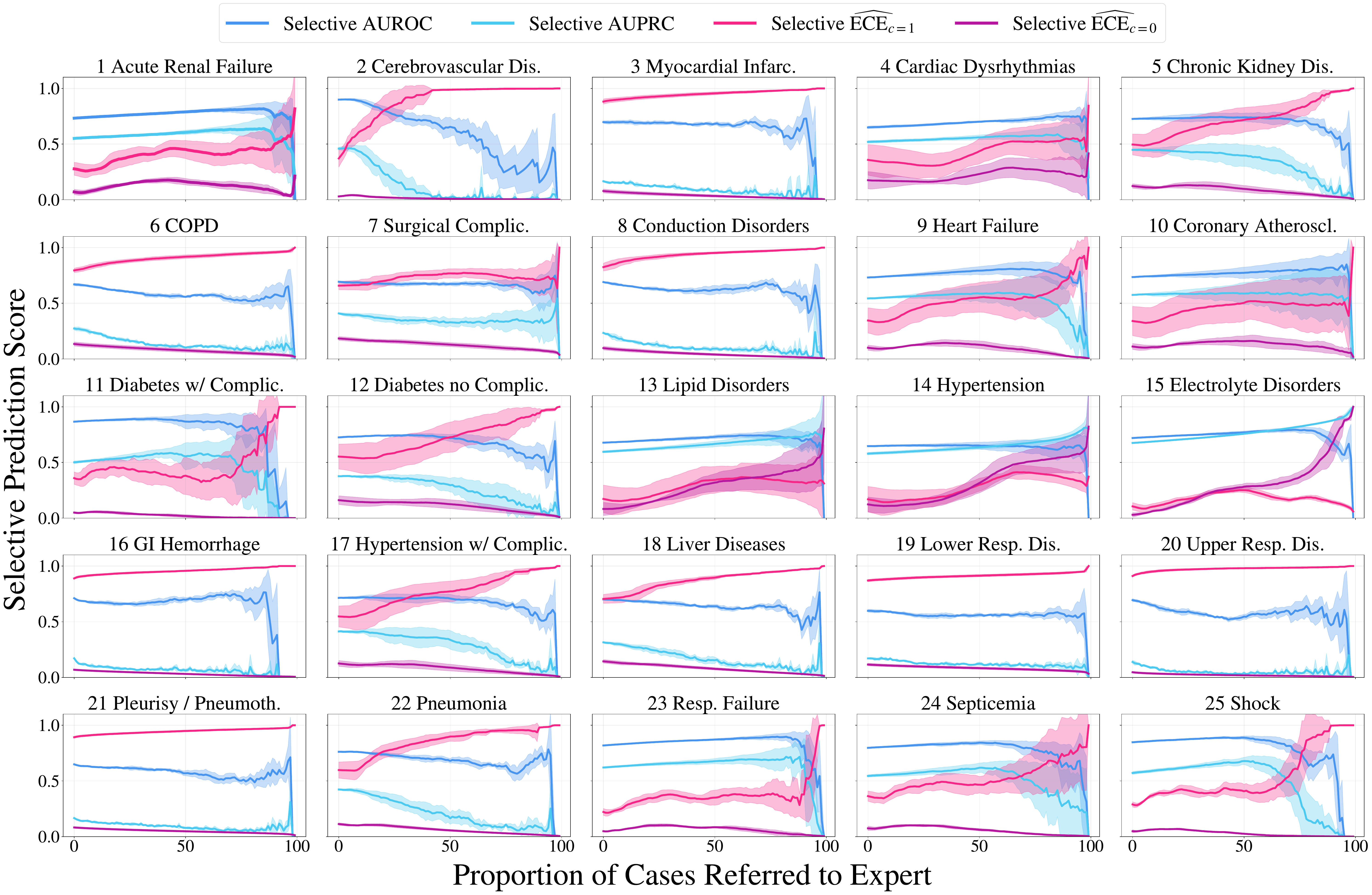}

    \includegraphics[width=\columnwidth]{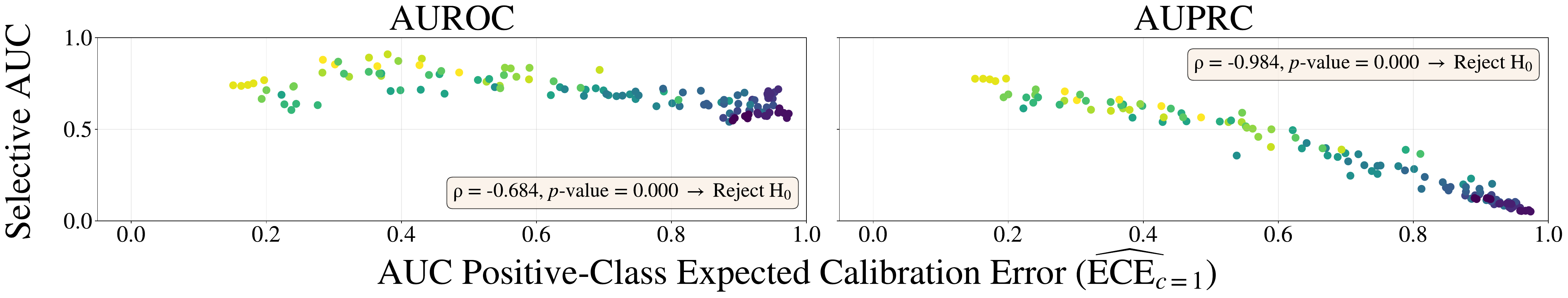}

    \includegraphics[width=\columnwidth]{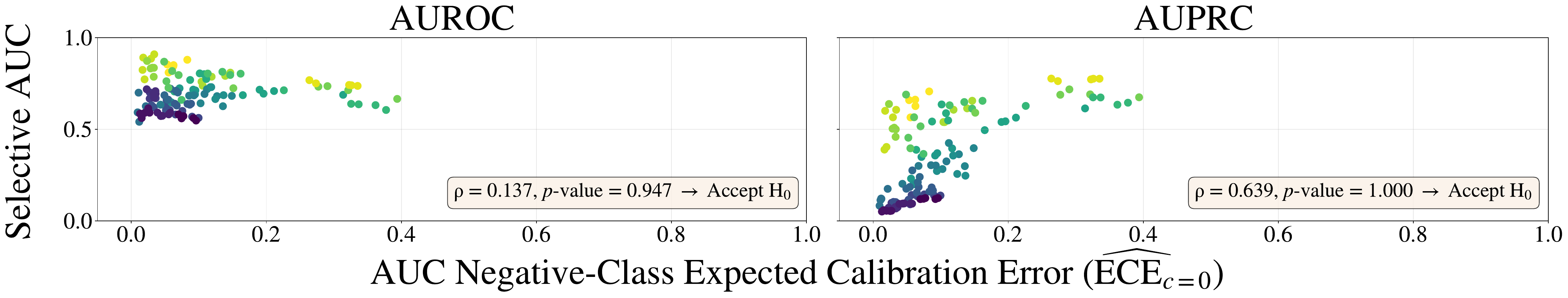}

    \caption{
    \textbf{DrFuse Stratified Calibration and Selective Prediction Across Clinical Conditions.}
    DrFuse exhibits elevated positive-class ECE across several conditions, indicating persistent overconfidence on underrepresented positive cases.
    Stratified ECE and selective AUROC/AUPRC closely mirror MedFuse trends, with higher positive-class ECE AUC associated with degraded selective performance.
    As in other models, negative-class ECE shows no clear monotonic relationship with selective AUC, suggesting that increased fusion complexity alone does not resolve calibration-driven reliability failures.
    }
    \label{fig:drfuse_25_grid}
\end{figure}

\begin{figure}[h!]
    \centering

    \includegraphics[width=\columnwidth]{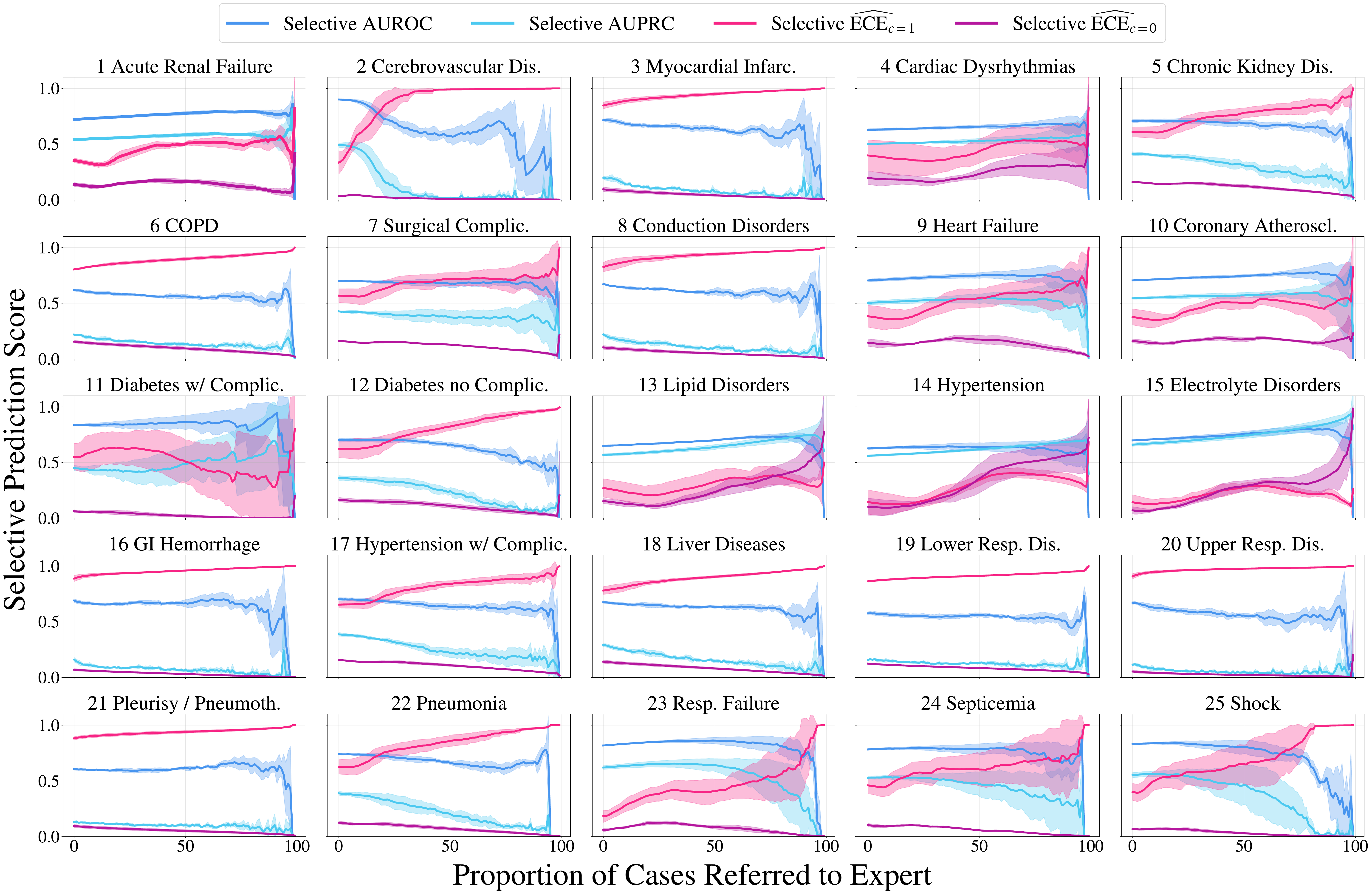}

    \includegraphics[width=\columnwidth]{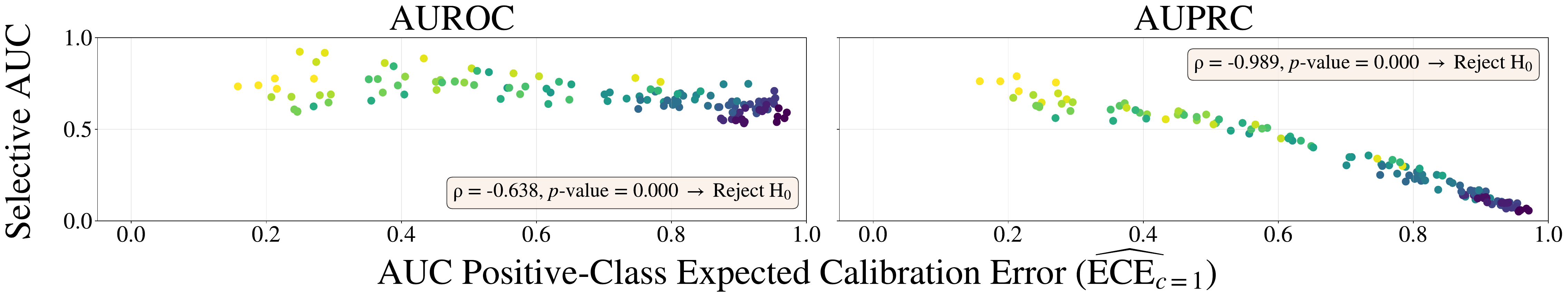}

    \includegraphics[width=\columnwidth]{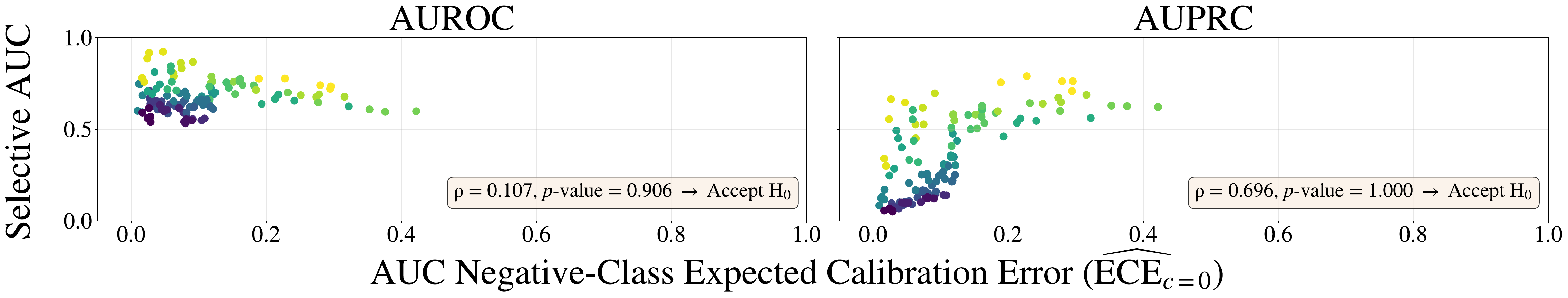}

    \caption{
    \textbf{MeTRA Stratified Calibration and Selective Prediction Across Clinical Conditions.}
    MeTRA shows class-dependent calibration behavior similar to MedFuse and DrFuse, with elevated positive-class ECE indicating overconfidence on underrepresented positives.
    Positive-class ECE AUC is negatively associated with selective AUC, while selective AUROC/AUPRC exhibit higher variance across seeds, consistent with other multimodal models, indicating that architectural complexity alone does not resolve calibration-driven selective instability.
    }
    \label{fig:metra_25_grid}
\end{figure}

Extending this analysis to DrFuse (\Cref{fig:drfuse_25_grid}) and MeTRA (\Cref{fig:metra_25_grid})  reveals broadly similar behaviors: both models produce the same selective and calibration trends observed in MedFuse, with only minor condition-specific deviations.  
These differences, however, do not translate into systematic performance or calibration gains, suggesting that architectural modifications, whether dynamic fusion or transformer-based attention, do not substantially alter the fundamental reliability dynamics across modalities.

\clearpage
\subsection*{Unimodal CXR}

\clearpage
\subsection*{Unimodal EHR}

\clearpage
\subsection*{MedFuse}

\clearpage
\subsection*{DrFuse}

\clearpage
\subsection*{MeTRA}

\clearpage
\section{Per-Condition Comparative Analysis}
\label{sec:appendix_evaluation_across_selected_condition}

\renewcommand{\thefigure}{D\arabic{figure}}
\setcounter{figure}{0}

\begin{figure*}[ht!]
    \centering
    \includegraphics[width=0.93\textwidth]{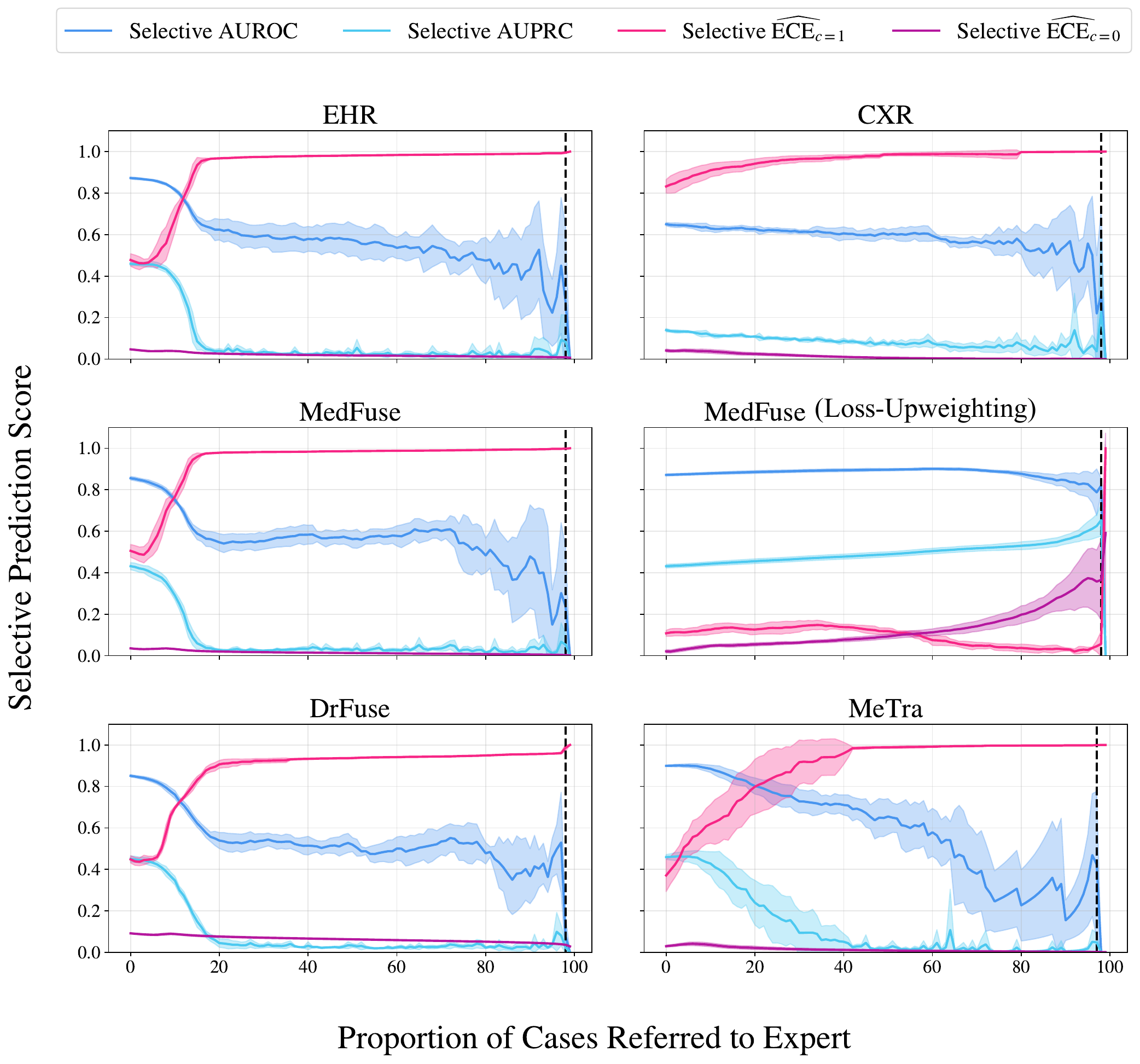}
    \caption{
    \textbf{Comparative analysis for Acute Cerebrovascular Disease.} 
    For this condition, EHR predominantly drives the discriminative performance of MedFuse, while CXR shows poor positive-class calibration and contributes to the oscillatory behavior of selective AUROC. 
    Applying label-weighted loss reduces early-threshold miscalibration in positive predictions but does not fully correct calibration at higher thresholds, leaving the overall selective prediction trend largely unchanged.
    }
    \label{fig:comparison_condition_2}
\end{figure*}
\begin{figure*}[ht!]
    \centering
    \includegraphics[width=\textwidth]{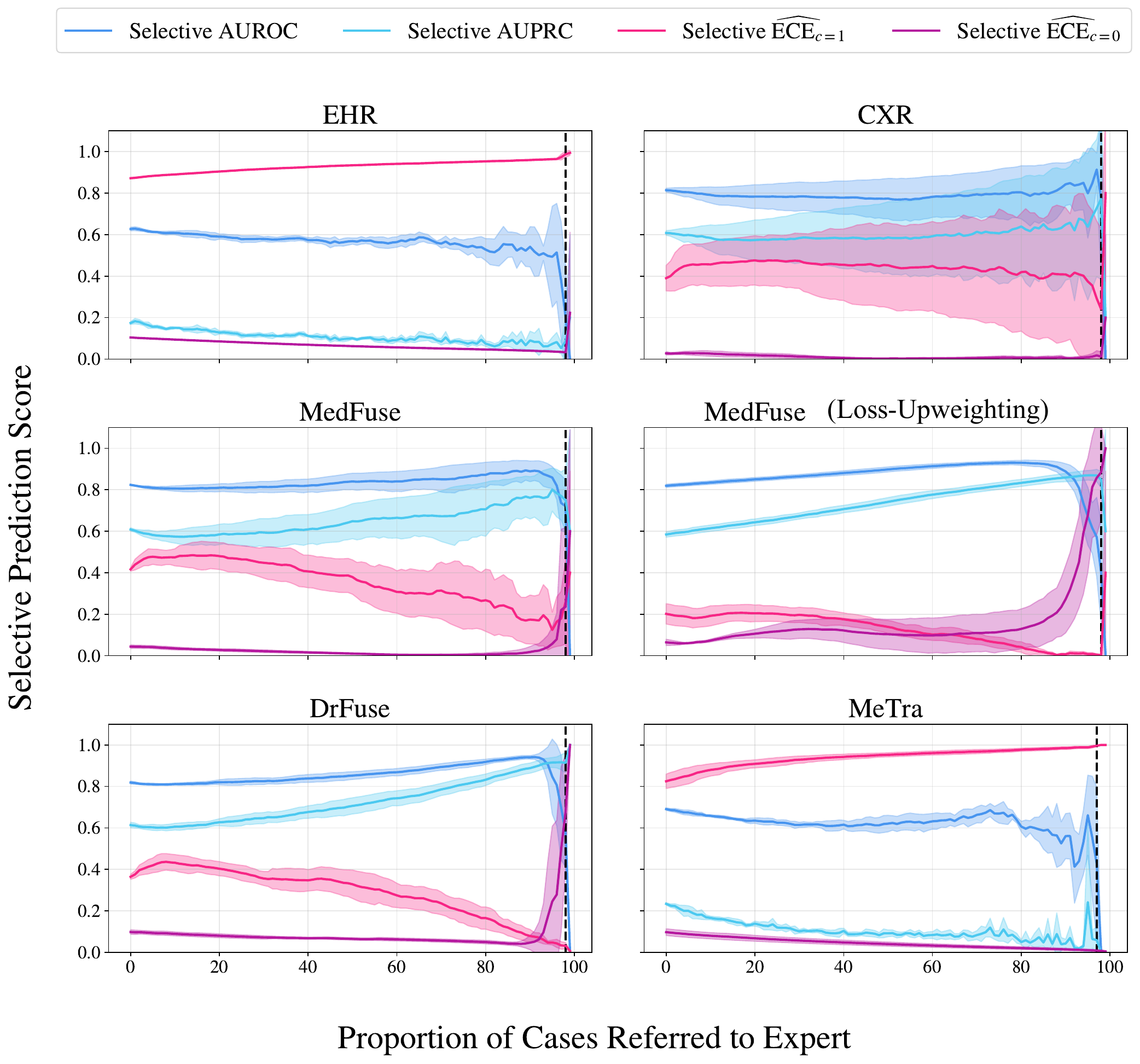}
    \caption{
    \textbf{Comparative analysis for Conduction Disorders.} 
    For this condition, the CXR model shows substantial seed-dependent variability, achieving higher mean performance but poorer calibration consistency than EHR. 
    MedFuse integrates both modalities to improve selective AUROC and stabilize performance, though positive-class ECE remains variable. 
    Label-weighted loss further reduces seed variance and attenuates positive-class ECE at higher confidence thresholds, yielding smoother and more reliable selective behavior, but with similar performance drops at the final thresholds.
    }
    \label{fig:comparison_condition_8}
\end{figure*}
\begin{figure*}[ht!]
    \centering
    \includegraphics[width=\textwidth]{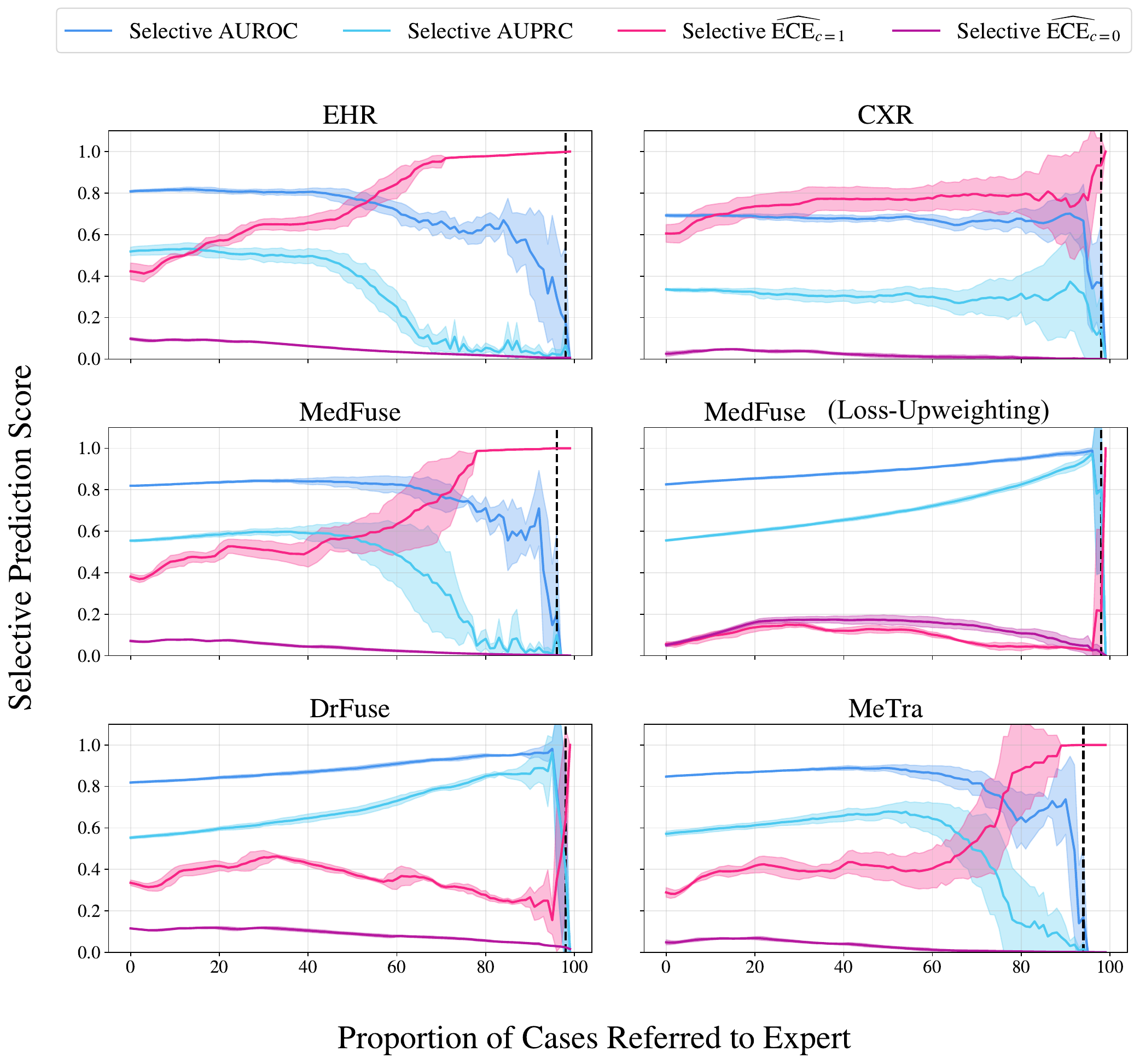}
    \caption{
    \textbf{Comparative analysis for Respiratory Shock.} 
    For this condition, unimodal EHR and CXR achieve comparable baseline performance, with EHR showing slightly lower positive-class ECE at early thresholds. 
    As rejection increases, however, EHR’s AUROC quickly deteriorates while CXR remains uniformly undercalibrated, and MedFuse largely inherits these calibration inconsistencies. 
    Label-weighted loss substantially improves both stability and calibration, suppressing positive-class ECE spikes and producing a clearer monotonic gain in selective AUROC. 
    This condition provides one of the strongest examples of how simple loss upweighting can directly reduce positive-class miscalibration and enhance the reliability of uncertainty estimates.
    }
    \label{fig:comparison_condition_25}
\end{figure*}

To provide qualitative insight into selective prediction behavior across models, we examine three representative clinical conditions: Acute Cerebrovascular Disease (2) (\Cref{fig:comparison_condition_2}), Conduction Disorders (8) (\Cref{fig:comparison_condition_8}), and Respiratory Shock (25) (\Cref{fig:comparison_condition_25}), across all model variants (EHR, CXR, MedFuse, MedFuse + Loss-Upweight, DrFuse and MeTra).
These examples highlight how calibration and selective prediction patterns vary across modalities and fusion strategies, and how simple loss upweighting affects positive-class calibration without fundamentally altering the selective trends observed across models.

\subsection{Label 2: Acute Cerebrovascular Disease}

\clearpage
\subsection{Label 8: Conduction Disorders}

\clearpage
\subsection{Label 25: Respiratory Shock}

\clearpage
\section{Regression Analysis of Calibration and Discriminative Performance}
\label{sec:regression_analysis}

\begin{figure*}[h!]
    \centering
    \includegraphics[width=\textwidth]{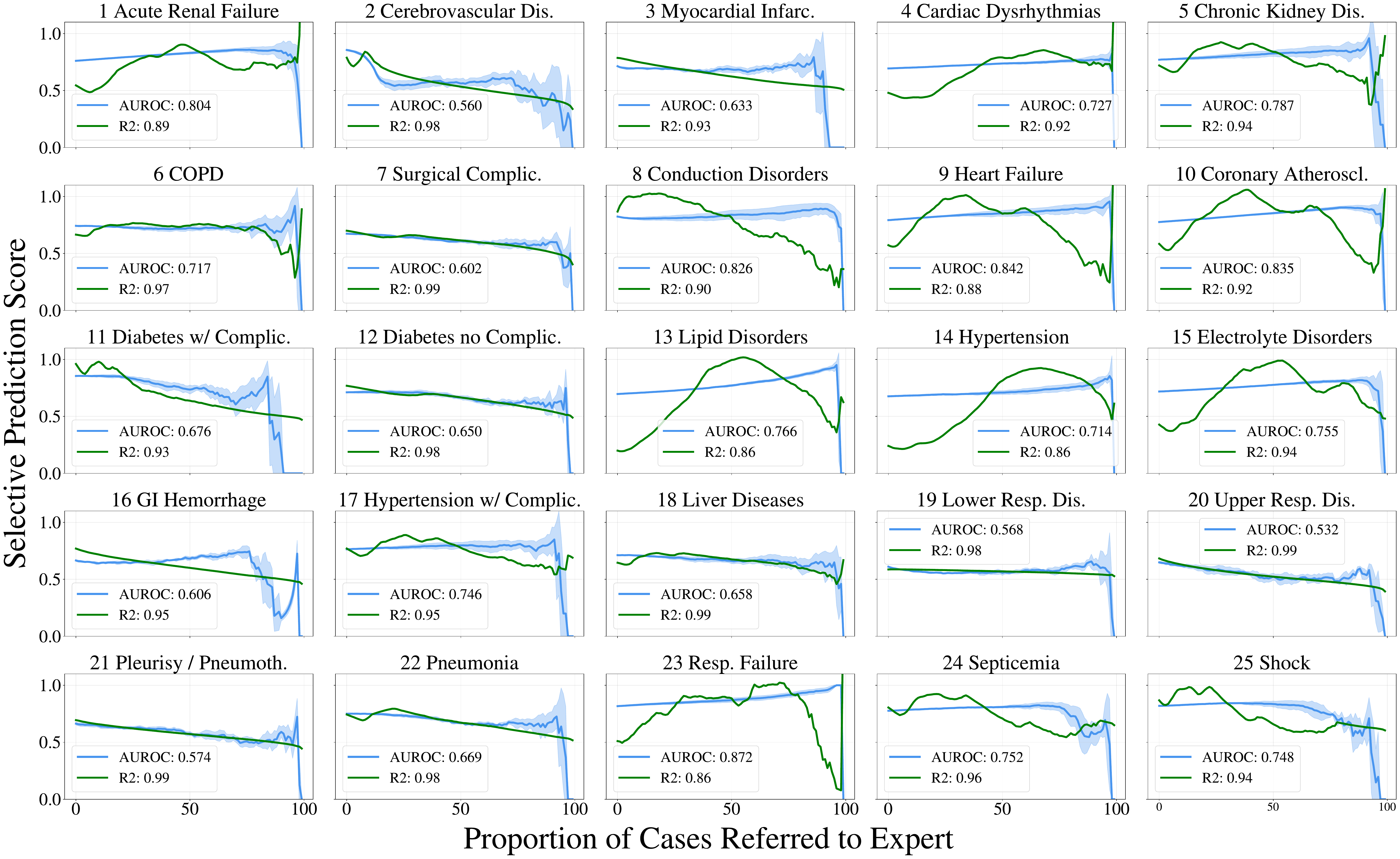}
    \caption{
    \textbf{Linear Regression of AUROC Against Stratified ECE.}
    Relationship between AUROC and both positive- and negative-class ECE across 25 conditions.  
    Regression slopes demonstrate a consistent, albeit not conclusive, negative association, indicating that poorer calibration modestly reduces discriminative performance in terms of AUROC.}
    \label{fig:regression_auroc}
\end{figure*}

\begin{figure*}[h!]
    \centering
    \includegraphics[width=\textwidth]{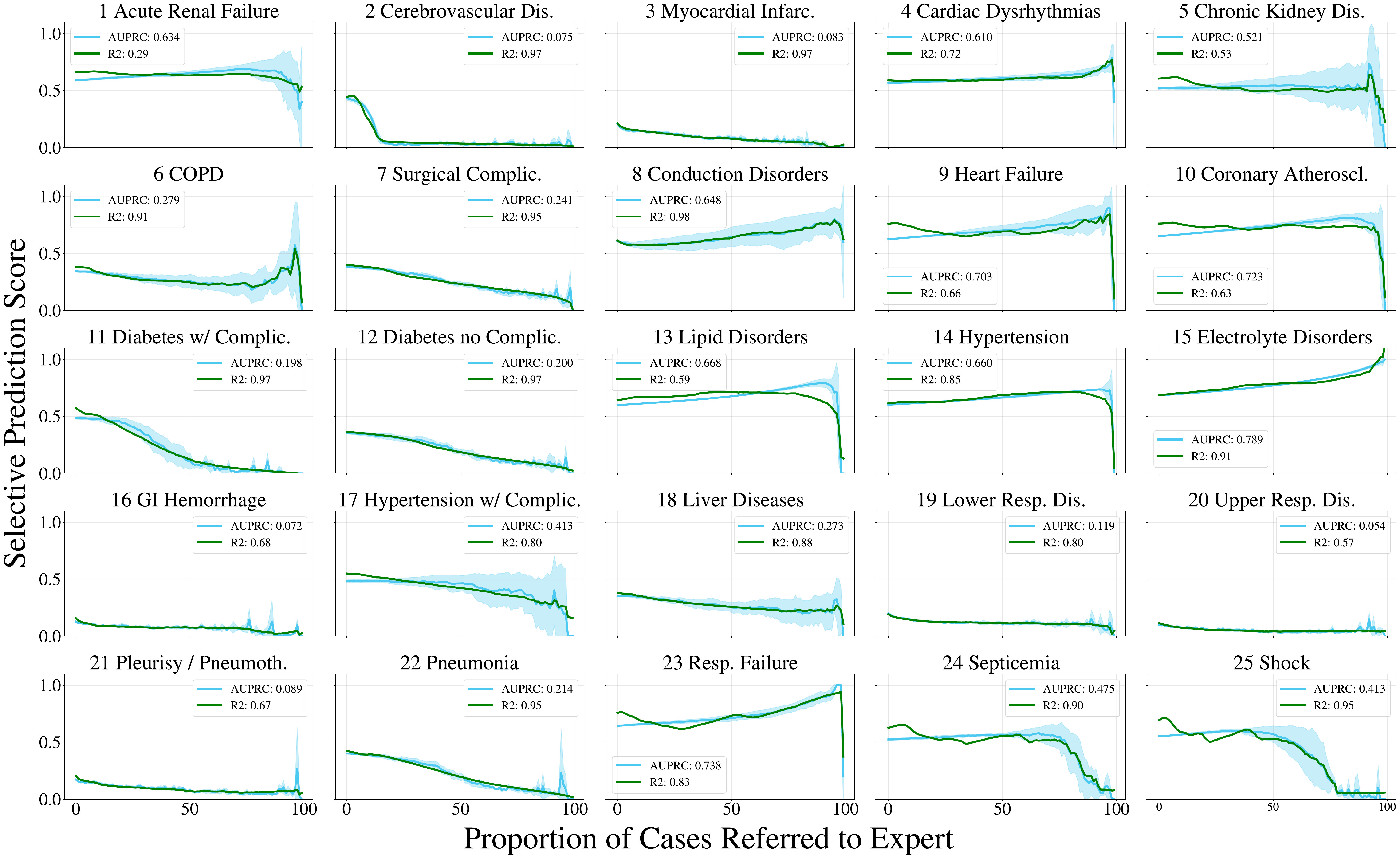}
    \caption{
    \textbf{Linear Regression of AUPRC Against Stratified ECE.}
    Relationship between AUPRC and class-stratified ECE across 25 conditions.  
    The correlation between ECE and AUPRC exhibits a clear, monotonic decline, confirming via another approach, that high positive-class miscalibration strongly predicts lower precision-recall performance under selective prediction.}
    \label{fig:regression_auprc}
\end{figure*}

To strengthen the link between calibration and selective performance, we fit linear regression models on MedFuse predictions across all 25 clinical conditions using class-stratified ECE (positive and negative) as predictors of both AUROC and AUPRC.   
Results reveal that stratified ECE, particularly the positive-class component, exhibits a strong negative correlation with AUPRC, reinforcing that miscalibrated confidence in positive predictions reliably forecasts degraded precision and recall under selective evaluation.
We plot the details of the regression curves for all 25 conditions using two independent predictors for ECE: AUROC (\Cref{fig:regression_auroc}) and AUPRC (\Cref{fig:regression_auprc}).

\clearpage

\end{appendices}

%% file: tables/table_clinical_info.tex
\begin{table*}[h!]

\caption{\textbf{Clinical Conditions and Dataset Characteristics.} 
Overview of the 25 conditions analyzed in this study, including their associated organ/system, primary diagnostic modality (CXR or EHR), and the prevalence of positive and negative samples in the training set.}
\vspace*{10pt}
    \centering
    \resizebox{\linewidth}{!}{
    \begin{tabular}{ll|cc|cc}
    \multirow{2}{*}[-1em]{\textbf{Clinical Condition}} &
    \multirow{2}{*}[-1em]{\textbf{Organ}} &
      \multicolumn{2}{c}{\textbf{Modality}} &
      \multicolumn{2}{c}{\textbf{Train Set Prevalence}} \\
      \midrule
      & {} & \textbf{CXR} & \textbf{EHR} 
      & \textbf{Positive} & \textbf{Negative} \\
      \midrule

     1 Acute and unspecified renal failure & Kidneys & - & \checkmark & 32.32\% (2,532) & 67.68\% (5,309) \\[3pt]

    2 Acute cerebrovascular disease & Brain (Cerebrovascular system) & - & \checkmark & 7.99\% (625) & 92.01\% (7,216) \\[3pt]
    
    3 Acute myocardial infarction & Heart (Coronary arteries) & - & \checkmark & 8.57\% (675) & 91.43\% (7,166) \\[3pt]
    
    4 Cardiac dysrhythmias & Heart (Conduction system) & - & \checkmark &	37.36\% (2,942)	& 62.64\% (4,899) \\[3pt]
    
    5 Chronic kidney disease & Kidneys & \checkmark & \checkmark & 23.34\% (1,824) & 76.66\% (6,017) \\[3pt]
    
    6 Chronic obstructive pulmonary disease & Lungs (Airways) & \checkmark & \checkmark & 16.40\% (1,269) & 83.60\% (6,572) \\[3pt]
    
    7 Complications of surgical procedures & Multiple systems (context-specific) & - & \checkmark & 21.75\% (1,703) &	78.25\% (6,138)	\\[3pt]
    
    8 Conduction disorders & Heart (Electrical system) & \checkmark & \checkmark & 10.79\% (838) & 89.21\% (7,003) \\[3pt]
    
    9 Congestive heart failure; nonhypertensive & Heart	& - & \checkmark & 31.03\% (2,451) & 68.97\% (5,390) \\[3pt]
    
    10 Coronary atherosclerosis and other heart disease & Heart (Coronary arteries)	& - & \checkmark & 31.00\% (2,452) &	69.00\% (5,389)	\\[3pt]
    
    11 Diabetes mellitus with complications & Endocrine system (Pancreas) + Target organs & - & \checkmark & 11.82\% (909) &	88.18\% (6,932) \\[3pt]
    
    12 Diabetes mellitus without complication & Endocrine system (Pancreas) & - & \checkmark & 20.58\% (1,634) & 79.42\% (6,207) \\[3pt]
    
    13 Disorders of lipid metabolism & Liver and Circulatory system & - & \checkmark & 41.05\% (3,207) & 58.95\% (4,634) \\[3pt]
    
    14 Essential hypertension & Cardiovascular system & - & \checkmark &	44.14\% (3,467)	& 55.86\% (4,374) \\[3pt]
    
    15 Fluid and electrolyte disorders & Kidneys and Endocrine system & - & \checkmark & 45.48\% (3,547) & 54.52\% (4,294) \\[3pt]
    
    16 Gastrointestinal hemorrhage & GI Tract & - & \checkmark & 7.08\% (562) & 92.92\% (7,279) \\[3pt]
    
    17 Hypertension with complications & Cardiovascular system	& - & \checkmark & 21.17\% (1,649) & 78.83\% (6,192) \\[3pt]
    
    18 Other liver diseases & Liver & \checkmark & \checkmark & 16.12\% (1,248)	& 83.88\% (6,593) \\[3pt]
    
    19 Other lower respiratory disease & Lungs (Lower airways) & \checkmark & \checkmark & 12.93\% (1,001) & 87.07\% (6,840)	\\[3pt]
    
    20 Other upper respiratory disease & Lungs/Nasal passages (Upper airways) & \checkmark & \checkmark & 6.29\% (500)	& 93.71\% (7,341) \\[3pt]
    
    21 Pleurisy; pneumothorax & Lungs/Pleura & \checkmark & \checkmark & 9.93\% (784) & 90.07\% (7,057)\\[3pt]
    
    22 Pneumonia (except caused by tuberculosis or std) & Lungs (Alveoli) & \checkmark & \checkmark & 18.84\% (1,489) & 81.16\% (6,352) \\[3pt]
    
    23 Respiratory failure & Lungs and/or Neuromuscular system	& - & \checkmark & 28.37\% (2,229) & 71.63\% (5,612) \\[3pt]
    
    24 Septicemia (except in labor) & Bloodstream (Systemic)	& \checkmark & \checkmark & 22.21\% (1,732) & 77.79\% (6,109) \\[3pt]
    
    25 Shock & Cardiovascular system (Systemic)	& - & \checkmark & 18.01\% (1,413) & 81.99\% (6,428) \\[3pt]

    \bottomrule
    \end{tabular}
    }
    \label{table:clinical_info}
\end{table*}

%% file: tables/table_eval_metrics_unimodals.tex
\begin{table*}[h!]
\caption{\textbf{Do Multimodal Models Improve or Degrade Selective Prediction Performance in this Task?} Condition-level performance of standard and selective prediction metrics.
MedFuse generally improves AUROC, AUPRC and selective AUROC, but calibration gains are very inconsistent compared to EHR.
\footnotesize{(Dark-bold: $p<0.05$, Wilcoxon signed-rank test, 5 seeds; Light-bold: highest mean, not significant)}}
\vspace*{3pt}
    \centering
    \resizebox{\linewidth}{!}{
    \begin{tabular}{c|ccccc|ccccc|cccccc}
     \multirow{2}{*}[-1.75em]{\textbf{\makecell{Clinical \\ Condition}}} 
    & \multicolumn{5}{c}{\textbf{EHR}}
    & \multicolumn{5}{c}{\textbf{CXR}}
    & \multicolumn{5}{c}{\textbf{MedFuse}}\\
    \midrule

    & \textbf{AUROC} 
    & \textbf{AUPRC} 
    & \textbf{\makecell{Selective \\ AUROC}}
    & \textbf{\makecell{Selective \\ AUPRC}}
    & \textbf{$\widehat{\mathbf{ECE}} \downarrow$} 

    & \textbf{AUROC} 
    & \textbf{AUPRC} 
    & \textbf{\makecell{Selective \\ AUROC}}
    & \textbf{\makecell{Selective \\ AUPRC}}
     & \textbf{$\widehat{\mathbf{ECE}} \downarrow$}

    & \textbf{AUROC} 
    & \textbf{AUPRC} 
    & \textbf{\makecell{Selective \\ AUROC}}
    & \textbf{\makecell{Selective \\ AUPRC}}
     & \textbf{$\widehat{\mathbf{ECE}} \downarrow$} \\
    \midrule

    1 & 0.724\tiny$\pm$0.003 & 0.564\tiny$\pm$0.010 & 0.772\tiny$\pm$0.100 & 0.627\tiny$\pm$0.100 & \cellcolor[gray]{0.95}\textbf{1.62\tiny$\pm$0.52} & 0.639\tiny$\pm$0.009 & 0.453\tiny$\pm$0.014 & 0.684\tiny$\pm$0.081 & 0.503\tiny$\pm$0.075 & 12.77\tiny$\pm$0.98 & \cellcolor[gray]{0.85}\textbf{0.761\tiny$\pm$0.005} & \cellcolor[gray]{0.85}\textbf{0.589\tiny$\pm$0.008} & \cellcolor[gray]{0.95}\textbf{0.804\tiny$\pm$0.111} & \cellcolor[gray]{0.95}\textbf{0.634\tiny$\pm$0.109} & 2.31\tiny$\pm$0.97 \\[3pt]

    2 & \cellcolor[gray]{0.85}\textbf{0.873\tiny$\pm$0.005} & \cellcolor[gray]{0.85}\textbf{0.459\tiny$\pm$0.010} & 0.569\tiny$\pm$0.180 & \cellcolor[gray]{0.95}\textbf{0.085\tiny$\pm$0.138} & 1.08\tiny$\pm$0.14 & 0.650\tiny$\pm$0.009 & 0.139\tiny$\pm$0.009 & \cellcolor[gray]{0.95}\textbf{0.577\tiny$\pm$0.121} & 0.084\tiny$\pm$0.057 & 3.99\tiny$\pm$0.31 & 0.856\tiny$\pm$0.010 & 0.432\tiny$\pm$0.019 & 0.560\tiny$\pm$0.176 & 0.075\tiny$\pm$0.116 & \cellcolor[gray]{0.95}\textbf{0.88\tiny$\pm$0.24} \\[3pt]
    
    3 & 0.693\tiny$\pm$0.014 & 0.175\tiny$\pm$0.016 & 0.617\tiny$\pm$0.150 & 0.089\tiny$\pm$0.070 & \cellcolor[gray]{0.95}\textbf{0.73\tiny$\pm$0.07} & 0.640\tiny$\pm$0.015 & 0.153\tiny$\pm$0.017 & 0.567\tiny$\pm$0.102 & \cellcolor[gray]{0.95}\textbf{0.092\tiny$\pm$0.037} & 4.35\tiny$\pm$0.42 & \cellcolor[gray]{0.95}\textbf{0.713\tiny$\pm$0.005} & \cellcolor[gray]{0.85}\textbf{0.209\tiny$\pm$0.010} & \cellcolor[gray]{0.95}\textbf{0.633\tiny$\pm$0.207} & 0.083\tiny$\pm$0.049 & 0.76\tiny$\pm$0.23 \\[3pt]
    
    4 & 0.598\tiny$\pm$0.004 & 0.462\tiny$\pm$0.012 & 0.618\tiny$\pm$0.069 & 0.491\tiny$\pm$0.065 & \cellcolor[gray]{0.85}\textbf{1.15\tiny$\pm$0.67} & 0.647\tiny$\pm$0.009 & 0.534\tiny$\pm$0.013 & 0.683\tiny$\pm$0.077 & 0.598\tiny$\pm$0.077 & 12.50\tiny$\pm$0.48 & \cellcolor[gray]{0.85}\textbf{0.694\tiny$\pm$0.006} & \cellcolor[gray]{0.85}\textbf{0.565\tiny$\pm$0.010} & \cellcolor[gray]{0.85}\textbf{0.727\tiny$\pm$0.079} & \cellcolor[gray]{0.95}\textbf{0.610\tiny$\pm$0.071} & 2.61\tiny$\pm$1.65 \\[3pt]
    
    5 & 0.696\tiny$\pm$0.006 & 0.402\tiny$\pm$0.012 & 0.642\tiny$\pm$0.113 & 0.244\tiny$\pm$0.113 & \cellcolor[gray]{0.95}\textbf{2.13\tiny$\pm$0.45} & 0.662\tiny$\pm$0.021 & 0.378\tiny$\pm$0.019 & 0.696\tiny$\pm$0.085 & 0.409\tiny$\pm$0.078 & 15.37\tiny$\pm$0.37 & \cellcolor[gray]{0.85}\textbf{0.770\tiny$\pm$0.003} & \cellcolor[gray]{0.85}\textbf{0.519\tiny$\pm$0.010} & \cellcolor[gray]{0.85}\textbf{0.787\tiny$\pm$0.177} & \cellcolor[gray]{0.95}\textbf{0.521\tiny$\pm$0.169} & 2.34\tiny$\pm$2.04 \\[3pt]
    
    6 & 0.612\tiny$\pm$0.005 & 0.203\tiny$\pm$0.008 & 0.564\tiny$\pm$0.089 & 0.140\tiny$\pm$0.045 & \cellcolor[gray]{0.95}\textbf{1.87\tiny$\pm$0.23} & 0.702\tiny$\pm$0.012 & 0.301\tiny$\pm$0.016 & 0.676\tiny$\pm$0.130 & 0.255\tiny$\pm$0.139 & 7.15\tiny$\pm$1.26 & \cellcolor[gray]{0.85}\textbf{0.742\tiny$\pm$0.004} & \cellcolor[gray]{0.85}\textbf{0.345\tiny$\pm$0.011} & \cellcolor[gray]{0.95}\textbf{0.717\tiny$\pm$0.120} & \cellcolor[gray]{0.95}\textbf{0.279\tiny$\pm$0.127} & 2.17\tiny$\pm$0.45 \\[3pt]
    
    7 & 0.649\tiny$\pm$0.007 & 0.365\tiny$\pm$0.013 & 0.578\tiny$\pm$0.098 & 0.221\tiny$\pm$0.083 & \cellcolor[gray]{0.95}\textbf{1.16\tiny$\pm$0.45} & 0.597\tiny$\pm$0.015 & 0.308\tiny$\pm$0.009 & 0.555\tiny$\pm$0.069 & \cellcolor[gray]{0.95}\textbf{0.263\tiny$\pm$0.067} & 9.15\tiny$\pm$0.58 & \cellcolor[gray]{0.85}\textbf{0.673\tiny$\pm$0.001} & \cellcolor[gray]{0.85}\textbf{0.383\tiny$\pm$0.003} & \cellcolor[gray]{0.85}\textbf{0.602\tiny$\pm$0.094} & 0.241\tiny$\pm$0.100 & 1.46\tiny$\pm$0.31 \\[3pt]
    
    8 & 0.627\tiny$\pm$0.010 & 0.174\tiny$\pm$0.007 & 0.558\tiny$\pm$0.098 & 0.109\tiny$\pm$0.054 & \cellcolor[gray]{0.85}\textbf{1.09\tiny$\pm$0.19} & 0.814\tiny$\pm$0.010 & 0.608\tiny$\pm$0.013 & 0.784\tiny$\pm$0.130 & 0.597\tiny$\pm$0.198 & 3.42\tiny$\pm$0.45 & \cellcolor[gray]{0.95}\textbf{0.824\tiny$\pm$0.004} & \cellcolor[gray]{0.95}\textbf{0.609\tiny$\pm$0.009} & \cellcolor[gray]{0.95}\textbf{0.826\tiny$\pm$0.099} & \cellcolor[gray]{0.95}\textbf{0.648\tiny$\pm$0.119} & 2.32\tiny$\pm$0.45 \\[3pt]
    
    9 & 0.686\tiny$\pm$0.005 & 0.463\tiny$\pm$0.011 & 0.690\tiny$\pm$0.113 & 0.419\tiny$\pm$0.128 & \cellcolor[gray]{0.85}\textbf{1.81\tiny$\pm$0.56} & 0.755\tiny$\pm$0.008 & 0.565\tiny$\pm$0.017 & 0.808\tiny$\pm$0.097 & 0.658\tiny$\pm$0.100 & 11.92\tiny$\pm$1.95 & \cellcolor[gray]{0.85}\textbf{0.792\tiny$\pm$0.004} & \cellcolor[gray]{0.85}\textbf{0.625\tiny$\pm$0.008} & \cellcolor[gray]{0.95}\textbf{0.842\tiny$\pm$0.110} & \cellcolor[gray]{0.95}\textbf{0.703\tiny$\pm$0.123} & 4.50\tiny$\pm$1.71 \\[3pt]
    
    10 & 0.697\tiny$\pm$0.004 & 0.517\tiny$\pm$0.008 & 0.658\tiny$\pm$0.096 & 0.412\tiny$\pm$0.144 & \cellcolor[gray]{0.85}\textbf{1.80\tiny$\pm$0.43} & 0.728\tiny$\pm$0.013 & 0.583\tiny$\pm$0.019 & 0.780\tiny$\pm$0.094 & 0.655\tiny$\pm$0.093 & 11.15\tiny$\pm$1.04 & \cellcolor[gray]{0.85}\textbf{0.776\tiny$\pm$0.002} & \cellcolor[gray]{0.85}\textbf{0.651\tiny$\pm$0.004} & \cellcolor[gray]{0.85}\textbf{0.835\tiny$\pm$0.109} & \cellcolor[gray]{0.85}\textbf{0.723\tiny$\pm$0.115} & 3.52\tiny$\pm$1.29 \\[3pt]
    
    11 & 0.755\tiny$\pm$0.015 & 0.304\tiny$\pm$0.014 & 0.623\tiny$\pm$0.181 & 0.100\tiny$\pm$0.070 & 1.41\tiny$\pm$0.26 & 0.608\tiny$\pm$0.019 & 0.171\tiny$\pm$0.013 & 0.583\tiny$\pm$0.104 & 0.133\tiny$\pm$0.064 & 5.42\tiny$\pm$0.83 & \cellcolor[gray]{0.85}\textbf{0.855\tiny$\pm$0.006} & \cellcolor[gray]{0.85}\textbf{0.485\tiny$\pm$0.012} & \cellcolor[gray]{0.85}\textbf{0.676\tiny$\pm$0.269} & \cellcolor[gray]{0.85}\textbf{0.198\tiny$\pm$0.193} & \cellcolor[gray]{0.95}\textbf{1.15\tiny$\pm$0.22} \\[3pt]
    
    12 & 0.623\tiny$\pm$0.012 & 0.286\tiny$\pm$0.009 & 0.556\tiny$\pm$0.085 & 0.195\tiny$\pm$0.061 & \cellcolor[gray]{0.95}\textbf{1.38\tiny$\pm$0.28} & 0.589\tiny$\pm$0.022 & 0.257\tiny$\pm$0.013 & 0.590\tiny$\pm$0.088 & \cellcolor[gray]{0.95}\textbf{0.214\tiny$\pm$0.049} & 8.45\tiny$\pm$1.35 & \cellcolor[gray]{0.85}\textbf{0.711\tiny$\pm$0.006} & \cellcolor[gray]{0.85}\textbf{0.355\tiny$\pm$0.010} & \cellcolor[gray]{0.85}\textbf{0.650\tiny$\pm$0.121} & 0.200\tiny$\pm$0.110 & 1.74\tiny$\pm$0.56 \\[3pt]
    
    13 & 0.650\tiny$\pm$0.003 & 0.553\tiny$\pm$0.010 & 0.642\tiny$\pm$0.085 & 0.620\tiny$\pm$0.070 & \cellcolor[gray]{0.85}\textbf{0.99\tiny$\pm$0.42} & 0.638\tiny$\pm$0.003 & 0.519\tiny$\pm$0.010 & 0.670\tiny$\pm$0.088 & 0.561\tiny$\pm$0.075 & 15.71\tiny$\pm$0.34 & \cellcolor[gray]{0.85}\textbf{0.696\tiny$\pm$0.002} & \cellcolor[gray]{0.85}\textbf{0.599\tiny$\pm$0.004} & \cellcolor[gray]{0.85}\textbf{0.766\tiny$\pm$0.139} & \cellcolor[gray]{0.85}\textbf{0.668\tiny$\pm$0.125} & 3.61\tiny$\pm$0.64 \\[3pt]
    
    14 & 0.591\tiny$\pm$0.011 & 0.529\tiny$\pm$0.012 & 0.572\tiny$\pm$0.067 & 0.594\tiny$\pm$0.068 & 2.77\tiny$\pm$0.48 & 0.611\tiny$\pm$0.011 & 0.529\tiny$\pm$0.009 & 0.637\tiny$\pm$0.075 & 0.565\tiny$\pm$0.061 & 18.90\tiny$\pm$1.04 & \cellcolor[gray]{0.85}\textbf{0.676\tiny$\pm$0.005} & \cellcolor[gray]{0.85}\textbf{0.603\tiny$\pm$0.006} & \cellcolor[gray]{0.85}\textbf{0.714\tiny$\pm$0.087} & \cellcolor[gray]{0.85}\textbf{0.660\tiny$\pm$0.085} & \cellcolor[gray]{0.95}\textbf{1.86\tiny$\pm$0.62} \\[3pt]
    
    15 & 0.703\tiny$\pm$0.005 & 0.676\tiny$\pm$0.005 & 0.715\tiny$\pm$0.103 & 0.782\tiny$\pm$0.083 & \cellcolor[gray]{0.85}\textbf{1.41\tiny$\pm$0.42} & 0.600\tiny$\pm$0.011 & 0.544\tiny$\pm$0.009 & 0.626\tiny$\pm$0.071 & 0.586\tiny$\pm$0.062 & 16.93\tiny$\pm$1.17 & \cellcolor[gray]{0.85}\textbf{0.718\tiny$\pm$0.003} & \cellcolor[gray]{0.85}\textbf{0.685\tiny$\pm$0.004} & \cellcolor[gray]{0.85}\textbf{0.755\tiny$\pm$0.114} & \cellcolor[gray]{0.85}\textbf{0.789\tiny$\pm$0.082} & 3.39\tiny$\pm$0.92 \\[3pt]
    
    16 & 0.628\tiny$\pm$0.007 & 0.111\tiny$\pm$0.008 & 0.590\tiny$\pm$0.148 & 0.073\tiny$\pm$0.033 & \cellcolor[gray]{0.85}\textbf{0.58\tiny$\pm$0.14} & 0.597\tiny$\pm$0.021 & 0.109\tiny$\pm$0.010 & 0.558\tiny$\pm$0.113 & \cellcolor[gray]{0.95}\textbf{0.077\tiny$\pm$0.028} & 4.36\tiny$\pm$0.27 & \cellcolor[gray]{0.85}\textbf{0.665\tiny$\pm$0.009} & \cellcolor[gray]{0.85}\textbf{0.128\tiny$\pm$0.003} & \cellcolor[gray]{0.95}\textbf{0.606\tiny$\pm$0.174} & 0.072\tiny$\pm$0.043 & 1.05\tiny$\pm$0.36 \\[3pt]
    
    17 & 0.690\tiny$\pm$0.005 & 0.377\tiny$\pm$0.012 & 0.618\tiny$\pm$0.108 & 0.204\tiny$\pm$0.093 & \cellcolor[gray]{0.95}\textbf{1.91\tiny$\pm$0.27} & 0.666\tiny$\pm$0.017 & 0.343\tiny$\pm$0.017 & 0.674\tiny$\pm$0.074 & 0.324\tiny$\pm$0.056 & 14.12\tiny$\pm$0.48 & \cellcolor[gray]{0.85}\textbf{0.763\tiny$\pm$0.004} & \cellcolor[gray]{0.85}\textbf{0.482\tiny$\pm$0.014} & \cellcolor[gray]{0.85}\textbf{0.746\tiny$\pm$0.188} & \cellcolor[gray]{0.95}\textbf{0.413\tiny$\pm$0.175} & 2.70\tiny$\pm$1.70 \\[3pt]
    
    18 & 0.643\tiny$\pm$0.008 & 0.285\tiny$\pm$0.005 & 0.580\tiny$\pm$0.083 & 0.158\tiny$\pm$0.075 & \cellcolor[gray]{0.85}\textbf{0.98\tiny$\pm$0.19} & 0.672\tiny$\pm$0.013 & 0.345\tiny$\pm$0.003 & \cellcolor[gray]{0.95}\textbf{0.674\tiny$\pm$0.081} & \cellcolor[gray]{0.95}\textbf{0.387\tiny$\pm$0.112} & 6.42\tiny$\pm$0.82 & \cellcolor[gray]{0.85}\textbf{0.712\tiny$\pm$0.003} & \cellcolor[gray]{0.95}\textbf{0.356\tiny$\pm$0.009} & 0.658\tiny$\pm$0.101 & 0.273\tiny$\pm$0.097 & 1.75\tiny$\pm$0.19 \\[3pt]
    
    19 & 0.594\tiny$\pm$0.008 & 0.172\tiny$\pm$0.008 & 0.543\tiny$\pm$0.080 & 0.120\tiny$\pm$0.032 & \cellcolor[gray]{0.85}\textbf{0.55\tiny$\pm$0.15} & 0.547\tiny$\pm$0.022 & 0.154\tiny$\pm$0.007 & 0.520\tiny$\pm$0.075 & \cellcolor[gray]{0.85}\textbf{0.150\tiny$\pm$0.065} & 5.55\tiny$\pm$1.19 & \cellcolor[gray]{0.85}\textbf{0.611\tiny$\pm$0.005} & \cellcolor[gray]{0.85}\textbf{0.199\tiny$\pm$0.008} & \cellcolor[gray]{0.85}\textbf{0.568\tiny$\pm$0.095} & 0.119\tiny$\pm$0.036 & 1.50\tiny$\pm$0.29 \\[3pt]
    
    20 & 0.623\tiny$\pm$0.010 & 0.091\tiny$\pm$0.007 & 0.531\tiny$\pm$0.113 & 0.056\tiny$\pm$0.031 & 0.98\tiny$\pm$0.15 & 0.615\tiny$\pm$0.019 & \cellcolor[gray]{0.95}\textbf{0.103\tiny$\pm$0.009} & 0.523\tiny$\pm$0.091 & \cellcolor[gray]{0.85}\textbf{0.060\tiny$\pm$0.029} & 3.03\tiny$\pm$0.34 & \cellcolor[gray]{0.95}\textbf{0.649\tiny$\pm$0.011} & 0.098\tiny$\pm$0.003 & \cellcolor[gray]{0.95}\textbf{0.532\tiny$\pm$0.109} & 0.054\tiny$\pm$0.027 & \cellcolor[gray]{0.85}\textbf{0.68\tiny$\pm$0.10} \\[3pt]
    
    21 & 0.591\tiny$\pm$0.009 & 0.129\tiny$\pm$0.006 & 0.573\tiny$\pm$0.129 & 0.097\tiny$\pm$0.037 & 1.91\tiny$\pm$0.38 & 0.658\tiny$\pm$0.023 & \cellcolor[gray]{0.95}\textbf{0.179\tiny$\pm$0.013} & 0.569\tiny$\pm$0.102 & \cellcolor[gray]{0.95}\textbf{0.104\tiny$\pm$0.048} & 5.11\tiny$\pm$0.98 & \cellcolor[gray]{0.95}\textbf{0.663\tiny$\pm$0.014} & 0.172\tiny$\pm$0.006 & \cellcolor[gray]{0.95}\textbf{0.574\tiny$\pm$0.103} & 0.089\tiny$\pm$0.056 & \cellcolor[gray]{0.95}\textbf{1.43\tiny$\pm$0.48} \\[3pt]
    
    22 & 0.722\tiny$\pm$0.007 & 0.368\tiny$\pm$0.007 & 0.607\tiny$\pm$0.119 & 0.184\tiny$\pm$0.114 & \cellcolor[gray]{0.85}\textbf{1.10\tiny$\pm$0.27} & 0.665\tiny$\pm$0.009 & 0.288\tiny$\pm$0.008 & 0.624\tiny$\pm$0.118 & 0.203\tiny$\pm$0.070 & 8.30\tiny$\pm$1.29 & \cellcolor[gray]{0.85}\textbf{0.750\tiny$\pm$0.004} & \cellcolor[gray]{0.85}\textbf{0.402\tiny$\pm$0.005} & \cellcolor[gray]{0.85}\textbf{0.669\tiny$\pm$0.154} & \cellcolor[gray]{0.95}\textbf{0.214\tiny$\pm$0.140} & 2.04\tiny$\pm$0.83 \\[3pt]
    
    23 & 0.803\tiny$\pm$0.009 & 0.620\tiny$\pm$0.013 & 0.804\tiny$\pm$0.146 & 0.607\tiny$\pm$0.177 & \cellcolor[gray]{0.95}\textbf{1.57\tiny$\pm$0.36} & 0.714\tiny$\pm$0.008 & 0.493\tiny$\pm$0.017 & 0.725\tiny$\pm$0.102 & 0.475\tiny$\pm$0.095 & 11.41\tiny$\pm$1.37 & \cellcolor[gray]{0.85}\textbf{0.817\tiny$\pm$0.005} & \cellcolor[gray]{0.85}\textbf{0.645\tiny$\pm$0.007} & \cellcolor[gray]{0.85}\textbf{0.872\tiny$\pm$0.101} & \cellcolor[gray]{0.85}\textbf{0.738\tiny$\pm$0.115} & 2.09\tiny$\pm$0.85 \\[3pt]
    
    24 & 0.754\tiny$\pm$0.006 & 0.483\tiny$\pm$0.010 & 0.691\tiny$\pm$0.141 & 0.358\tiny$\pm$0.177 & \cellcolor[gray]{0.95}\textbf{1.88\tiny$\pm$0.40} & 0.649\tiny$\pm$0.005 & 0.350\tiny$\pm$0.008 & 0.626\tiny$\pm$0.099 & 0.302\tiny$\pm$0.097 & 11.07\tiny$\pm$1.03 & \cellcolor[gray]{0.85}\textbf{0.776\tiny$\pm$0.004} & \cellcolor[gray]{0.85}\textbf{0.525\tiny$\pm$0.010} & \cellcolor[gray]{0.85}\textbf{0.752\tiny$\pm$0.151} & \cellcolor[gray]{0.85}\textbf{0.475\tiny$\pm$0.178} & 2.38\tiny$\pm$0.68 \\[3pt]
    
    25 & 0.808\tiny$\pm$0.006 & 0.519\tiny$\pm$0.023 & 0.704\tiny$\pm$0.179 & 0.316\tiny$\pm$0.214 & \cellcolor[gray]{0.95}\textbf{1.55\tiny$\pm$0.48} & 0.693\tiny$\pm$0.008 & 0.336\tiny$\pm$0.006 & 0.659\tiny$\pm$0.110 & 0.301\tiny$\pm$0.106 & 8.71\tiny$\pm$0.31 & \cellcolor[gray]{0.85}\textbf{0.820\tiny$\pm$0.003} & \cellcolor[gray]{0.95}\textbf{0.554\tiny$\pm$0.005} & \cellcolor[gray]{0.85}\textbf{0.748\tiny$\pm$0.203} & \cellcolor[gray]{0.95}\textbf{0.413\tiny$\pm$0.242} & 1.65\tiny$\pm$0.35 \\[3pt]
    
    \midrule
    
    \textbf{Average} & 0.681\tiny$\pm$0.007 & 0.371\tiny$\pm$0.010 & 0.625\tiny$\pm$0.115 & 0.292\tiny$\pm$0.091 & \cellcolor[gray]{0.95}\textbf{1.42\tiny$\pm$0.34} & 0.654\tiny$\pm$0.013 & 0.350\tiny$\pm$0.012 & 0.643\tiny$\pm$0.095 & 0.342\tiny$\pm$0.077 & 9.41\tiny$\pm$0.83 & \cellcolor[gray]{0.95}\textbf{0.739\tiny$\pm$0.005} & \cellcolor[gray]{0.95}\textbf{0.449\tiny$\pm$0.008} & \cellcolor[gray]{0.95}\textbf{0.705\tiny$\pm$0.136} & \cellcolor[gray]{0.95}\textbf{0.396\tiny$\pm$0.112} & 2.08\tiny$\pm$0.73 \\[3pt]

    \bottomrule
    \end{tabular}
    }
    \label{table:eval_unimodals}
\end{table*}

%% file: tables/table_eval_metrics_new_multimodals.tex
\begin{table*}[h!]
\caption{
    \textbf{Condition-level Performance of Evaluation Metrics Across Multimodal Architectures.}
    MedFuse, DrFuse, and MeTra achieve broadly comparable discrimination performance across conditions, with MedFuse consistently attaining the highest AUROC/AUPRC values, while all models exhibit elevated calibration error.
    Shading and bolding are applied \emph{within each model} to compare standard AUC metrics against their selective AUC counterparts for the same condition.
    Under correct uncertainty estimation, selective AUROC/AUPRC should be statistically significantly higher than standard metrics, reflecting effective rejection of uncertain predictions.
    Instead, we observe no consistent improvement across architectures: standard AUC metrics frequently dominate their selective counterparts, indicating that selective prediction fails to provide systematic gains.
    These results further suggest that increased architectural complexity alone does not improve reliability or resolve calibration-driven failures in this task.
    \footnotesize{(Dark-bold: $p<0.05$, Wilcoxon signed-rank test, 5 seeds; Light-bold: highest mean, not significant)}}
    \centering
    \resizebox{\linewidth}{!}{
    \begin{tabular}{c|ccccc|ccccc|cccccc}
     \multirow{2}{*}[-1.75em]{\textbf{\makecell{Clinical \\ Condition}}} 
    & \multicolumn{5}{c}{\textbf{MedFuse}}
    & \multicolumn{5}{c}{\textbf{DrFuse}}
    & \multicolumn{5}{c}{\textbf{MeTra}}\\
    \midrule

    & \textbf{AUROC} 
    & \textbf{AUPRC} 
    & \textbf{\makecell{Selective \\ AUROC}}
    & \textbf{\makecell{Selective \\ AUPRC}}
    & \textbf{$\widehat{\mathbf{ECE}} \downarrow$} 

    & \textbf{AUROC} 
    & \textbf{AUPRC} 
    & \textbf{\makecell{Selective \\ AUROC}}
    & \textbf{\makecell{Selective \\ AUPRC}}
     & \textbf{$\widehat{\mathbf{ECE}} \downarrow$}

    & \textbf{AUROC} 
    & \textbf{AUPRC} 
    & \textbf{\makecell{Selective \\ AUROC}}
    & \textbf{\makecell{Selective \\ AUPRC}}
     & \textbf{$\widehat{\mathbf{ECE}} \downarrow$}\\
    \midrule
    
    1 & {0.761\tiny$\pm$0.005} & {0.589\tiny$\pm$0.008} & \cellcolor[gray]{0.85}\textbf{0.804\tiny$\pm$0.111} & \cellcolor[gray]{0.85}\textbf{0.634\tiny$\pm$0.109} & {2.31\tiny$\pm$0.97} & 0.732\tiny$\pm$0.004 & 0.551\tiny$\pm$0.008 & \cellcolor[gray]{0.85}\textbf{0.767\tiny$\pm$0.104} & \cellcolor[gray]{0.85}\textbf{0.585\tiny$\pm$0.113} & 4.06\tiny$\pm$1.21 & 0.722\tiny$\pm$0.009 & 0.541\tiny$\pm$0.011 & \cellcolor[gray]{0.85}\textbf{0.759\tiny$\pm$0.083} & \cellcolor[gray]{0.85}\textbf{0.573\tiny$\pm$0.071} & 2.73\tiny$\pm$0.39 \\[3pt]
    
    2 & \cellcolor[gray]{0.85}\textbf{0.856\tiny$\pm$0.010} & \cellcolor[gray]{0.85}\textbf{0.432\tiny$\pm$0.019} & 0.560\tiny$\pm$0.176 & 0.075\tiny$\pm$0.116 & {0.88\tiny$\pm$0.24} & \cellcolor[gray]{0.85}\textbf{0.899\tiny$\pm$0.004} & \cellcolor[gray]{0.85}\textbf{0.460\tiny$\pm$0.013} & 0.584\tiny$\pm$0.272 & {0.118\tiny$\pm$0.166} & 1.06\tiny$\pm$0.44 & \cellcolor[gray]{0.85}\textbf{0.900\tiny$\pm$0.004} & \cellcolor[gray]{0.85}\textbf{0.490\tiny$\pm$0.022} & {0.602\tiny$\pm$0.253} & 0.112\tiny$\pm$0.172 & 1.13\tiny$\pm$0.32 \\[3pt]
    
    3 & \cellcolor[gray]{0.85}\textbf{0.713\tiny$\pm$0.005} & \cellcolor[gray]{0.85}\textbf{0.209\tiny$\pm$0.010} & 0.633\tiny$\pm$0.207 & 0.083\tiny$\pm$0.049 & {0.76\tiny$\pm$0.23} & \cellcolor[gray]{0.85}\textbf{0.697\tiny$\pm$0.013} & \cellcolor[gray]{0.85}\textbf{0.166\tiny$\pm$0.010} & {0.637\tiny$\pm$0.169} & {0.090\tiny$\pm$0.069} & 2.21\tiny$\pm$0.58 & \cellcolor[gray]{0.85}\textbf{0.717\tiny$\pm$0.010} & \cellcolor[gray]{0.85}\textbf{0.197\tiny$\pm$0.014} & 0.610\tiny$\pm$0.158 & 0.085\tiny$\pm$0.058 & 1.98\tiny$\pm$1.12 \\[3pt]
    
    4 & {0.694\tiny$\pm$0.006} & {0.565\tiny$\pm$0.010} & \cellcolor[gray]{0.85}\textbf{0.727\tiny$\pm$0.079} & \cellcolor[gray]{0.85}\textbf{0.610\tiny$\pm$0.071} & {2.61\tiny$\pm$1.65} & 0.650\tiny$\pm$0.012 & 0.520\tiny$\pm$0.013 & \cellcolor[gray]{0.85}\textbf{0.692\tiny$\pm$0.084} & \cellcolor[gray]{0.85}\textbf{0.547\tiny$\pm$0.089} & 3.42\tiny$\pm$1.66 & 0.628\tiny$\pm$0.011 & 0.500\tiny$\pm$0.008 & \cellcolor[gray]{0.85}\textbf{0.648\tiny$\pm$0.076} & \cellcolor[gray]{0.85}\textbf{0.529\tiny$\pm$0.079} & 3.59\tiny$\pm$1.65 \\[3pt]
    
    5 & {0.770\tiny$\pm$0.003} & {0.519\tiny$\pm$0.010} & \cellcolor[gray]{0.85}\textbf{0.787\tiny$\pm$0.177} & \cellcolor[gray]{0.85}\textbf{0.521\tiny$\pm$0.169} & {2.34\tiny$\pm$2.04} & \cellcolor[gray]{0.85}\textbf{0.726\tiny$\pm$0.006} & \cellcolor[gray]{0.85}\textbf{0.448\tiny$\pm$0.009} & 0.693\tiny$\pm$0.127 & 0.334\tiny$\pm$0.152 & 3.59\tiny$\pm$2.23 & \cellcolor[gray]{0.85}\textbf{0.709\tiny$\pm$0.017} & \cellcolor[gray]{0.85}\textbf{0.414\tiny$\pm$0.019} & 0.668\tiny$\pm$0.101 & 0.299\tiny$\pm$0.105 & 2.68\tiny$\pm$1.31 \\[3pt]
    
    6 & \cellcolor[gray]{0.85}\textbf{0.742\tiny$\pm$0.004} & \cellcolor[gray]{0.85}\textbf{0.345\tiny$\pm$0.011} & {0.717\tiny$\pm$0.120} & {0.279\tiny$\pm$0.127} & 2.17\tiny$\pm$0.45 & \cellcolor[gray]{0.85}\textbf{0.670\tiny$\pm$0.013} & \cellcolor[gray]{0.85}\textbf{0.274\tiny$\pm$0.015} & 0.573\tiny$\pm$0.099 & 0.137\tiny$\pm$0.060 & {1.94\tiny$\pm$0.66} & \cellcolor[gray]{0.85}\textbf{0.618\tiny$\pm$0.010} & \cellcolor[gray]{0.85}\textbf{0.219\tiny$\pm$0.007} & 0.557\tiny$\pm$0.086 & 0.140\tiny$\pm$0.042 & 2.52\tiny$\pm$0.61 \\[3pt]
    
    7 & \cellcolor[gray]{0.85}\textbf{0.673\tiny$\pm$0.001} & \cellcolor[gray]{0.85}\textbf{0.383\tiny$\pm$0.003} & 0.602\tiny$\pm$0.094 & 0.241\tiny$\pm$0.100 & {1.46\tiny$\pm$0.31} & \cellcolor[gray]{0.85}\textbf{0.693\tiny$\pm$0.008} & \cellcolor[gray]{0.85}\textbf{0.409\tiny$\pm$0.008} & 0.662\tiny$\pm$0.097 & 0.349\tiny$\pm$0.097 & 1.90\tiny$\pm$0.93 & \cellcolor[gray]{0.85}\textbf{0.701\tiny$\pm$0.003} & \cellcolor[gray]{0.85}\textbf{0.427\tiny$\pm$0.005} & {0.673\tiny$\pm$0.091} & {0.380\tiny$\pm$0.104} & 1.76\tiny$\pm$0.51 \\[3pt]
    
    8 & {0.824\tiny$\pm$0.004} & {0.609\tiny$\pm$0.009} & \cellcolor[gray]{0.85}\textbf{0.826\tiny$\pm$0.099} & \cellcolor[gray]{0.85}\textbf{0.648\tiny$\pm$0.119} & 2.32\tiny$\pm$0.45 & \cellcolor[gray]{0.85}\textbf{0.691\tiny$\pm$0.007} & \cellcolor[gray]{0.85}\textbf{0.234\tiny$\pm$0.006} & 0.611\tiny$\pm$0.121 & 0.104\tiny$\pm$0.066 & {2.30\tiny$\pm$0.70} & \cellcolor[gray]{0.85}\textbf{0.674\tiny$\pm$0.007} & \cellcolor[gray]{0.85}\textbf{0.220\tiny$\pm$0.011} & 0.592\tiny$\pm$0.111 & 0.107\tiny$\pm$0.051 & 2.61\tiny$\pm$0.72 \\[3pt]
    
    9 & {0.792\tiny$\pm$0.004} & {0.625\tiny$\pm$0.008} & \cellcolor[gray]{0.85}\textbf{0.842\tiny$\pm$0.110} & \cellcolor[gray]{0.85}\textbf{0.703\tiny$\pm$0.123} & 4.50\tiny$\pm$1.71 & 0.732\tiny$\pm$0.006 & \cellcolor[gray]{0.85}\textbf{0.544\tiny$\pm$0.009} & \cellcolor[gray]{0.85}\textbf{0.751\tiny$\pm$0.137} & 0.527\tiny$\pm$0.169 & 3.95\tiny$\pm$2.06 & 0.705\tiny$\pm$0.013 & 0.503\tiny$\pm$0.019 & \cellcolor[gray]{0.85}\textbf{0.723\tiny$\pm$0.103} & \cellcolor[gray]{0.85}\textbf{0.515\tiny$\pm$0.120} & {2.02\tiny$\pm$0.62} \\[3pt]
    
    10 & {0.776\tiny$\pm$0.002} & {0.651\tiny$\pm$0.004} & \cellcolor[gray]{0.85}\textbf{0.835\tiny$\pm$0.109} & \cellcolor[gray]{0.85}\textbf{0.723\tiny$\pm$0.115} & 3.52\tiny$\pm$1.29 & 0.735\tiny$\pm$0.005 & \cellcolor[gray]{0.85}\textbf{0.575\tiny$\pm$0.005} & \cellcolor[gray]{0.85}\textbf{0.764\tiny$\pm$0.123} & 0.574\tiny$\pm$0.154 & 4.17\tiny$\pm$3.96 & 0.705\tiny$\pm$0.008 & 0.545\tiny$\pm$0.009 & \cellcolor[gray]{0.85}\textbf{0.734\tiny$\pm$0.093} & \cellcolor[gray]{0.85}\textbf{0.560\tiny$\pm$0.102} & {3.28\tiny$\pm$0.71} \\[3pt]
    
    11 & \cellcolor[gray]{0.85}\textbf{0.855\tiny$\pm$0.006} & \cellcolor[gray]{0.85}\textbf{0.485\tiny$\pm$0.012} & 0.676\tiny$\pm$0.269 & 0.198\tiny$\pm$0.193 & {1.15\tiny$\pm$0.22} & \cellcolor[gray]{0.85}\textbf{0.867\tiny$\pm$0.005} & \cellcolor[gray]{0.85}\textbf{0.502\tiny$\pm$0.015} & 0.775\tiny$\pm$0.267 & 0.464\tiny$\pm$0.226 & 1.41\tiny$\pm$0.45 & \cellcolor[gray]{0.85}\textbf{0.838\tiny$\pm$0.008} & 0.450\tiny$\pm$0.027 & {0.822\tiny$\pm$0.192} & \cellcolor[gray]{0.85}\textbf{0.487\tiny$\pm$0.234} & 2.22\tiny$\pm$1.11 \\[3pt]
    
    12 & \cellcolor[gray]{0.85}\textbf{0.711\tiny$\pm$0.006} & \cellcolor[gray]{0.85}\textbf{0.355\tiny$\pm$0.010} & 0.650\tiny$\pm$0.121 & 0.200\tiny$\pm$0.110 & {1.74\tiny$\pm$0.56} & \cellcolor[gray]{0.85}\textbf{0.726\tiny$\pm$0.007} & \cellcolor[gray]{0.85}\textbf{0.377\tiny$\pm$0.007} & {0.673\tiny$\pm$0.131} & {0.262\tiny$\pm$0.132} & 3.52\tiny$\pm$1.24 & \cellcolor[gray]{0.85}\textbf{0.700\tiny$\pm$0.020} & \cellcolor[gray]{0.85}\textbf{0.361\tiny$\pm$0.024} & 0.620\tiny$\pm$0.116 & 0.223\tiny$\pm$0.117 & 2.22\tiny$\pm$0.45 \\[3pt]
    
    13 & {0.696\tiny$\pm$0.002} & {0.599\tiny$\pm$0.004} & \cellcolor[gray]{0.85}\textbf{0.766\tiny$\pm$0.139} & \cellcolor[gray]{0.85}\textbf{0.668\tiny$\pm$0.125} & 3.61\tiny$\pm$0.64 & 0.678\tiny$\pm$0.005 & 0.596\tiny$\pm$0.010 & \cellcolor[gray]{0.85}\textbf{0.706\tiny$\pm$0.089} & \cellcolor[gray]{0.85}\textbf{0.670\tiny$\pm$0.094} & 3.72\tiny$\pm$1.65 & 0.650\tiny$\pm$0.008 & 0.568\tiny$\pm$0.011 & \cellcolor[gray]{0.85}\textbf{0.682\tiny$\pm$0.083} & \cellcolor[gray]{0.85}\textbf{0.647\tiny$\pm$0.085} & {2.57\tiny$\pm$1.67} \\[3pt]
    
    14 & {0.676\tiny$\pm$0.005} & {0.603\tiny$\pm$0.006} & \cellcolor[gray]{0.85}\textbf{0.714\tiny$\pm$0.087} & \cellcolor[gray]{0.85}\textbf{0.660\tiny$\pm$0.085} & {1.86\tiny$\pm$0.62} & \cellcolor[gray]{0.85}\textbf{0.647\tiny$\pm$0.008} & 0.580\tiny$\pm$0.013 & 0.639\tiny$\pm$0.086 & \cellcolor[gray]{0.85}\textbf{0.650\tiny$\pm$0.074} & 2.02\tiny$\pm$1.20 & \cellcolor[gray]{0.85}\textbf{0.627\tiny$\pm$0.011} & 0.560\tiny$\pm$0.009 & 0.623\tiny$\pm$0.087 & \cellcolor[gray]{0.85}\textbf{0.620\tiny$\pm$0.073} & 2.58\tiny$\pm$1.81 \\[3pt]
    
    15 & 0.718\tiny$\pm$0.003 & {0.685\tiny$\pm$0.004} & \cellcolor[gray]{0.85}\textbf{0.755\tiny$\pm$0.114} & \cellcolor[gray]{0.85}\textbf{0.789\tiny$\pm$0.082} & {3.39\tiny$\pm$0.92} & {0.720\tiny$\pm$0.005} & 0.676\tiny$\pm$0.005 & \cellcolor[gray]{0.85}\textbf{0.738\tiny$\pm$0.099} & \cellcolor[gray]{0.85}\textbf{0.773\tiny$\pm$0.074} & 3.78\tiny$\pm$0.92 & 0.698\tiny$\pm$0.007 & 0.660\tiny$\pm$0.017 & \cellcolor[gray]{0.85}\textbf{0.742\tiny$\pm$0.096} & \cellcolor[gray]{0.85}\textbf{0.754\tiny$\pm$0.085} & 3.54\tiny$\pm$0.48 \\[3pt]
    
    16 & \cellcolor[gray]{0.85}\textbf{0.665\tiny$\pm$0.009} & \cellcolor[gray]{0.85}\textbf{0.128\tiny$\pm$0.003} & 0.606\tiny$\pm$0.174 & {0.072\tiny$\pm$0.043} & 1.05\tiny$\pm$0.36 & \cellcolor[gray]{0.85}\textbf{0.711\tiny$\pm$0.006} & \cellcolor[gray]{0.85}\textbf{0.171\tiny$\pm$0.009} & 0.625\tiny$\pm$0.221 & 0.065\tiny$\pm$0.044 & {1.01\tiny$\pm$0.19} & \cellcolor[gray]{0.85}\textbf{0.690\tiny$\pm$0.017} & \cellcolor[gray]{0.85}\textbf{0.159\tiny$\pm$0.023} & {0.625\tiny$\pm$0.168} & 0.069\tiny$\pm$0.055 & 1.34\tiny$\pm$0.49 \\[3pt]
    
    17 & \cellcolor[gray]{0.85}\textbf{0.763\tiny$\pm$0.004} & \cellcolor[gray]{0.85}\textbf{0.482\tiny$\pm$0.014} & {0.746\tiny$\pm$0.188} & {0.413\tiny$\pm$0.175} & 2.70\tiny$\pm$1.70 & \cellcolor[gray]{0.85}\textbf{0.714\tiny$\pm$0.006} & \cellcolor[gray]{0.85}\textbf{0.413\tiny$\pm$0.008} & 0.665\tiny$\pm$0.123 & 0.281\tiny$\pm$0.138 & 3.71\tiny$\pm$2.60 & \cellcolor[gray]{0.85}\textbf{0.701\tiny$\pm$0.017} & \cellcolor[gray]{0.85}\textbf{0.386\tiny$\pm$0.018} & 0.635\tiny$\pm$0.103 & 0.247\tiny$\pm$0.098 & {2.52\tiny$\pm$0.89} \\[3pt]
    
    18 & \cellcolor[gray]{0.85}\textbf{0.712\tiny$\pm$0.003} & \cellcolor[gray]{0.85}\textbf{0.356\tiny$\pm$0.009} & {0.658\tiny$\pm$0.101} & {0.273\tiny$\pm$0.097} & {1.75\tiny$\pm$0.19} & \cellcolor[gray]{0.85}\textbf{0.700\tiny$\pm$0.005} & \cellcolor[gray]{0.85}\textbf{0.315\tiny$\pm$0.008} & 0.625\tiny$\pm$0.115 & 0.167\tiny$\pm$0.094 & 2.10\tiny$\pm$0.48 & \cellcolor[gray]{0.85}\textbf{0.674\tiny$\pm$0.010} & \cellcolor[gray]{0.85}\textbf{0.288\tiny$\pm$0.016} & 0.612\tiny$\pm$0.113 & 0.157\tiny$\pm$0.073 & 2.37\tiny$\pm$1.06 \\[3pt]
    
    19 & \cellcolor[gray]{0.85}\textbf{0.611\tiny$\pm$0.005} & \cellcolor[gray]{0.85}\textbf{0.199\tiny$\pm$0.008} & {0.568\tiny$\pm$0.095} & 0.119\tiny$\pm$0.036 & {1.50\tiny$\pm$0.29} & \cellcolor[gray]{0.85}\textbf{0.598\tiny$\pm$0.013} & \cellcolor[gray]{0.85}\textbf{0.171\tiny$\pm$0.006} & 0.556\tiny$\pm$0.082 & 0.120\tiny$\pm$0.032 & 1.57\tiny$\pm$0.58 & \cellcolor[gray]{0.85}\textbf{0.575\tiny$\pm$0.016} & \cellcolor[gray]{0.85}\textbf{0.157\tiny$\pm$0.006} & 0.538\tiny$\pm$0.080 & {0.124\tiny$\pm$0.036} & 1.67\tiny$\pm$0.44 \\[3pt]
    
    20 & \cellcolor[gray]{0.85}\textbf{0.649\tiny$\pm$0.011} & \cellcolor[gray]{0.85}\textbf{0.098\tiny$\pm$0.003} & 0.532\tiny$\pm$0.109 & 0.054\tiny$\pm$0.027 & {0.68\tiny$\pm$0.10} & \cellcolor[gray]{0.85}\textbf{0.694\tiny$\pm$0.015} & \cellcolor[gray]{0.85}\textbf{0.139\tiny$\pm$0.016} & {0.565\tiny$\pm$0.137} & 0.053\tiny$\pm$0.055 & 1.23\tiny$\pm$0.58 & \cellcolor[gray]{0.85}\textbf{0.671\tiny$\pm$0.016} & \cellcolor[gray]{0.85}\textbf{0.113\tiny$\pm$0.013} & 0.554\tiny$\pm$0.130 & {0.057\tiny$\pm$0.056} & 1.29\tiny$\pm$0.44 \\[3pt]
    
    21 & \cellcolor[gray]{0.85}\textbf{0.663\tiny$\pm$0.014} & \cellcolor[gray]{0.85}\textbf{0.172\tiny$\pm$0.006} & 0.574\tiny$\pm$0.103 & 0.089\tiny$\pm$0.056 & 1.43\tiny$\pm$0.48 & \cellcolor[gray]{0.85}\textbf{0.649\tiny$\pm$0.003} & \cellcolor[gray]{0.85}\textbf{0.166\tiny$\pm$0.007} & 0.569\tiny$\pm$0.108 & 0.093\tiny$\pm$0.046 & {1.37\tiny$\pm$0.77} & \cellcolor[gray]{0.85}\textbf{0.606\tiny$\pm$0.010} & \cellcolor[gray]{0.85}\textbf{0.131\tiny$\pm$0.007} & {0.596\tiny$\pm$0.114} & {0.097\tiny$\pm$0.036} & 2.26\tiny$\pm$0.22 \\[3pt]
    
    22 & \cellcolor[gray]{0.85}\textbf{0.750\tiny$\pm$0.004} & \cellcolor[gray]{0.85}\textbf{0.402\tiny$\pm$0.005} & {0.669\tiny$\pm$0.154} & {0.214\tiny$\pm$0.140} & 2.04\tiny$\pm$0.83 & \cellcolor[gray]{0.85}\textbf{0.762\tiny$\pm$0.002} & \cellcolor[gray]{0.85}\textbf{0.423\tiny$\pm$0.008} & 0.668\tiny$\pm$0.152 & 0.209\tiny$\pm$0.138 & 2.20\tiny$\pm$1.12 & \cellcolor[gray]{0.85}\textbf{0.740\tiny$\pm$0.008} & \cellcolor[gray]{0.85}\textbf{0.388\tiny$\pm$0.019} & 0.649\tiny$\pm$0.162 & 0.210\tiny$\pm$0.124 & {2.00\tiny$\pm$1.18} \\[3pt]
    
    23 & 0.817\tiny$\pm$0.005 & {0.645\tiny$\pm$0.007} & \cellcolor[gray]{0.85}\textbf{0.872\tiny$\pm$0.101} & \cellcolor[gray]{0.85}\textbf{0.738\tiny$\pm$0.115} & {2.09\tiny$\pm$0.85} & {0.819\tiny$\pm$0.003} & 0.621\tiny$\pm$0.008 & \cellcolor[gray]{0.85}\textbf{0.829\tiny$\pm$0.151} & \cellcolor[gray]{0.85}\textbf{0.630\tiny$\pm$0.169} & 2.94\tiny$\pm$1.00 & \cellcolor[gray]{0.85}\textbf{0.818\tiny$\pm$0.005} & \cellcolor[gray]{0.85}\textbf{0.620\tiny$\pm$0.019} & 0.800\tiny$\pm$0.170 & 0.543\tiny$\pm$0.209 & 2.36\tiny$\pm$0.82 \\[3pt]
    
    24 & \cellcolor[gray]{0.85}\textbf{0.776\tiny$\pm$0.004} & \cellcolor[gray]{0.85}\textbf{0.525\tiny$\pm$0.010} & 0.752\tiny$\pm$0.151 & 0.475\tiny$\pm$0.178 & {2.38\tiny$\pm$0.68} & \cellcolor[gray]{0.85}\textbf{0.797\tiny$\pm$0.005} & \cellcolor[gray]{0.85}\textbf{0.545\tiny$\pm$0.012} & {0.776\tiny$\pm$0.155} & {0.513\tiny$\pm$0.197} & 2.76\tiny$\pm$0.80 & \cellcolor[gray]{0.85}\textbf{0.783\tiny$\pm$0.008} & \cellcolor[gray]{0.85}\textbf{0.528\tiny$\pm$0.015} & 0.747\tiny$\pm$0.155 & 0.436\tiny$\pm$0.190 & 2.57\tiny$\pm$0.80 \\[3pt]
    
    25 & \cellcolor[gray]{0.85}\textbf{0.820\tiny$\pm$0.003} & \cellcolor[gray]{0.85}\textbf{0.554\tiny$\pm$0.005} & 0.748\tiny$\pm$0.203 & 0.413\tiny$\pm$0.242 & {1.65\tiny$\pm$0.35} & \cellcolor[gray]{0.85}\textbf{0.847\tiny$\pm$0.004} & \cellcolor[gray]{0.85}\textbf{0.571\tiny$\pm$0.014} & {0.772\tiny$\pm$0.233} & {0.484\tiny$\pm$0.253} & 1.71\tiny$\pm$0.67 & \cellcolor[gray]{0.85}\textbf{0.830\tiny$\pm$0.008} & \cellcolor[gray]{0.85}\textbf{0.552\tiny$\pm$0.026} & 0.719\tiny$\pm$0.205 & 0.367\tiny$\pm$0.236 & 2.18\tiny$\pm$1.00 \\[3pt]
    
    \midrule
    
    \textbf{Average} & 0.739\tiny$\pm$0.005 & 0.449\tiny$\pm$0.008 & 0.705\tiny$\pm$0.136 & 0.396\tiny$\pm$0.112 & 2.08\tiny$\pm$0.73 & 0.726\tiny$\pm$0.007 & 0.418\tiny$\pm$0.010 & 0.677\tiny$\pm$0.139 & 0.352\tiny$\pm$0.116 & 2.54\tiny$\pm$1.15 & 0.707\tiny$\pm$0.010 & 0.399\tiny$\pm$0.015 & 0.661\tiny$\pm$0.125 & 0.333\tiny$\pm$0.104 & 2.32\tiny$\pm$0.83 \\[3pt]

    \bottomrule
    \end{tabular}
    }
    \label{table:new_eval_multimodals}
\end{table*}

%% file: tables/table_ece_new_multimodals.tex
\begin{table*}[h!]
    \caption{
    \textbf{Condition-level Performance of ECE Stratification Across Multimodal Architectures.}
    Positive-class ECE dominates across conditions, highlighting systematic overconfidence in positive predictions across all multimodal architectures, independent of fusion complexity.
    Although MedFuse attains slightly lower mean ECE values, these differences are not consistently statistically significant across conditions.
    This indicates that no state-of-the-art multimodal architecture reliably resolves class-dependent calibration errors for this clinical task.
    \footnotesize{(Dark-bold: $p<0.05$, Wilcoxon signed-rank test, 5 seeds; Light-bold: highest mean, not significant)}
    }
    \centering
    \small 
    \begin{adjustbox}{max width=\linewidth}  
    \begin{tabular}{c|ccc|ccc|cccc}
    \multirow{2}{*}[-1.75em]{\textbf{\makecell{Clinical \\ Condition}}} 

    & \multicolumn{3}{c}{\textbf{MedFuse}}
    & \multicolumn{3}{c}{\textbf{DrFuse}}
    & \multicolumn{3}{c}{\textbf{MeTra}}\\
    \midrule
    \addlinespace[10pt]
    
    & \textbf{$\widehat{\mathbf{ECE}} \downarrow$}
    & \textbf{$\widehat{\mathbf{ECE}}_{c=1} \downarrow$}
    & \textbf{$\widehat{\mathbf{ECE}}_{c=0} \downarrow$}
    
     & \textbf{$\widehat{\mathbf{ECE}} \downarrow$}
    & \textbf{$\widehat{\mathbf{ECE}}_{c=1} \downarrow$}
    & \textbf{$\widehat{\mathbf{ECE}}_{c=0} \downarrow$}
    
     & \textbf{$\widehat{\mathbf{ECE}} \downarrow$}
    & \textbf{$\widehat{\mathbf{ECE}}_{c=1} \downarrow$}
    & \textbf{$\widehat{\mathbf{ECE}}_{c=0} \downarrow$}\\
    \midrule

    1 & \cellcolor[gray]{0.95}\textbf{2.31\tiny$\pm$0.97} & \cellcolor[gray]{0.95}\textbf{24.85\tiny$\pm$3.96} & 9.97\tiny$\pm$1.14 & 4.06\tiny$\pm$1.21 & 27.74\tiny$\pm$5.78 & \cellcolor[gray]{0.95}\textbf{7.01\tiny$\pm$2.40} & 2.73\tiny$\pm$0.39 & 35.27\tiny$\pm$2.04 & 13.65\tiny$\pm$1.70 \\[3pt]
    
    2 & \cellcolor[gray]{0.95}\textbf{0.88\tiny$\pm$0.24} & 50.56\tiny$\pm$3.07 & 3.56\tiny$\pm$0.19 & 1.06\tiny$\pm$0.44 & 37.08\tiny$\pm$7.82 & \cellcolor[gray]{0.95}\textbf{2.98\tiny$\pm$0.46} & 1.13\tiny$\pm$0.32 & \cellcolor[gray]{0.95}\textbf{33.53\tiny$\pm$10.49} & 3.52\tiny$\pm$0.15 \\[3pt]
    
    3 & \cellcolor[gray]{0.85}\textbf{0.76\tiny$\pm$0.23} & 86.93\tiny$\pm$0.14 & 8.05\tiny$\pm$0.25 & 2.21\tiny$\pm$0.58 & 88.15\tiny$\pm$2.27 & \cellcolor[gray]{0.95}\textbf{7.83\tiny$\pm$1.71} & 1.98\tiny$\pm$1.12 & \cellcolor[gray]{0.95}\textbf{84.65\tiny$\pm$2.99} & 9.32\tiny$\pm$1.95 \\[3pt]
    
    4 & \cellcolor[gray]{0.95}\textbf{2.61\tiny$\pm$1.65} & \cellcolor[gray]{0.95}\textbf{26.90\tiny$\pm$9.52} & \cellcolor[gray]{0.95}\textbf{13.57\tiny$\pm$2.63} & 3.42\tiny$\pm$1.66 & 35.79\tiny$\pm$13.68 & 17.54\tiny$\pm$7.67 & 3.59\tiny$\pm$1.65 & 39.75\tiny$\pm$13.97 & 19.53\tiny$\pm$5.75 \\[3pt]
    
    5 & \cellcolor[gray]{0.95}\textbf{2.34\tiny$\pm$2.04} & \cellcolor[gray]{0.95}\textbf{40.71\tiny$\pm$8.56} & \cellcolor[gray]{0.95}\textbf{10.21\tiny$\pm$1.06} & 3.59\tiny$\pm$2.23 & 49.62\tiny$\pm$9.38 & 12.31\tiny$\pm$1.96 & 2.68\tiny$\pm$1.31 & 60.85\tiny$\pm$4.41 & 16.17\tiny$\pm$0.66 \\[3pt]
    
    6 & 2.17\tiny$\pm$0.45 & \cellcolor[gray]{0.85}\textbf{61.34\tiny$\pm$3.97} & \cellcolor[gray]{0.85}\textbf{9.27\tiny$\pm$0.68} & \cellcolor[gray]{0.95}\textbf{1.94\tiny$\pm$0.66} & 79.53\tiny$\pm$1.91 & 13.44\tiny$\pm$1.82 & 2.52\tiny$\pm$0.61 & 80.43\tiny$\pm$0.66 & 15.52\tiny$\pm$1.17 \\[3pt]
    
    7 & \cellcolor[gray]{0.95}\textbf{1.46\tiny$\pm$0.31} & 66.13\tiny$\pm$1.23 & 19.12\tiny$\pm$0.40 & 1.90\tiny$\pm$0.93 & 65.82\tiny$\pm$3.70 & 18.38\tiny$\pm$1.68 & 1.76\tiny$\pm$0.51 & \cellcolor[gray]{0.95}\textbf{56.89\tiny$\pm$6.12} & \cellcolor[gray]{0.95}\textbf{16.28\tiny$\pm$0.59} \\[3pt]
    
    8 & 2.32\tiny$\pm$0.45 & \cellcolor[gray]{0.85}\textbf{41.64\tiny$\pm$0.88} & \cellcolor[gray]{0.85}\textbf{4.38\tiny$\pm$0.89} & \cellcolor[gray]{0.95}\textbf{2.30\tiny$\pm$0.70} & 82.47\tiny$\pm$3.45 & 9.75\tiny$\pm$1.62 & 2.61\tiny$\pm$0.72 & 82.51\tiny$\pm$4.15 & 10.35\tiny$\pm$1.64 \\[3pt]
    
    9 & 4.50\tiny$\pm$1.71 & \cellcolor[gray]{0.95}\textbf{24.13\tiny$\pm$9.07} & \cellcolor[gray]{0.95}\textbf{4.69\tiny$\pm$1.96} & 3.95\tiny$\pm$2.06 & 34.55\tiny$\pm$11.33 & 10.07\tiny$\pm$2.26 & \cellcolor[gray]{0.95}\textbf{2.02\tiny$\pm$0.62} & 38.30\tiny$\pm$9.77 & 14.65\tiny$\pm$3.27 \\[3pt]
    
    10 & 3.52\tiny$\pm$1.29 & \cellcolor[gray]{0.95}\textbf{22.44\tiny$\pm$4.96} & \cellcolor[gray]{0.95}\textbf{7.08\tiny$\pm$1.88} & 4.17\tiny$\pm$3.96 & 34.09\tiny$\pm$13.14 & 11.16\tiny$\pm$2.60 & \cellcolor[gray]{0.95}\textbf{3.28\tiny$\pm$0.71} & 37.68\tiny$\pm$7.23 & 16.13\tiny$\pm$1.71 \\[3pt]
    
    11 & \cellcolor[gray]{0.95}\textbf{1.15\tiny$\pm$0.22} & 47.13\tiny$\pm$2.20 & 6.30\tiny$\pm$0.34 & 1.41\tiny$\pm$0.45 & \cellcolor[gray]{0.95}\textbf{35.95\tiny$\pm$5.61} & \cellcolor[gray]{0.95}\textbf{5.01\tiny$\pm$0.89} & 2.22\tiny$\pm$1.11 & 55.17\tiny$\pm$11.94 & 6.25\tiny$\pm$1.24 \\[3pt]
    
    12 & \cellcolor[gray]{0.95}\textbf{1.74\tiny$\pm$0.56} & 68.84\tiny$\pm$2.85 & 17.64\tiny$\pm$0.79 & 3.52\tiny$\pm$1.24 & \cellcolor[gray]{0.95}\textbf{55.32\tiny$\pm$11.18} & \cellcolor[gray]{0.95}\textbf{16.14\tiny$\pm$3.87} & 2.22\tiny$\pm$0.45 & 62.38\tiny$\pm$8.01 & 16.60\tiny$\pm$1.96 \\[3pt]
    
    13 & 3.61\tiny$\pm$0.64 & \cellcolor[gray]{0.95}\textbf{12.51\tiny$\pm$4.83} & \cellcolor[gray]{0.95}\textbf{4.88\tiny$\pm$0.85} & 3.72\tiny$\pm$1.65 & 17.27\tiny$\pm$12.41 & 8.27\tiny$\pm$5.89 & \cellcolor[gray]{0.95}\textbf{2.57\tiny$\pm$1.67} & 27.10\tiny$\pm$8.21 & 15.39\tiny$\pm$2.24 \\[3pt]
    
    14 & \cellcolor[gray]{0.95}\textbf{1.86\tiny$\pm$0.62} & \cellcolor[gray]{0.95}\textbf{12.64\tiny$\pm$5.92} & \cellcolor[gray]{0.95}\textbf{9.09\tiny$\pm$4.73} & 2.02\tiny$\pm$1.20 & 16.93\tiny$\pm$11.53 & 12.38\tiny$\pm$6.58 & 2.58\tiny$\pm$1.81 & 14.50\tiny$\pm$10.83 & 10.32\tiny$\pm$7.64 \\[3pt]
    
    15 & \cellcolor[gray]{0.95}\textbf{3.39\tiny$\pm$0.92} & 12.05\tiny$\pm$2.66 & 3.95\tiny$\pm$0.52 & 3.78\tiny$\pm$0.92 & \cellcolor[gray]{0.95}\textbf{10.59\tiny$\pm$3.32} & \cellcolor[gray]{0.95}\textbf{3.01\tiny$\pm$1.87} & 3.54\tiny$\pm$0.48 & 14.38\tiny$\pm$6.59 & 7.02\tiny$\pm$3.27 \\[3pt]
    
    16 & 1.05\tiny$\pm$0.36 & 89.64\tiny$\pm$0.80 & 7.21\tiny$\pm$0.42 & \cellcolor[gray]{0.95}\textbf{1.01\tiny$\pm$0.19} & 88.91\tiny$\pm$0.80 & \cellcolor[gray]{0.95}\textbf{6.84\tiny$\pm$0.74} & 1.34\tiny$\pm$0.49 & \cellcolor[gray]{0.95}\textbf{88.77\tiny$\pm$2.63} & 6.89\tiny$\pm$0.98 \\[3pt]
    
    17 & 2.70\tiny$\pm$1.70 & \cellcolor[gray]{0.95}\textbf{46.22\tiny$\pm$7.91} & \cellcolor[gray]{0.95}\textbf{10.70\tiny$\pm$0.96} & 3.71\tiny$\pm$2.60 & 54.74\tiny$\pm$9.45 & 12.52\tiny$\pm$2.57 & \cellcolor[gray]{0.95}\textbf{2.52\tiny$\pm$0.89} & 65.36\tiny$\pm$3.14 & 15.65\tiny$\pm$0.56 \\[3pt]
    
    18 & \cellcolor[gray]{0.95}\textbf{1.75\tiny$\pm$0.19} & \cellcolor[gray]{0.95}\textbf{60.84\tiny$\pm$4.55} & \cellcolor[gray]{0.85}\textbf{11.05\tiny$\pm$0.92} & 2.10\tiny$\pm$0.48 & 70.47\tiny$\pm$3.99 & 14.39\tiny$\pm$1.54 & 2.37\tiny$\pm$1.06 & 78.01\tiny$\pm$3.50 & 14.10\tiny$\pm$1.39 \\[3pt]
    
    19 & \cellcolor[gray]{0.95}\textbf{1.50\tiny$\pm$0.29} & 86.71\tiny$\pm$0.43 & \cellcolor[gray]{0.95}\textbf{11.15\tiny$\pm$0.47} & 1.57\tiny$\pm$0.58 & 86.99\tiny$\pm$1.08 & 11.54\tiny$\pm$1.14 & 1.67\tiny$\pm$0.44 & \cellcolor[gray]{0.95}\textbf{86.28\tiny$\pm$0.82} & 12.37\tiny$\pm$0.55 \\[3pt]
    
    20 & \cellcolor[gray]{0.95}\textbf{0.68\tiny$\pm$0.10} & 92.06\tiny$\pm$0.39 & 5.53\tiny$\pm$0.33 & 1.23\tiny$\pm$0.58 & 91.04\tiny$\pm$0.93 & \cellcolor[gray]{0.95}\textbf{4.63\tiny$\pm$0.81} & 1.29\tiny$\pm$0.44 & \cellcolor[gray]{0.95}\textbf{90.78\tiny$\pm$2.42} & 5.28\tiny$\pm$1.12 \\[3pt]
    
    21 & 1.43\tiny$\pm$0.48 & \cellcolor[gray]{0.95}\textbf{87.93\tiny$\pm$0.48} & 8.56\tiny$\pm$0.60 & \cellcolor[gray]{0.95}\textbf{1.37\tiny$\pm$0.77} & 89.23\tiny$\pm$0.93 & \cellcolor[gray]{0.95}\textbf{8.20\tiny$\pm$0.81} & 2.26\tiny$\pm$0.22 & 88.25\tiny$\pm$1.49 & 9.52\tiny$\pm$1.33 \\[3pt]
    
    22 & 2.04\tiny$\pm$0.83 & 61.13\tiny$\pm$2.02 & 11.72\tiny$\pm$0.84 & 2.20\tiny$\pm$1.12 & \cellcolor[gray]{0.95}\textbf{59.74\tiny$\pm$7.21} & \cellcolor[gray]{0.95}\textbf{11.26\tiny$\pm$1.04} & \cellcolor[gray]{0.95}\textbf{2.00\tiny$\pm$1.18} & 62.64\tiny$\pm$6.49 & 12.51\tiny$\pm$1.24 \\[3pt]
    
    23 & \cellcolor[gray]{0.95}\textbf{2.09\tiny$\pm$0.85} & 25.38\tiny$\pm$3.17 & 7.17\tiny$\pm$0.15 & 2.94\tiny$\pm$1.00 & 22.12\tiny$\pm$3.08 & \cellcolor[gray]{0.95}\textbf{4.87\tiny$\pm$0.58} & 2.36\tiny$\pm$0.82 & \cellcolor[gray]{0.95}\textbf{18.30\tiny$\pm$5.56} & 5.82\tiny$\pm$1.39 \\[3pt]
    
    24 & \cellcolor[gray]{0.95}\textbf{2.38\tiny$\pm$0.68} & 43.16\tiny$\pm$1.67 & 9.60\tiny$\pm$0.84 & 2.76\tiny$\pm$0.80 & \cellcolor[gray]{0.95}\textbf{36.35\tiny$\pm$4.61} & \cellcolor[gray]{0.95}\textbf{7.43\tiny$\pm$1.94} & 2.57\tiny$\pm$0.80 & 45.99\tiny$\pm$7.53 & 10.29\tiny$\pm$1.44 \\[3pt]
    
    25 & \cellcolor[gray]{0.95}\textbf{1.65\tiny$\pm$0.35} & 38.14\tiny$\pm$1.48 & 7.12\tiny$\pm$0.45 & 1.71\tiny$\pm$0.67 & \cellcolor[gray]{0.85}\textbf{28.90\tiny$\pm$2.32} & \cellcolor[gray]{0.85}\textbf{4.83\tiny$\pm$1.20} & 2.18\tiny$\pm$1.00 & 40.13\tiny$\pm$7.60 & 7.04\tiny$\pm$0.57 \\[3pt]
    
    \midrule
    
    \textbf{Average} & 2.08\tiny$\pm$0.73 & 49.20\tiny$\pm$3.47 & 8.86\tiny$\pm$0.97 & 2.54\tiny$\pm$1.15 & 51.98\tiny$\pm$6.04 & 9.67\tiny$\pm$2.23 & 2.32\tiny$\pm$0.83 & 55.52\tiny$\pm$5.94 & 11.61\tiny$\pm$1.82 \\[3pt]
    
    \bottomrule
    \end{tabular}

    \end{adjustbox}
    \label{table:new_ece_multimodals}
\end{table*}

%% file: tables/table_eval_metrics_upscaled_multimodals.tex
\begin{table*}[h!]
\caption{
    \textbf{Condition-level Performance of Evaluation Metrics Across Multimodal Architectures under Loss-Upweighting.}
    Although the performance of all three models is comparable, MedFuse outperforms both DrFuse and MeTra across different conditions and evaluation metrics, with the downside that all models present high ECE scores and the difference between models is not statistically significant.
    This further confirms that complex architectures do not necessarily improve predictive performance and do not solve the reliability issues observed in this clinical task.
    \footnotesize{(Dark-bold: $p<0.05$, Wilcoxon signed-rank test, 5 seeds; Light-bold: highest mean, not significant)}
}
    \centering
    \resizebox{\linewidth}{!}{
    \begin{tabular}{c|ccccc|ccccc|cccccc}
     \multirow{2}{*}[-1.5em]{\textbf{\makecell{Clinical \\ Condition}}}  
    & \multicolumn{5}{c}{\textbf{MedFuse (Loss-Upweighting)}}
    & \multicolumn{5}{c}{\textbf{DrFuse (Loss-Upweighting)}}
    & \multicolumn{5}{c}{\textbf{MeTra (Loss-Upweighting)}}\\[-1pt]
    \midrule

    & \textbf{AUROC} 
    & \textbf{AUPRC} 
    & \textbf{\makecell{Selective \\ AUROC}}
    & \textbf{\makecell{Selective \\ AUPRC}}
    & \textbf{$\widehat{\mathbf{ECE}} \downarrow$} 

    & \textbf{AUROC} 
    & \textbf{AUPRC} 
    & \textbf{\makecell{Selective \\ AUROC}}
    & \textbf{\makecell{Selective \\ AUPRC}}
     & \textbf{$\widehat{\mathbf{ECE}} \downarrow$}

    & \textbf{AUROC} 
    & \textbf{AUPRC} 
    & \textbf{\makecell{Selective \\ AUROC}}
    & \textbf{\makecell{Selective \\ AUPRC}}
     & \textbf{$\widehat{\mathbf{ECE}} \downarrow$}\\[5pt]
    \midrule

    1 & \cellcolor[gray]{0.95}\textbf{0.761\tiny$\pm$0.003} & \cellcolor[gray]{0.95}\textbf{0.588\tiny$\pm$0.008} & \cellcolor[gray]{0.95}\textbf{0.756\tiny$\pm$0.088} & \cellcolor[gray]{0.95}\textbf{0.672\tiny$\pm$0.065} & \cellcolor[gray]{0.95}\textbf{3.09\tiny$\pm$0.86} & 0.731\tiny$\pm$0.004 & 0.547\tiny$\pm$0.007 & 0.717\tiny$\pm$0.090 & 0.628\tiny$\pm$0.081 & 4.58\tiny$\pm$0.98 & 0.715\tiny$\pm$0.008 & 0.533\tiny$\pm$0.012 & 0.721\tiny$\pm$0.088 & 0.618\tiny$\pm$0.077 & 3.35\tiny$\pm$1.48 \\[3pt]

    2 & 0.871\tiny$\pm$0.006 & 0.431\tiny$\pm$0.010 & 0.872\tiny$\pm$0.093 & 0.491\tiny$\pm$0.069 & \cellcolor[gray]{0.95}\textbf{2.38\tiny$\pm$0.66} & 0.895\tiny$\pm$0.005 & 0.473\tiny$\pm$0.010 & \cellcolor[gray]{0.95}\textbf{0.878\tiny$\pm$0.153} & \cellcolor[gray]{0.95}\textbf{0.547\tiny$\pm$0.135} & 4.44\tiny$\pm$1.50 & \cellcolor[gray]{0.95}\textbf{0.897\tiny$\pm$0.007} & \cellcolor[gray]{0.95}\textbf{0.478\tiny$\pm$0.034} & 0.871\tiny$\pm$0.111 & 0.546\tiny$\pm$0.078 & 3.79\tiny$\pm$0.79 \\[3pt]
    
    3 & \cellcolor[gray]{0.95}\textbf{0.716\tiny$\pm$0.003} & \cellcolor[gray]{0.95}\textbf{0.205\tiny$\pm$0.005} & \cellcolor[gray]{0.95}\textbf{0.744\tiny$\pm$0.167} & \cellcolor[gray]{0.95}\textbf{0.260\tiny$\pm$0.097} & \cellcolor[gray]{0.95}\textbf{9.58\tiny$\pm$2.12} & 0.693\tiny$\pm$0.007 & 0.175\tiny$\pm$0.007 & 0.687\tiny$\pm$0.118 & 0.205\tiny$\pm$0.073 & 13.34\tiny$\pm$6.55 & 0.699\tiny$\pm$0.027 & 0.182\tiny$\pm$0.012 & 0.694\tiny$\pm$0.143 & 0.205\tiny$\pm$0.070 & 10.09\tiny$\pm$3.99 \\[3pt]
    
    4 & \cellcolor[gray]{0.95}\textbf{0.692\tiny$\pm$0.005} & \cellcolor[gray]{0.95}\textbf{0.563\tiny$\pm$0.007} & \cellcolor[gray]{0.95}\textbf{0.650\tiny$\pm$0.084} & \cellcolor[gray]{0.95}\textbf{0.631\tiny$\pm$0.060} & 2.89\tiny$\pm$1.26 & 0.621\tiny$\pm$0.005 & 0.492\tiny$\pm$0.005 & 0.589\tiny$\pm$0.073 & 0.550\tiny$\pm$0.066 & 4.58\tiny$\pm$2.03 & 0.630\tiny$\pm$0.009 & 0.496\tiny$\pm$0.010 & 0.595\tiny$\pm$0.073 & 0.543\tiny$\pm$0.068 & \cellcolor[gray]{0.95}\textbf{1.93\tiny$\pm$0.65} \\[3pt]
    
    5 & \cellcolor[gray]{0.95}\textbf{0.760\tiny$\pm$0.006} & \cellcolor[gray]{0.95}\textbf{0.496\tiny$\pm$0.012} & \cellcolor[gray]{0.95}\textbf{0.805\tiny$\pm$0.094} & \cellcolor[gray]{0.95}\textbf{0.599\tiny$\pm$0.113} & \cellcolor[gray]{0.95}\textbf{4.60\tiny$\pm$1.52} & 0.718\tiny$\pm$0.002 & 0.437\tiny$\pm$0.006 & 0.726\tiny$\pm$0.085 & 0.522\tiny$\pm$0.088 & 5.90\tiny$\pm$2.72 & 0.696\tiny$\pm$0.015 & 0.402\tiny$\pm$0.013 & 0.678\tiny$\pm$0.095 & 0.467\tiny$\pm$0.080 & 4.98\tiny$\pm$3.28 \\[3pt]
    
    6 & \cellcolor[gray]{0.95}\textbf{0.747\tiny$\pm$0.007} & \cellcolor[gray]{0.95}\textbf{0.348\tiny$\pm$0.009} & \cellcolor[gray]{0.95}\textbf{0.803\tiny$\pm$0.099} & \cellcolor[gray]{0.95}\textbf{0.444\tiny$\pm$0.102} & 4.78\tiny$\pm$2.64 & 0.653\tiny$\pm$0.005 & 0.259\tiny$\pm$0.004 & 0.619\tiny$\pm$0.076 & 0.320\tiny$\pm$0.089 & 4.36\tiny$\pm$2.76 & 0.615\tiny$\pm$0.009 & 0.206\tiny$\pm$0.007 & 0.590\tiny$\pm$0.072 & 0.223\tiny$\pm$0.032 & \cellcolor[gray]{0.95}\textbf{3.57\tiny$\pm$1.35} \\[3pt]
    
    7 & 0.670\tiny$\pm$0.003 & 0.390\tiny$\pm$0.009 & 0.645\tiny$\pm$0.074 & 0.476\tiny$\pm$0.095 & 5.90\tiny$\pm$1.22 & 0.690\tiny$\pm$0.003 & 0.409\tiny$\pm$0.004 & 0.669\tiny$\pm$0.074 & \cellcolor[gray]{0.95}\textbf{0.512\tiny$\pm$0.106} & \cellcolor[gray]{0.95}\textbf{2.46\tiny$\pm$1.79} & \cellcolor[gray]{0.95}\textbf{0.699\tiny$\pm$0.012} & \cellcolor[gray]{0.95}\textbf{0.416\tiny$\pm$0.018} & \cellcolor[gray]{0.95}\textbf{0.683\tiny$\pm$0.078} & 0.505\tiny$\pm$0.093 & 2.96\tiny$\pm$0.57 \\[3pt]
    
    8 & \cellcolor[gray]{0.95}\textbf{0.819\tiny$\pm$0.009} & \cellcolor[gray]{0.95}\textbf{0.584\tiny$\pm$0.014} & \cellcolor[gray]{0.95}\textbf{0.860\tiny$\pm$0.122} & \cellcolor[gray]{0.95}\textbf{0.734\tiny$\pm$0.103} & \cellcolor[gray]{0.95}\textbf{4.61\tiny$\pm$0.84} & 0.673\tiny$\pm$0.007 & 0.233\tiny$\pm$0.009 & 0.720\tiny$\pm$0.090 & 0.325\tiny$\pm$0.115 & 8.29\tiny$\pm$2.82 & 0.674\tiny$\pm$0.005 & 0.210\tiny$\pm$0.012 & 0.684\tiny$\pm$0.093 & 0.250\tiny$\pm$0.064 & 6.13\tiny$\pm$3.48 \\[3pt]
    
    9 & \cellcolor[gray]{0.95}\textbf{0.789\tiny$\pm$0.007} & \cellcolor[gray]{0.95}\textbf{0.615\tiny$\pm$0.007} & \cellcolor[gray]{0.95}\textbf{0.840\tiny$\pm$0.101} & \cellcolor[gray]{0.95}\textbf{0.727\tiny$\pm$0.121} & 5.97\tiny$\pm$2.62 & 0.722\tiny$\pm$0.004 & 0.533\tiny$\pm$0.011 & 0.754\tiny$\pm$0.090 & 0.648\tiny$\pm$0.108 & 8.07\tiny$\pm$2.26 & 0.690\tiny$\pm$0.017 & 0.481\tiny$\pm$0.025 & 0.708\tiny$\pm$0.086 & 0.571\tiny$\pm$0.094 & \cellcolor[gray]{0.95}\textbf{4.83\tiny$\pm$3.15} \\[3pt]
    
    10 & \cellcolor[gray]{0.95}\textbf{0.773\tiny$\pm$0.005} & \cellcolor[gray]{0.95}\textbf{0.641\tiny$\pm$0.009} & \cellcolor[gray]{0.95}\textbf{0.809\tiny$\pm$0.093} & \cellcolor[gray]{0.95}\textbf{0.732\tiny$\pm$0.100} & 5.86\tiny$\pm$2.01 & 0.725\tiny$\pm$0.007 & 0.566\tiny$\pm$0.007 & 0.689\tiny$\pm$0.102 & 0.644\tiny$\pm$0.066 & 4.17\tiny$\pm$3.15 & 0.696\tiny$\pm$0.012 & 0.535\tiny$\pm$0.015 & 0.681\tiny$\pm$0.093 & 0.608\tiny$\pm$0.073 & \cellcolor[gray]{0.95}\textbf{2.16\tiny$\pm$1.13} \\[3pt]
    
    11 & 0.861\tiny$\pm$0.006 & 0.484\tiny$\pm$0.026 & 0.901\tiny$\pm$0.105 & 0.607\tiny$\pm$0.126 & \cellcolor[gray]{0.95}\textbf{4.62\tiny$\pm$0.72} & \cellcolor[gray]{0.85}\textbf{0.867\tiny$\pm$0.003} & \cellcolor[gray]{0.95}\textbf{0.506\tiny$\pm$0.004} & \cellcolor[gray]{0.95}\textbf{0.904\tiny$\pm$0.097} & \cellcolor[gray]{0.95}\textbf{0.639\tiny$\pm$0.120} & 6.14\tiny$\pm$2.96 & 0.821\tiny$\pm$0.017 & 0.426\tiny$\pm$0.012 & 0.833\tiny$\pm$0.109 & 0.553\tiny$\pm$0.117 & 6.68\tiny$\pm$2.72 \\[3pt]
    
    12 & 0.713\tiny$\pm$0.008 & 0.367\tiny$\pm$0.013 & \cellcolor[gray]{0.95}\textbf{0.720\tiny$\pm$0.107} & 0.413\tiny$\pm$0.091 & \cellcolor[gray]{0.95}\textbf{6.83\tiny$\pm$1.26} & \cellcolor[gray]{0.95}\textbf{0.722\tiny$\pm$0.004} & \cellcolor[gray]{0.95}\textbf{0.379\tiny$\pm$0.005} & 0.712\tiny$\pm$0.101 & \cellcolor[gray]{0.85}\textbf{0.444\tiny$\pm$0.098} & 7.58\tiny$\pm$4.02 & 0.690\tiny$\pm$0.012 & 0.349\tiny$\pm$0.017 & 0.651\tiny$\pm$0.095 & 0.390\tiny$\pm$0.064 & 9.57\tiny$\pm$3.54 \\[3pt]
    
    13 & \cellcolor[gray]{0.95}\textbf{0.689\tiny$\pm$0.004} & 0.590\tiny$\pm$0.008 & \cellcolor[gray]{0.95}\textbf{0.733\tiny$\pm$0.123} & 0.655\tiny$\pm$0.113 & 5.94\tiny$\pm$0.46 & 0.674\tiny$\pm$0.004 & \cellcolor[gray]{0.95}\textbf{0.592\tiny$\pm$0.006} & 0.639\tiny$\pm$0.081 & \cellcolor[gray]{0.95}\textbf{0.672\tiny$\pm$0.073} & 3.74\tiny$\pm$3.10 & 0.650\tiny$\pm$0.004 & 0.563\tiny$\pm$0.006 & 0.631\tiny$\pm$0.083 & 0.644\tiny$\pm$0.076 & \cellcolor[gray]{0.95}\textbf{3.22\tiny$\pm$1.88} \\[3pt]
    
    14 & \cellcolor[gray]{0.95}\textbf{0.666\tiny$\pm$0.009} & \cellcolor[gray]{0.95}\textbf{0.600\tiny$\pm$0.007} & \cellcolor[gray]{0.95}\textbf{0.684\tiny$\pm$0.085} & \cellcolor[gray]{0.95}\textbf{0.659\tiny$\pm$0.094} & \cellcolor[gray]{0.95}\textbf{2.25\tiny$\pm$0.77} & 0.641\tiny$\pm$0.005 & 0.573\tiny$\pm$0.005 & 0.585\tiny$\pm$0.072 & 0.637\tiny$\pm$0.066 & 2.47\tiny$\pm$1.43 & 0.624\tiny$\pm$0.004 & 0.561\tiny$\pm$0.009 & 0.588\tiny$\pm$0.077 & 0.625\tiny$\pm$0.069 & 2.47\tiny$\pm$0.35 \\[3pt]
    
    15 & 0.714\tiny$\pm$0.002 & 0.676\tiny$\pm$0.004 & \cellcolor[gray]{0.95}\textbf{0.727\tiny$\pm$0.086} & \cellcolor[gray]{0.95}\textbf{0.772\tiny$\pm$0.078} & 3.22\tiny$\pm$0.62 & \cellcolor[gray]{0.95}\textbf{0.721\tiny$\pm$0.005} & \cellcolor[gray]{0.95}\textbf{0.676\tiny$\pm$0.007} & 0.698\tiny$\pm$0.100 & 0.767\tiny$\pm$0.068 & \cellcolor[gray]{0.95}\textbf{2.17\tiny$\pm$1.03} & 0.696\tiny$\pm$0.008 & 0.663\tiny$\pm$0.009 & 0.712\tiny$\pm$0.089 & 0.766\tiny$\pm$0.083 & 3.05\tiny$\pm$0.91 \\[3pt]
    
    16 & 0.656\tiny$\pm$0.010 & 0.128\tiny$\pm$0.005 & 0.731\tiny$\pm$0.095 & 0.169\tiny$\pm$0.050 & \cellcolor[gray]{0.95}\textbf{7.21\tiny$\pm$3.68} & \cellcolor[gray]{0.85}\textbf{0.725\tiny$\pm$0.009} & \cellcolor[gray]{0.85}\textbf{0.183\tiny$\pm$0.013} & \cellcolor[gray]{0.85}\textbf{0.790\tiny$\pm$0.125} & \cellcolor[gray]{0.95}\textbf{0.271\tiny$\pm$0.135} & 9.68\tiny$\pm$3.80 & 0.667\tiny$\pm$0.008 & 0.151\tiny$\pm$0.013 & 0.713\tiny$\pm$0.115 & 0.226\tiny$\pm$0.130 & 14.73\tiny$\pm$6.17 \\[3pt]
    
    17 & \cellcolor[gray]{0.95}\textbf{0.752\tiny$\pm$0.007} & \cellcolor[gray]{0.95}\textbf{0.461\tiny$\pm$0.010} & \cellcolor[gray]{0.95}\textbf{0.803\tiny$\pm$0.105} & \cellcolor[gray]{0.95}\textbf{0.563\tiny$\pm$0.117} & 5.09\tiny$\pm$1.61 & 0.707\tiny$\pm$0.003 & 0.404\tiny$\pm$0.010 & 0.720\tiny$\pm$0.085 & 0.491\tiny$\pm$0.093 & 6.41\tiny$\pm$2.93 & 0.687\tiny$\pm$0.016 & 0.374\tiny$\pm$0.015 & 0.667\tiny$\pm$0.093 & 0.441\tiny$\pm$0.077 & \cellcolor[gray]{0.95}\textbf{5.00\tiny$\pm$3.25} \\[3pt]
    
    18 & \cellcolor[gray]{0.95}\textbf{0.717\tiny$\pm$0.007} & \cellcolor[gray]{0.95}\textbf{0.366\tiny$\pm$0.020} & \cellcolor[gray]{0.95}\textbf{0.722\tiny$\pm$0.090} & \cellcolor[gray]{0.95}\textbf{0.467\tiny$\pm$0.103} & 5.71\tiny$\pm$2.67 & 0.686\tiny$\pm$0.006 & 0.302\tiny$\pm$0.004 & 0.681\tiny$\pm$0.083 & 0.363\tiny$\pm$0.070 & \cellcolor[gray]{0.95}\textbf{5.24\tiny$\pm$2.51} & 0.671\tiny$\pm$0.009 & 0.283\tiny$\pm$0.011 & 0.660\tiny$\pm$0.075 & 0.333\tiny$\pm$0.070 & 7.99\tiny$\pm$3.32 \\[3pt]
    
    19 & \cellcolor[gray]{0.95}\textbf{0.612\tiny$\pm$0.005} & \cellcolor[gray]{0.95}\textbf{0.198\tiny$\pm$0.004} & \cellcolor[gray]{0.95}\textbf{0.623\tiny$\pm$0.067} & \cellcolor[gray]{0.95}\textbf{0.251\tiny$\pm$0.065} & 3.72\tiny$\pm$1.10 & 0.606\tiny$\pm$0.005 & 0.173\tiny$\pm$0.003 & 0.560\tiny$\pm$0.078 & 0.187\tiny$\pm$0.023 & \cellcolor[gray]{0.95}\textbf{3.36\tiny$\pm$2.03} & 0.564\tiny$\pm$0.013 & 0.156\tiny$\pm$0.006 & 0.547\tiny$\pm$0.062 & 0.179\tiny$\pm$0.051 & 9.74\tiny$\pm$4.91 \\[3pt]
    
    20 & \cellcolor[gray]{0.95}\textbf{0.699\tiny$\pm$0.010} & 0.144\tiny$\pm$0.007 & 0.707\tiny$\pm$0.079 & 0.200\tiny$\pm$0.086 & \cellcolor[gray]{0.95}\textbf{4.44\tiny$\pm$1.45} & 0.689\tiny$\pm$0.011 & \cellcolor[gray]{0.95}\textbf{0.159\tiny$\pm$0.009} & \cellcolor[gray]{0.85}\textbf{0.741\tiny$\pm$0.088} & \cellcolor[gray]{0.85}\textbf{0.260\tiny$\pm$0.136} & 5.23\tiny$\pm$2.60 & 0.654\tiny$\pm$0.030 & 0.109\tiny$\pm$0.010 & 0.676\tiny$\pm$0.107 & 0.148\tiny$\pm$0.074 & 6.48\tiny$\pm$3.84 \\[3pt]
    
    21 & \cellcolor[gray]{0.95}\textbf{0.699\tiny$\pm$0.009} & \cellcolor[gray]{0.95}\textbf{0.178\tiny$\pm$0.013} & \cellcolor[gray]{0.95}\textbf{0.653\tiny$\pm$0.088} & \cellcolor[gray]{0.95}\textbf{0.205\tiny$\pm$0.053} & 6.53\tiny$\pm$3.43 & 0.638\tiny$\pm$0.010 & 0.154\tiny$\pm$0.002 & 0.622\tiny$\pm$0.078 & 0.178\tiny$\pm$0.031 & \cellcolor[gray]{0.95}\textbf{4.69\tiny$\pm$1.85} & 0.596\tiny$\pm$0.013 & 0.129\tiny$\pm$0.009 & 0.615\tiny$\pm$0.089 & 0.154\tiny$\pm$0.077 & 7.05\tiny$\pm$1.97 \\[3pt]
    
    22 & \cellcolor[gray]{0.95}\textbf{0.763\tiny$\pm$0.004} & 0.415\tiny$\pm$0.010 & \cellcolor[gray]{0.95}\textbf{0.788\tiny$\pm$0.093} & 0.502\tiny$\pm$0.096 & 3.89\tiny$\pm$0.89 & 0.752\tiny$\pm$0.004 & \cellcolor[gray]{0.95}\textbf{0.417\tiny$\pm$0.012} & 0.786\tiny$\pm$0.086 & \cellcolor[gray]{0.95}\textbf{0.507\tiny$\pm$0.094} & 3.10\tiny$\pm$1.37 & 0.732\tiny$\pm$0.004 & 0.373\tiny$\pm$0.008 & 0.757\tiny$\pm$0.117 & 0.440\tiny$\pm$0.095 & \cellcolor[gray]{0.95}\textbf{2.57\tiny$\pm$0.98} \\[3pt]
    
    23 & \cellcolor[gray]{0.95}\textbf{0.818\tiny$\pm$0.006} & \cellcolor[gray]{0.95}\textbf{0.652\tiny$\pm$0.006} & 0.826\tiny$\pm$0.099 & \cellcolor[gray]{0.95}\textbf{0.757\tiny$\pm$0.075} & \cellcolor[gray]{0.95}\textbf{2.19\tiny$\pm$0.77} & 0.818\tiny$\pm$0.003 & 0.616\tiny$\pm$0.006 & 0.820\tiny$\pm$0.103 & 0.708\tiny$\pm$0.072 & 2.36\tiny$\pm$1.04 & 0.810\tiny$\pm$0.009 & 0.602\tiny$\pm$0.020 & \cellcolor[gray]{0.95}\textbf{0.834\tiny$\pm$0.124} & 0.663\tiny$\pm$0.110 & 3.16\tiny$\pm$1.69 \\[3pt]
    
    24 & 0.774\tiny$\pm$0.005 & 0.510\tiny$\pm$0.010 & 0.815\tiny$\pm$0.090 & 0.621\tiny$\pm$0.103 & 5.60\tiny$\pm$0.95 & \cellcolor[gray]{0.85}\textbf{0.792\tiny$\pm$0.003} & \cellcolor[gray]{0.95}\textbf{0.534\tiny$\pm$0.007} & \cellcolor[gray]{0.95}\textbf{0.829\tiny$\pm$0.091} & \cellcolor[gray]{0.95}\textbf{0.656\tiny$\pm$0.099} & 4.51\tiny$\pm$0.86 & 0.777\tiny$\pm$0.007 & 0.517\tiny$\pm$0.014 & 0.823\tiny$\pm$0.092 & 0.632\tiny$\pm$0.098 & \cellcolor[gray]{0.85}\textbf{2.02\tiny$\pm$0.35} \\[3pt]
    
    25 & 0.826\tiny$\pm$0.004 & 0.556\tiny$\pm$0.004 & 0.886\tiny$\pm$0.115 & \cellcolor[gray]{0.95}\textbf{0.698\tiny$\pm$0.142} & 5.09\tiny$\pm$0.55 & \cellcolor[gray]{0.85}\textbf{0.845\tiny$\pm$0.002} & \cellcolor[gray]{0.95}\textbf{0.557\tiny$\pm$0.008} & \cellcolor[gray]{0.95}\textbf{0.888\tiny$\pm$0.120} & 0.679\tiny$\pm$0.133 & 5.20\tiny$\pm$1.25 & 0.826\tiny$\pm$0.004 & 0.548\tiny$\pm$0.019 & 0.867\tiny$\pm$0.145 & 0.661\tiny$\pm$0.160 & \cellcolor[gray]{0.85}\textbf{2.99\tiny$\pm$1.34} \\[3pt]
    
    \midrule
    
    \textbf{Average} & 0.742\tiny$\pm$0.006 & 0.447\tiny$\pm$0.010 & 0.764\tiny$\pm$0.098 & 0.532\tiny$\pm$0.093 & 4.88\tiny$\pm$1.47 & 0.720\tiny$\pm$0.005 & 0.414\tiny$\pm$0.007 & 0.721\tiny$\pm$0.094 & 0.494\tiny$\pm$0.089 & 5.28\tiny$\pm$2.37 & 0.699\tiny$\pm$0.011 & 0.390\tiny$\pm$0.014 & 0.699\tiny$\pm$0.096 & 0.456\tiny$\pm$0.083 & 5.22\tiny$\pm$2.28 \\[3pt]
    
    \bottomrule
    \end{tabular}
    }
    \label{table:new_eval_multimodals_ups}
\end{table*}

%% file: tables/table_ece_upscaled_multimodals.tex
\begin{table*}[h!]
\caption{
    \textbf{Condition-level Performance of ECE Stratification Across Multimodal Architectures under Loss-Upweighting.}
    Class-stratified calibration metrics under loss upweighting show reduced positive-class ECE for many conditions, but it still remains dominant across architectures.
    No single multimodal model consistently achieves the lowest ECE across conditions, indicating that loss upweighting mitigates but does not resolve class-dependent calibration errors.
    \footnotesize{(Dark-bold: $p<0.05$, Wilcoxon signed-rank test, 5 seeds; Light-bold: highest mean, not significant)}
    }
    \centering
    \resizebox{\linewidth}{!}{
    \begin{tabular}{c|ccc|ccc|cccc}
    \multirow{2}{*}[-1.75em]{\textbf{\makecell{Clinical \\ Condition}}} 

    & \multicolumn{3}{c}{\textbf{MedFuse (Loss-Upweighting)}}
    & \multicolumn{3}{c}{\textbf{DrFuse (Loss-Upweighting)}}
    & \multicolumn{3}{c}{\textbf{MeTra (Loss-Upweighting)}}\\
    \midrule
    \addlinespace[10pt]
    & \textbf{$\widehat{\mathbf{ECE}} \downarrow$}
    & \textbf{$\widehat{\mathbf{ECE}}_{c=1} \downarrow$}
    & \textbf{$\widehat{\mathbf{ECE}}_{c=0} \downarrow$}
    
     & \textbf{$\widehat{\mathbf{ECE}} \downarrow$}
    & \textbf{$\widehat{\mathbf{ECE}}_{c=1} \downarrow$}
    & \textbf{$\widehat{\mathbf{ECE}}_{c=0} \downarrow$}
    
     & \textbf{$\widehat{\mathbf{ECE}} \downarrow$}
    & \textbf{$\widehat{\mathbf{ECE}}_{c=1} \downarrow$}
    & \textbf{$\widehat{\mathbf{ECE}}_{c=0} \downarrow$}
    \\
    \midrule

    1 & \cellcolor[gray]{0.95}\textbf{3.09\tiny$\pm$0.86} & \cellcolor[gray]{0.95}\textbf{3.81\tiny$\pm$1.13} & \cellcolor[gray]{0.95}\textbf{5.01\tiny$\pm$1.96} & 4.58\tiny$\pm$0.98 & 4.57\tiny$\pm$0.84 & 6.32\tiny$\pm$1.96 & 3.35\tiny$\pm$1.48 & 7.19\tiny$\pm$4.48 & 7.36\tiny$\pm$2.53 \\[3pt]
    
    2 & \cellcolor[gray]{0.95}\textbf{2.38\tiny$\pm$0.66} & 10.82\tiny$\pm$1.46 & \cellcolor[gray]{0.95}\textbf{2.10\tiny$\pm$0.76} & 4.44\tiny$\pm$1.50 & 9.75\tiny$\pm$2.09 & 4.69\tiny$\pm$1.66 & 3.79\tiny$\pm$0.79 & \cellcolor[gray]{0.95}\textbf{5.78\tiny$\pm$1.66} & 4.41\tiny$\pm$0.85 \\[3pt]
    
    3 & \cellcolor[gray]{0.95}\textbf{9.58\tiny$\pm$2.12} & 12.14\tiny$\pm$1.93 & \cellcolor[gray]{0.95}\textbf{11.33\tiny$\pm$2.29} & 13.34\tiny$\pm$6.55 & \cellcolor[gray]{0.95}\textbf{11.71\tiny$\pm$2.50} & 15.62\tiny$\pm$7.08 & 10.09\tiny$\pm$3.99 & 18.20\tiny$\pm$13.24 & 12.53\tiny$\pm$5.48 \\[3pt]
    
    4 & 2.89\tiny$\pm$1.26 & 7.52\tiny$\pm$2.45 & \cellcolor[gray]{0.95}\textbf{5.32\tiny$\pm$5.21} & 4.58\tiny$\pm$2.03 & \cellcolor[gray]{0.95}\textbf{7.31\tiny$\pm$2.60} & 11.83\tiny$\pm$4.69 & \cellcolor[gray]{0.95}\textbf{1.93\tiny$\pm$0.65} & 11.11\tiny$\pm$5.16 & 8.07\tiny$\pm$3.52 \\[3pt]
    
    5 & \cellcolor[gray]{0.95}\textbf{4.60\tiny$\pm$1.52} & \cellcolor[gray]{0.95}\textbf{5.05\tiny$\pm$2.19} & \cellcolor[gray]{0.95}\textbf{5.40\tiny$\pm$2.97} & 5.90\tiny$\pm$2.72 & 5.60\tiny$\pm$2.70 & 8.21\tiny$\pm$4.41 & 4.98\tiny$\pm$3.28 & 12.78\tiny$\pm$12.05 & 9.92\tiny$\pm$7.48 \\[3pt]
    
    6 & 4.78\tiny$\pm$2.64 & 13.89\tiny$\pm$7.07 & 5.96\tiny$\pm$3.21 & 4.36\tiny$\pm$2.76 & \cellcolor[gray]{0.95}\textbf{4.41\tiny$\pm$1.88} & \cellcolor[gray]{0.95}\textbf{5.47\tiny$\pm$3.56} & \cellcolor[gray]{0.95}\textbf{3.57\tiny$\pm$1.35} & 11.84\tiny$\pm$3.42 & 5.74\tiny$\pm$2.10 \\[3pt]
    
    7 & 5.90\tiny$\pm$1.22 & 9.79\tiny$\pm$1.85 & 10.40\tiny$\pm$1.49 & \cellcolor[gray]{0.95}\textbf{2.46\tiny$\pm$1.79} & \cellcolor[gray]{0.85}\textbf{4.01\tiny$\pm$1.00} & \cellcolor[gray]{0.95}\textbf{3.26\tiny$\pm$2.91} & 2.96\tiny$\pm$0.57 & 8.72\tiny$\pm$3.99 & 5.77\tiny$\pm$2.06 \\[3pt]
    
    8 & \cellcolor[gray]{0.95}\textbf{4.61\tiny$\pm$0.84} & 20.18\tiny$\pm$4.92 & \cellcolor[gray]{0.95}\textbf{6.43\tiny$\pm$1.51} & 8.29\tiny$\pm$2.82 & \cellcolor[gray]{0.95}\textbf{9.40\tiny$\pm$5.07} & 9.16\tiny$\pm$2.14 & 6.13\tiny$\pm$3.48 & 14.85\tiny$\pm$11.31 & 8.36\tiny$\pm$4.59 \\[3pt]
    
    9 & 5.97\tiny$\pm$2.62 & 6.67\tiny$\pm$1.70 & \cellcolor[gray]{0.95}\textbf{7.93\tiny$\pm$5.47} & 8.07\tiny$\pm$2.26 & \cellcolor[gray]{0.95}\textbf{5.75\tiny$\pm$1.75} & 12.77\tiny$\pm$4.47 & \cellcolor[gray]{0.95}\textbf{4.83\tiny$\pm$3.15} & 11.04\tiny$\pm$5.53 & 10.03\tiny$\pm$4.27 \\[3pt]
    
    10 & 5.86\tiny$\pm$2.01 & \cellcolor[gray]{0.95}\textbf{5.35\tiny$\pm$3.38} & \cellcolor[gray]{0.95}\textbf{7.46\tiny$\pm$4.30} & 4.17\tiny$\pm$3.15 & 6.17\tiny$\pm$2.38 & 8.70\tiny$\pm$5.61 & \cellcolor[gray]{0.95}\textbf{2.16\tiny$\pm$1.13} & 13.40\tiny$\pm$14.01 & 9.35\tiny$\pm$5.23 \\[3pt]
    
    11 & \cellcolor[gray]{0.95}\textbf{4.62\tiny$\pm$0.72} & \cellcolor[gray]{0.95}\textbf{4.86\tiny$\pm$1.49} & \cellcolor[gray]{0.95}\textbf{4.98\tiny$\pm$0.82} & 6.14\tiny$\pm$2.96 & 6.17\tiny$\pm$1.61 & 7.17\tiny$\pm$3.73 & 6.68\tiny$\pm$2.72 & 8.75\tiny$\pm$3.53 & 8.44\tiny$\pm$3.56 \\[3pt]
    
    12 & \cellcolor[gray]{0.95}\textbf{6.83\tiny$\pm$1.26} & \cellcolor[gray]{0.95}\textbf{8.12\tiny$\pm$2.16} & \cellcolor[gray]{0.95}\textbf{10.47\tiny$\pm$2.69} & 7.58\tiny$\pm$4.02 & 11.77\tiny$\pm$3.28 & 12.57\tiny$\pm$5.78 & 9.57\tiny$\pm$3.54 & 17.23\tiny$\pm$2.96 & 16.67\tiny$\pm$4.44 \\[3pt]
    
    13 & 5.94\tiny$\pm$0.46 & \cellcolor[gray]{0.95}\textbf{7.16\tiny$\pm$1.24} & \cellcolor[gray]{0.95}\textbf{10.32\tiny$\pm$3.75} & 3.74\tiny$\pm$3.10 & 7.17\tiny$\pm$3.51 & 10.62\tiny$\pm$7.21 & \cellcolor[gray]{0.95}\textbf{3.22\tiny$\pm$1.88} & 14.21\tiny$\pm$11.52 & 13.23\tiny$\pm$6.32 \\[3pt]
    
    14 & \cellcolor[gray]{0.95}\textbf{2.25\tiny$\pm$0.77} & \cellcolor[gray]{0.95}\textbf{7.84\tiny$\pm$3.34} & \cellcolor[gray]{0.95}\textbf{6.50\tiny$\pm$4.92} & 2.47\tiny$\pm$1.43 & 9.74\tiny$\pm$4.32 & 8.68\tiny$\pm$7.05 & 2.47\tiny$\pm$0.35 & 9.95\tiny$\pm$2.70 & 11.95\tiny$\pm$2.34 \\[3pt]
    
    15 & 3.22\tiny$\pm$0.62 & \cellcolor[gray]{0.95}\textbf{4.22\tiny$\pm$1.38} & 4.33\tiny$\pm$1.55 & \cellcolor[gray]{0.95}\textbf{2.17\tiny$\pm$1.03} & 4.51\tiny$\pm$2.29 & \cellcolor[gray]{0.95}\textbf{3.34\tiny$\pm$1.86} & 3.05\tiny$\pm$0.91 & 5.77\tiny$\pm$1.29 & 5.72\tiny$\pm$1.70 \\[3pt]
    
    16 & \cellcolor[gray]{0.95}\textbf{7.21\tiny$\pm$3.68} & 12.76\tiny$\pm$6.14 & \cellcolor[gray]{0.95}\textbf{7.65\tiny$\pm$3.81} & 9.68\tiny$\pm$3.80 & \cellcolor[gray]{0.95}\textbf{7.86\tiny$\pm$2.60} & 10.48\tiny$\pm$4.14 & 14.73\tiny$\pm$6.17 & 13.94\tiny$\pm$3.39 & 16.54\tiny$\pm$6.96 \\[3pt]
    
    17 & 5.09\tiny$\pm$1.61 & 6.16\tiny$\pm$2.69 & \cellcolor[gray]{0.95}\textbf{5.87\tiny$\pm$3.00} & 6.41\tiny$\pm$2.93 & \cellcolor[gray]{0.95}\textbf{5.70\tiny$\pm$2.92} & 8.56\tiny$\pm$4.13 & \cellcolor[gray]{0.95}\textbf{5.00\tiny$\pm$3.25} & 11.68\tiny$\pm$10.50 & 9.28\tiny$\pm$6.46 \\[3pt]
    
    18 & 5.71\tiny$\pm$2.67 & 8.68\tiny$\pm$3.09 & 6.80\tiny$\pm$3.54 & \cellcolor[gray]{0.95}\textbf{5.24\tiny$\pm$2.51} & \cellcolor[gray]{0.95}\textbf{6.50\tiny$\pm$3.35} & \cellcolor[gray]{0.95}\textbf{6.22\tiny$\pm$3.20} & 7.99\tiny$\pm$3.32 & 9.92\tiny$\pm$4.03 & 11.24\tiny$\pm$5.20 \\[3pt]
    
    19 & 3.72\tiny$\pm$1.10 & \cellcolor[gray]{0.95}\textbf{6.00\tiny$\pm$2.71} & \cellcolor[gray]{0.95}\textbf{3.86\tiny$\pm$1.39} & \cellcolor[gray]{0.95}\textbf{3.36\tiny$\pm$2.03} & 7.05\tiny$\pm$2.53 & 4.39\tiny$\pm$2.81 & 9.74\tiny$\pm$4.91 & 12.10\tiny$\pm$13.11 & 12.23\tiny$\pm$7.18 \\[3pt]
    
    20 & \cellcolor[gray]{0.95}\textbf{4.44\tiny$\pm$1.45} & \cellcolor[gray]{0.95}\textbf{18.17\tiny$\pm$1.09} & \cellcolor[gray]{0.95}\textbf{5.18\tiny$\pm$1.68} & 5.23\tiny$\pm$2.60 & 24.25\tiny$\pm$2.43 & 6.09\tiny$\pm$3.30 & 6.48\tiny$\pm$3.84 & 20.08\tiny$\pm$8.64 & 7.84\tiny$\pm$3.91 \\[3pt]
    
    21 & 6.53\tiny$\pm$3.43 & \cellcolor[gray]{0.95}\textbf{7.57\tiny$\pm$3.75} & 7.36\tiny$\pm$4.02 & \cellcolor[gray]{0.95}\textbf{4.69\tiny$\pm$1.85} & 12.96\tiny$\pm$2.23 & \cellcolor[gray]{0.95}\textbf{5.69\tiny$\pm$2.19} & 7.05\tiny$\pm$1.97 & 19.47\tiny$\pm$6.77 & 9.59\tiny$\pm$2.77 \\[3pt]
    
    22 & 3.89\tiny$\pm$0.89 & \cellcolor[gray]{0.95}\textbf{5.27\tiny$\pm$1.06} & 5.56\tiny$\pm$1.54 & 3.10\tiny$\pm$1.37 & 5.57\tiny$\pm$1.14 & 4.29\tiny$\pm$1.37 & \cellcolor[gray]{0.95}\textbf{2.57\tiny$\pm$0.98} & 6.21\tiny$\pm$0.90 & \cellcolor[gray]{0.95}\textbf{3.95\tiny$\pm$1.18} \\[3pt]
    
    23 & \cellcolor[gray]{0.95}\textbf{2.19\tiny$\pm$0.77} & 6.05\tiny$\pm$0.85 & \cellcolor[gray]{0.95}\textbf{3.27\tiny$\pm$1.91} & 2.36\tiny$\pm$1.04 & \cellcolor[gray]{0.95}\textbf{3.67\tiny$\pm$0.81} & 3.32\tiny$\pm$1.17 & 3.16\tiny$\pm$1.69 & 7.09\tiny$\pm$2.35 & 6.23\tiny$\pm$3.19 \\[3pt]
    
    24 & 5.60\tiny$\pm$0.95 & 5.24\tiny$\pm$0.89 & 6.70\tiny$\pm$1.13 & 4.51\tiny$\pm$0.86 & 5.25\tiny$\pm$0.73 & 6.30\tiny$\pm$1.23 & \cellcolor[gray]{0.85}\textbf{2.02\tiny$\pm$0.35} & \cellcolor[gray]{0.95}\textbf{5.23\tiny$\pm$1.63} & \cellcolor[gray]{0.85}\textbf{2.87\tiny$\pm$0.69} \\[3pt]
    
    25 & 5.09\tiny$\pm$0.55 & 5.63\tiny$\pm$1.12 & 5.11\tiny$\pm$0.80 & 5.20\tiny$\pm$1.25 & \cellcolor[gray]{0.95}\textbf{4.11\tiny$\pm$1.67} & 5.83\tiny$\pm$1.52 & \cellcolor[gray]{0.85}\textbf{2.99\tiny$\pm$1.34} & 6.58\tiny$\pm$2.98 & \cellcolor[gray]{0.95}\textbf{3.94\tiny$\pm$1.55} \\[3pt]
    
    \midrule
    
    \textbf{Average} & 4.88\tiny$\pm$1.47 & 8.36\tiny$\pm$2.44 & 6.45\tiny$\pm$2.63 & 5.28\tiny$\pm$2.37 & 7.64\tiny$\pm$2.33 & 7.58\tiny$\pm$3.57 & 5.22\tiny$\pm$2.28 & 11.32\tiny$\pm$6.05 & 8.85\tiny$\pm$3.82 \\[3pt]

    \bottomrule
    \end{tabular}
    }
    \label{table:new_ece_multimodals_ups}
\end{table*}

%% file: tables/table_ece_unimodals_binning.tex
\setlength{\tabcolsep}{6pt}
\begin{table*}[h!]
\caption{
    \textbf{Robustness of ECE Estimates Across Binning Strategies for Unimodal Models.}
    Aggregate and class-stratified ECE are reported for unimodal baselines and MedFuse across different numbers of bins.
    Positive-class calibration error remains substantially higher than negative-class calibration error across binning choices, indicating that the observed class-dependent miscalibration is not driven by a single ECE bin configuration.
}
    \centering
    \scriptsize
    \begin{adjustbox}{max width=\linewidth}
    \begin{tabular}{c|ccc|ccc|cccc}
    \multirow{2}{*}[-1.5em]{\textbf{Bins}} 
    & \multicolumn{3}{c}{\textbf{EHR}}
    & \multicolumn{3}{c}{\textbf{CXR}}
    & \multicolumn{3}{c}{\textbf{MedFuse}}\\
    \midrule

    & \textbf{$\widehat{\mathbf{ECE}} \downarrow$}
    & \textbf{$\widehat{\mathbf{ECE}}_{c=1} \downarrow $}
    & \textbf{$\widehat{\mathbf{ECE}}_{c=0} \downarrow$}

    & \textbf{$\widehat{\mathbf{ECE}} \downarrow$}
    & \textbf{$\widehat{\mathbf{ECE}}_{c=1} \downarrow$}
    & \textbf{$\widehat{\mathbf{ECE}}_{c=0} \downarrow$}
    
    & \textbf{$\widehat{\mathbf{ECE}} \downarrow$}
    & \textbf{$\widehat{\mathbf{ECE}}_{c=1} \downarrow$}
    & \textbf{$\widehat{\mathbf{ECE}}_{c=0} \downarrow$}
    \\[2pt]
    \midrule

    5 & 1.07\tiny{$\pm$0.30} & 60.35\tiny{$\pm$1.08} & 14.26\tiny{$\pm$0.48} & 9.26\tiny{$\pm$0.88} & 54.64\tiny{$\pm$5.64} & 5.68\tiny{$\pm$1.75} & 1.85\tiny{$\pm$0.78} & 49.19\tiny{$\pm$3.46} & 8.85\tiny{$\pm$0.98} \\[3pt]

    10 &  1.42\tiny{$\pm$0.34} &  60.35\tiny{$\pm$1.08} &  14.27\tiny{$\pm$0.48} &  9.41\tiny{$\pm$0.83} &  54.67\tiny{$\pm$5.63} &  5.74\tiny{$\pm$1.71} &  2.08\tiny{$\pm$0.73} &  49.20\tiny{$\pm$3.47} &  8.86\tiny{$\pm$0.97} \\[3pt]
    
    15 & 1.64\tiny{$\pm$0.32} & 60.70\tiny{$\pm$1.14} & 14.39\tiny{$\pm$0.44} & 9.55\tiny{$\pm$0.81} & 54.97\tiny{$\pm$5.65} & 5.87\tiny{$\pm$1.67} & 2.30\tiny{$\pm$0.71} & 49.48\tiny{$\pm$3.45} & 8.94\tiny{$\pm$0.95} \\[3pt]
    
    20 & 1.84\tiny{$\pm$0.40} & 60.81\tiny{$\pm$1.11} & 14.42\tiny{$\pm$0.45} & 9.59\tiny{$\pm$0.83} & 54.95\tiny{$\pm$5.66} & 5.93\tiny{$\pm$1.68} & 2.41\tiny{$\pm$0.70} & 49.39\tiny{$\pm$3.46} & 8.97\tiny{$\pm$0.95} \\[3pt]
    
    50 & 2.81\tiny{$\pm$0.42} & 61.27\tiny{$\pm$1.14} & 14.61\tiny{$\pm$0.45} & 10.01\tiny{$\pm$0.77} & 55.59\tiny{$\pm$5.63} & 6.39\tiny{$\pm$1.54} & 3.26\tiny{$\pm$0.62} & 49.83\tiny{$\pm$3.45} & 9.17\tiny{$\pm$0.89} \\[3pt]

    \bottomrule
    \end{tabular}
    \end{adjustbox}
    \label{table:ece_unimodals_binning}
\end{table*}

%% file: tables/table_ece_multimodals_binning.tex
\setlength{\tabcolsep}{6pt}
\begin{table*}[h!]
\caption{
    \textbf{Robustness of ECE Estimates Across Binning Strategies for Multimodal Models.}
    Aggregate and class-stratified ECE are reported for multimodal architectures across different numbers of bins.
    Across MedFuse, DrFuse, and MeTra, positive-class ECE remains consistently larger than negative-class ECE, showing that class-dependent miscalibration persists across binning strategies and model families.
    }
    \centering
    \scriptsize
    \begin{adjustbox}{max width=\linewidth}
    \begin{tabular}{c|ccc|ccc|cccc}
    \multirow{2}{*}[-1.5em]{\textbf{Bins}} 
    & \multicolumn{3}{c}{\textbf{MedFuse}}
    & \multicolumn{3}{c}{\textbf{DrFuse}}
    & \multicolumn{3}{c}{\textbf{MeTra}}\\
    \midrule

    & \textbf{$\widehat{\mathbf{ECE}} \downarrow$}
    & \textbf{$\widehat{\mathbf{ECE}}_{c=1} \downarrow $}
    & \textbf{$\widehat{\mathbf{ECE}}_{c=0} \downarrow$}

    & \textbf{$\widehat{\mathbf{ECE}} \downarrow$}
    & \textbf{$\widehat{\mathbf{ECE}}_{c=1} \downarrow$}
    & \textbf{$\widehat{\mathbf{ECE}}_{c=0} \downarrow$}
    
    & \textbf{$\widehat{\mathbf{ECE}} \downarrow$}
    & \textbf{$\widehat{\mathbf{ECE}}_{c=1} \downarrow$}
    & \textbf{$\widehat{\mathbf{ECE}}_{c=0} \downarrow$}
    \\[2pt]
    \midrule

    5 & 1.85\tiny{$\pm$0.78} & 49.19\tiny{$\pm$3.46} & 8.85\tiny{$\pm$0.98} & 2.42\tiny{$\pm$1.21} & 51.95\tiny{$\pm$6.04} & 9.64\tiny{$\pm$2.26} & 2.08\tiny{$\pm$0.90} & 55.47\tiny{$\pm$5.96} & 11.58\tiny{$\pm$1.84} \\[3pt]

     10 &  2.08\tiny{$\pm$0.73} &  49.20\tiny{$\pm$3.47} &  8.86\tiny{$\pm$0.97} &  2.54\tiny{$\pm$1.15} &  51.98\tiny{$\pm$6.04} &  9.67\tiny{$\pm$2.23} &  2.32\tiny{$\pm$0.83} &  55.52\tiny{$\pm$5.94} &  11.61\tiny{$\pm$1.82}\\[3pt]
    
    15 & 2.30\tiny{$\pm$0.71} & 49.48\tiny{$\pm$3.45} & 8.94\tiny{$\pm$0.95} & 2.76\tiny{$\pm$1.11} & 52.18\tiny{$\pm$5.98} & 9.79\tiny{$\pm$2.22} & 2.55\tiny{$\pm$0.78} & 55.74\tiny{$\pm$5.85} & 11.71\tiny{$\pm$1.77} \\[3pt]
    
    20 & 2.41\tiny{$\pm$0.70} & 49.39\tiny{$\pm$3.46} & 8.97\tiny{$\pm$0.95} & 2.87\tiny{$\pm$1.07} & 52.13\tiny{$\pm$6.00} & 9.79\tiny{$\pm$2.19} & 2.67\tiny{$\pm$0.74} & 55.72\tiny{$\pm$5.84} & 11.69\tiny{$\pm$1.76} \\[3pt]
    
    50 & 3.26\tiny{$\pm$0.62} & 49.83\tiny{$\pm$3.45} & 9.17\tiny{$\pm$0.89} & 3.62\tiny{$\pm$0.96} & 52.69\tiny{$\pm$5.86} & 10.08\tiny{$\pm$2.09} & 3.47\tiny{$\pm$0.63} & 56.10\tiny{$\pm$5.61} & 11.87\tiny{$\pm$1.71} \\[3pt]

    \bottomrule
    \end{tabular}
    \end{adjustbox}
    \label{table:ece_multimodals_binning}
\end{table*}

%% file: tables/table_aece_unimodals.tex
\begin{table*}[h!]
\caption{
    \textbf{Adaptive ECE for Unimodal Models.}
    Aggregate and class-stratified Adaptive Expected Calibration Error (AECE) are reported for unimodal baselines.
    Positive-class AECE remains substantially higher than negative-class AECE, confirming that minority-class miscalibration persists when calibration error is estimated using adaptive equal-frequency bins.
    \footnotesize{(Dark-bold: $p<0.05$, Wilcoxon signed-rank test, 5 seeds; Light-bold: highest mean, not significant)}
}
    \centering
    \resizebox{\linewidth}{!}{
    \begin{tabular}{c|ccc|ccc|cccc}
    \multirow{2}{*}[-1.75em]{\textbf{\makecell{Clinical \\ Condition}}} 

    & \multicolumn{3}{c}{\textbf{EHR}}
    & \multicolumn{3}{c}{\textbf{CXR}}
    & \multicolumn{3}{c}{\textbf{MedFuse}}\\
    \midrule
    \addlinespace[10pt]
    & \textbf{$\widehat{\mathbf{AECE}} \downarrow$}
    & \textbf{$\widehat{\mathbf{AECE}}_{c=1} \downarrow$}
    & \textbf{$\widehat{\mathbf{AECE}}_{c=0} \downarrow$}
    
     & \textbf{$\widehat{\mathbf{AECE}} \downarrow$}
    & \textbf{$\widehat{\mathbf{AECE}}_{c=1} \downarrow$}
    & \textbf{$\widehat{\mathbf{AECE}}_{c=0} \downarrow$}
    
     & \textbf{$\widehat{\mathbf{AECE}} \downarrow$}
    & \textbf{$\widehat{\mathbf{AECE}}_{c=1} \downarrow$}
    & \textbf{$\widehat{\mathbf{AECE}}_{c=0} \downarrow$}
    \\
    \midrule

    1 & \cellcolor[gray]{0.95}\textbf{2.32\tiny$\pm$0.48} & 57.80\tiny$\pm$0.36 & 27.38\tiny$\pm$0.09 & 14.39\tiny$\pm$1.15 & 56.87\tiny$\pm$3.22 & 29.87\tiny$\pm$3.00 & 3.22\tiny$\pm$0.91 & \cellcolor[gray]{0.95}\textbf{54.09\tiny$\pm$2.26} & \cellcolor[gray]{0.95}\textbf{25.83\tiny$\pm$2.03} \\[3pt]

    2 & \cellcolor[gray]{0.95}\textbf{1.02\tiny$\pm$0.21} & \cellcolor[gray]{0.85}\textbf{66.59\tiny$\pm$0.37} & 6.23\tiny$\pm$0.14 & 5.87\tiny$\pm$0.47 & 86.92\tiny$\pm$2.62 & 6.84\tiny$\pm$1.62 & 1.29\tiny$\pm$0.26 & 67.96\tiny$\pm$1.40 & \cellcolor[gray]{0.95}\textbf{5.40\tiny$\pm$0.18} \\[3pt]
    
    3 & \cellcolor[gray]{0.95}\textbf{1.37\tiny$\pm$0.12} & 88.27\tiny$\pm$0.33 & 8.22\tiny$\pm$0.06 & 6.75\tiny$\pm$1.33 & \cellcolor[gray]{0.95}\textbf{85.30\tiny$\pm$3.50} & 8.38\tiny$\pm$2.50 & 1.46\tiny$\pm$0.30 & 86.93\tiny$\pm$0.14 & \cellcolor[gray]{0.95}\textbf{8.05\tiny$\pm$0.25} \\[3pt]
    
    4 & \cellcolor[gray]{0.85}\textbf{2.37\tiny$\pm$0.50} & 60.63\tiny$\pm$0.13 & 36.44\tiny$\pm$0.12 & 14.13\tiny$\pm$1.07 & \cellcolor[gray]{0.95}\textbf{52.81\tiny$\pm$4.49} & 32.66\tiny$\pm$4.52 & 3.93\tiny$\pm$1.67 & 55.81\tiny$\pm$3.90 & \cellcolor[gray]{0.95}\textbf{32.44\tiny$\pm$3.51} \\[3pt]
    
    5 & \cellcolor[gray]{0.95}\textbf{2.65\tiny$\pm$0.37} & 70.95\tiny$\pm$0.28 & 21.08\tiny$\pm$0.07 & 17.08\tiny$\pm$0.70 & \cellcolor[gray]{0.95}\textbf{59.96\tiny$\pm$1.82} & 22.65\tiny$\pm$1.99 & 3.64\tiny$\pm$1.62 & 61.94\tiny$\pm$4.35 & \cellcolor[gray]{0.95}\textbf{18.95\tiny$\pm$3.07} \\[3pt]
    
    6 & \cellcolor[gray]{0.95}\textbf{2.58\tiny$\pm$0.14} & 81.50\tiny$\pm$0.22 & 16.00\tiny$\pm$0.07 & 8.93\tiny$\pm$1.03 & 74.71\tiny$\pm$9.10 & \cellcolor[gray]{0.95}\textbf{11.19\tiny$\pm$4.40} & 3.30\tiny$\pm$0.47 & \cellcolor[gray]{0.95}\textbf{72.08\tiny$\pm$2.20} & 13.20\tiny$\pm$1.03 \\[3pt]
    
    7 & \cellcolor[gray]{0.95}\textbf{1.99\tiny$\pm$0.25} & 73.47\tiny$\pm$0.43 & 20.83\tiny$\pm$0.08 & 11.31\tiny$\pm$0.89 & 75.57\tiny$\pm$2.16 & \cellcolor[gray]{0.85}\textbf{17.34\tiny$\pm$1.60} & 2.10\tiny$\pm$0.35 & \cellcolor[gray]{0.95}\textbf{72.23\tiny$\pm$0.49} & 20.56\tiny$\pm$0.37 \\[3pt]
    
    8 & \cellcolor[gray]{0.85}\textbf{1.45\tiny$\pm$0.24} & 87.20\tiny$\pm$0.28 & 10.42\tiny$\pm$0.04 & 4.95\tiny$\pm$1.48 & \cellcolor[gray]{0.95}\textbf{52.20\tiny$\pm$7.29} & 6.55\tiny$\pm$3.01 & 2.85\tiny$\pm$0.49 & 54.39\tiny$\pm$3.81 & \cellcolor[gray]{0.95}\textbf{6.16\tiny$\pm$1.40} \\[3pt]
    
    9 & \cellcolor[gray]{0.85}\textbf{3.22\tiny$\pm$0.52} & 61.89\tiny$\pm$0.34 & 27.76\tiny$\pm$0.06 & 13.63\tiny$\pm$3.29 & \cellcolor[gray]{0.85}\textbf{41.85\tiny$\pm$5.80} & 28.29\tiny$\pm$5.00 & 5.63\tiny$\pm$1.61 & 50.90\tiny$\pm$5.67 & \cellcolor[gray]{0.95}\textbf{21.83\tiny$\pm$4.31} \\[3pt]
    
    10 & \cellcolor[gray]{0.95}\textbf{2.97\tiny$\pm$0.25} & 62.15\tiny$\pm$0.19 & 27.59\tiny$\pm$0.12 & 12.29\tiny$\pm$1.25 & 50.81\tiny$\pm$4.00 & 24.76\tiny$\pm$4.45 & 4.60\tiny$\pm$1.31 & \cellcolor[gray]{0.95}\textbf{50.67\tiny$\pm$3.65} & \cellcolor[gray]{0.95}\textbf{24.37\tiny$\pm$3.03} \\[3pt]
    
    11 & 2.41\tiny$\pm$0.61 & 80.37\tiny$\pm$1.11 & 10.92\tiny$\pm$0.15 & 9.29\tiny$\pm$0.68 & 82.04\tiny$\pm$4.22 & 11.91\tiny$\pm$2.77 & \cellcolor[gray]{0.95}\textbf{1.44\tiny$\pm$0.33} & \cellcolor[gray]{0.95}\textbf{65.92\tiny$\pm$1.70} & \cellcolor[gray]{0.95}\textbf{9.03\tiny$\pm$0.38} \\[3pt]
    
    12 & \cellcolor[gray]{0.95}\textbf{2.05\tiny$\pm$0.62} & 77.14\tiny$\pm$0.38 & 19.78\tiny$\pm$0.07 & 12.72\tiny$\pm$1.51 & 73.26\tiny$\pm$4.44 & 21.51\tiny$\pm$4.13 & 2.63\tiny$\pm$0.40 & \cellcolor[gray]{0.95}\textbf{71.76\tiny$\pm$1.61} & \cellcolor[gray]{0.95}\textbf{18.97\tiny$\pm$1.05} \\[3pt]
    
    13 & \cellcolor[gray]{0.85}\textbf{2.70\tiny$\pm$0.27} & 54.87\tiny$\pm$0.23 & 38.19\tiny$\pm$0.07 & 16.87\tiny$\pm$0.76 & \cellcolor[gray]{0.95}\textbf{42.87\tiny$\pm$6.30} & 43.89\tiny$\pm$6.23 & 4.51\tiny$\pm$0.75 & 49.43\tiny$\pm$2.11 & \cellcolor[gray]{0.95}\textbf{36.43\tiny$\pm$2.12} \\[3pt]
    
    14 & \cellcolor[gray]{0.95}\textbf{3.42\tiny$\pm$0.36} & 54.28\tiny$\pm$0.39 & 43.48\tiny$\pm$0.08 & 19.16\tiny$\pm$1.41 & \cellcolor[gray]{0.95}\textbf{49.45\tiny$\pm$5.90} & \cellcolor[gray]{0.95}\textbf{39.29\tiny$\pm$6.64} & 3.75\tiny$\pm$0.81 & 50.60\tiny$\pm$2.22 & 39.56\tiny$\pm$2.06 \\[3pt]
    
    15 & \cellcolor[gray]{0.85}\textbf{2.37\tiny$\pm$0.34} & 47.74\tiny$\pm$0.15 & 39.49\tiny$\pm$0.24 & 16.95\tiny$\pm$1.31 & 52.02\tiny$\pm$1.91 & 38.58\tiny$\pm$1.96 & 3.67\tiny$\pm$0.92 & \cellcolor[gray]{0.95}\textbf{46.28\tiny$\pm$1.19} & \cellcolor[gray]{0.95}\textbf{36.83\tiny$\pm$1.55} \\[3pt]
    
    16 & \cellcolor[gray]{0.95}\textbf{1.69\tiny$\pm$0.31} & 91.64\tiny$\pm$0.15 & 6.89\tiny$\pm$0.09 & 5.59\tiny$\pm$0.76 & \cellcolor[gray]{0.95}\textbf{89.56\tiny$\pm$2.11} & \cellcolor[gray]{0.95}\textbf{6.79\tiny$\pm$1.40} & 1.73\tiny$\pm$0.34 & 89.64\tiny$\pm$0.80 & 7.21\tiny$\pm$0.42 \\[3pt]
    
    17 & \cellcolor[gray]{0.95}\textbf{2.24\tiny$\pm$0.33} & 73.82\tiny$\pm$0.26 & 19.26\tiny$\pm$0.09 & 15.74\tiny$\pm$0.32 & \cellcolor[gray]{0.95}\textbf{64.14\tiny$\pm$0.99} & 19.92\tiny$\pm$0.97 & 3.86\tiny$\pm$1.29 & 64.76\tiny$\pm$4.25 & \cellcolor[gray]{0.95}\textbf{17.59\tiny$\pm$2.93} \\[3pt]
    
    18 & \cellcolor[gray]{0.95}\textbf{1.92\tiny$\pm$0.37} & 80.92\tiny$\pm$0.15 & 15.41\tiny$\pm$0.14 & 9.63\tiny$\pm$1.13 & \cellcolor[gray]{0.95}\textbf{70.76\tiny$\pm$3.99} & \cellcolor[gray]{0.95}\textbf{14.27\tiny$\pm$3.05} & 3.08\tiny$\pm$0.99 & 72.07\tiny$\pm$2.49 & 15.49\tiny$\pm$0.95 \\[3pt]
    
    19 & \cellcolor[gray]{0.85}\textbf{1.44\tiny$\pm$0.32} & 86.08\tiny$\pm$0.27 & 12.65\tiny$\pm$0.12 & 9.99\tiny$\pm$1.09 & \cellcolor[gray]{0.95}\textbf{83.60\tiny$\pm$3.35} & 13.59\tiny$\pm$2.87 & 2.11\tiny$\pm$0.15 & 86.71\tiny$\pm$0.43 & \cellcolor[gray]{0.95}\textbf{11.15\tiny$\pm$0.47} \\[3pt]
    
    20 & 1.68\tiny$\pm$0.33 & 91.94\tiny$\pm$0.23 & 6.23\tiny$\pm$0.09 & 4.52\tiny$\pm$0.68 & \cellcolor[gray]{0.85}\textbf{89.21\tiny$\pm$1.18} & \cellcolor[gray]{0.95}\textbf{4.93\tiny$\pm$1.02} & \cellcolor[gray]{0.95}\textbf{1.34\tiny$\pm$0.28} & 92.06\tiny$\pm$0.39 & 5.53\tiny$\pm$0.33 \\[3pt]
    
    21 & 2.37\tiny$\pm$0.54 & 88.61\tiny$\pm$0.18 & 9.83\tiny$\pm$0.14 & 6.52\tiny$\pm$0.41 & \cellcolor[gray]{0.95}\textbf{86.20\tiny$\pm$3.27} & \cellcolor[gray]{0.95}\textbf{6.45\tiny$\pm$2.00} & \cellcolor[gray]{0.95}\textbf{1.81\tiny$\pm$0.17} & 87.93\tiny$\pm$0.48 & 8.56\tiny$\pm$0.60 \\[3pt]
    
    22 & \cellcolor[gray]{0.95}\textbf{1.97\tiny$\pm$0.22} & 72.53\tiny$\pm$0.21 & 16.97\tiny$\pm$0.13 & 10.18\tiny$\pm$0.78 & 75.37\tiny$\pm$4.74 & \cellcolor[gray]{0.95}\textbf{14.55\tiny$\pm$3.48} & 2.22\tiny$\pm$0.65 & \cellcolor[gray]{0.95}\textbf{71.68\tiny$\pm$1.20} & 14.94\tiny$\pm$1.05 \\[3pt]
    
    23 & \cellcolor[gray]{0.95}\textbf{2.13\tiny$\pm$0.40} & 54.41\tiny$\pm$0.79 & 21.87\tiny$\pm$0.39 & 11.90\tiny$\pm$0.92 & 61.64\tiny$\pm$5.01 & \cellcolor[gray]{0.95}\textbf{18.32\tiny$\pm$3.81} & 2.56\tiny$\pm$0.92 & \cellcolor[gray]{0.95}\textbf{53.61\tiny$\pm$1.61} & 18.67\tiny$\pm$1.00 \\[3pt]
    
    24 & \cellcolor[gray]{0.85}\textbf{2.07\tiny$\pm$0.26} & 65.40\tiny$\pm$0.70 & 18.50\tiny$\pm$0.22 & 12.62\tiny$\pm$0.80 & 71.73\tiny$\pm$4.30 & \cellcolor[gray]{0.95}\textbf{16.52\tiny$\pm$2.98} & 3.01\tiny$\pm$0.48 & \cellcolor[gray]{0.95}\textbf{62.07\tiny$\pm$1.01} & 17.19\tiny$\pm$1.02 \\[3pt]
    
    25 & \cellcolor[gray]{0.95}\textbf{2.04\tiny$\pm$0.44} & 63.56\tiny$\pm$0.95 & 14.04\tiny$\pm$0.27 & 10.41\tiny$\pm$1.08 & 72.14\tiny$\pm$3.23 & \cellcolor[gray]{0.95}\textbf{12.75\tiny$\pm$1.89} & 2.67\tiny$\pm$0.29 & \cellcolor[gray]{0.95}\textbf{59.90\tiny$\pm$1.37} & 12.83\tiny$\pm$0.63 \\[3pt]
    
    \midrule
    
    \textbf{Average} & 2.18\tiny$\pm$0.35 & 71.75\tiny$\pm$0.36 & 19.82\tiny$\pm$0.13 & 11.26\tiny$\pm$1.05 & 68.04\tiny$\pm$3.96 & 18.87\tiny$\pm$3.09 & 2.90\tiny$\pm$0.71 & 66.06\tiny$\pm$2.03 & 17.87\tiny$\pm$1.43 \\[3pt]

    \bottomrule
    \end{tabular}
    }
    \label{table:aece_unimodals}
\end{table*}

%% file: tables/table_aece_new_multimodals.tex
\begin{table*}[h!]
\caption{
    \textbf{Adaptive ECE for Multimodal Models.}
    Aggregate and class-stratified Adaptive Expected Calibration Error (AECE) are reported for standard multimodal architectures.
    The positive class exhibits substantially higher AECE than the negative class across MedFuse, DrFuse, and MeTra, reinforcing that multimodal performance gains do not eliminate class-dependent calibration errors.
    \footnotesize{(Dark-bold: $p<0.05$, Wilcoxon signed-rank test, 5 seeds; Light-bold: highest mean, not significant)}
    }
    \centering
    \resizebox{\linewidth}{!}{
    \begin{tabular}{c|ccc|ccc|cccc}
    \multirow{2}{*}[-1.75em]{\textbf{\makecell{Clinical \\ Condition}}} 

    & \multicolumn{3}{c}{\textbf{MedFuse}}
    & \multicolumn{3}{c}{\textbf{DrFuse}}
    & \multicolumn{3}{c}{\textbf{MeTra}}\\
    \midrule
    \addlinespace[10pt]
    & \textbf{$\widehat{\mathbf{AECE}} \downarrow$}
    & \textbf{$\widehat{\mathbf{AECE}}_{c=1} \downarrow$}
    & \textbf{$\widehat{\mathbf{AECE}}_{c=0} \downarrow$}
    
     & \textbf{$\widehat{\mathbf{AECE}} \downarrow$}
    & \textbf{$\widehat{\mathbf{AECE}}_{c=1} \downarrow$}
    & \textbf{$\widehat{\mathbf{AECE}}_{c=0} \downarrow$}
    
     & \textbf{$\widehat{\mathbf{AECE}} \downarrow$}
    & \textbf{$\widehat{\mathbf{AECE}}_{c=1} \downarrow$}
    & \textbf{$\widehat{\mathbf{AECE}}_{c=0} \downarrow$}
    \\
    \midrule

    1 & 3.22\tiny$\pm$0.91 & \cellcolor[gray]{0.95}\textbf{54.09\tiny$\pm$2.26} & \cellcolor[gray]{0.95}\textbf{25.83\tiny$\pm$2.03} & 5.16\tiny$\pm$1.20 & 55.57\tiny$\pm$3.44 & 26.30\tiny$\pm$2.71 & \cellcolor[gray]{0.95}\textbf{3.01\tiny$\pm$0.39} & 59.40\tiny$\pm$1.02 & 26.08\tiny$\pm$1.02 \\[3pt]

    2 & \cellcolor[gray]{0.95}\textbf{1.29\tiny$\pm$0.26} & 67.96\tiny$\pm$1.40 & \cellcolor[gray]{0.95}\textbf{5.40\tiny$\pm$0.18} & 1.92\tiny$\pm$1.05 & 60.49\tiny$\pm$4.83 & 6.21\tiny$\pm$1.11 & 2.02\tiny$\pm$0.62 & \cellcolor[gray]{0.95}\textbf{59.96\tiny$\pm$5.55} & 6.44\tiny$\pm$1.01 \\[3pt]
    
    3 & \cellcolor[gray]{0.95}\textbf{1.46\tiny$\pm$0.30} & 86.93\tiny$\pm$0.14 & 8.05\tiny$\pm$0.25 & 2.37\tiny$\pm$0.35 & 88.15\tiny$\pm$2.27 & \cellcolor[gray]{0.95}\textbf{7.83\tiny$\pm$1.71} & 2.11\tiny$\pm$1.13 & \cellcolor[gray]{0.95}\textbf{84.75\tiny$\pm$2.92} & 9.34\tiny$\pm$1.95 \\[3pt]
    
    4 & \cellcolor[gray]{0.95}\textbf{3.93\tiny$\pm$1.67} & \cellcolor[gray]{0.95}\textbf{55.81\tiny$\pm$3.90} & \cellcolor[gray]{0.95}\textbf{32.44\tiny$\pm$3.51} & 4.82\tiny$\pm$2.29 & 58.41\tiny$\pm$5.60 & 34.07\tiny$\pm$4.56 & 5.16\tiny$\pm$1.72 & 59.83\tiny$\pm$4.55 & 34.46\tiny$\pm$4.44 \\[3pt]
    
    5 & 3.64\tiny$\pm$1.62 & \cellcolor[gray]{0.95}\textbf{61.94\tiny$\pm$4.35} & \cellcolor[gray]{0.95}\textbf{18.95\tiny$\pm$3.07} & 4.64\tiny$\pm$2.03 & 66.72\tiny$\pm$4.71 & 19.40\tiny$\pm$3.56 & \cellcolor[gray]{0.95}\textbf{3.36\tiny$\pm$1.01} & 70.00\tiny$\pm$1.94 & 20.06\tiny$\pm$1.60 \\[3pt]
    
    6 & 3.30\tiny$\pm$0.47 & \cellcolor[gray]{0.85}\textbf{72.08\tiny$\pm$2.20} & \cellcolor[gray]{0.95}\textbf{13.20\tiny$\pm$1.03} & \cellcolor[gray]{0.95}\textbf{2.09\tiny$\pm$0.41} & 80.69\tiny$\pm$1.41 & 13.62\tiny$\pm$1.68 & 3.04\tiny$\pm$0.72 & 80.71\tiny$\pm$0.80 & 15.64\tiny$\pm$1.15 \\[3pt]
    
    7 & \cellcolor[gray]{0.95}\textbf{2.10\tiny$\pm$0.35} & 72.23\tiny$\pm$0.49 & 20.56\tiny$\pm$0.37 & 2.59\tiny$\pm$0.81 & 71.65\tiny$\pm$2.04 & \cellcolor[gray]{0.95}\textbf{19.77\tiny$\pm$1.74} & 2.65\tiny$\pm$0.56 & \cellcolor[gray]{0.95}\textbf{68.49\tiny$\pm$2.80} & 20.47\tiny$\pm$1.49 \\[3pt]
    
    8 & \cellcolor[gray]{0.95}\textbf{2.85\tiny$\pm$0.49} & \cellcolor[gray]{0.85}\textbf{54.39\tiny$\pm$3.81} & \cellcolor[gray]{0.85}\textbf{6.16\tiny$\pm$1.40} & 2.95\tiny$\pm$0.72 & 83.42\tiny$\pm$2.79 & 9.94\tiny$\pm$1.79 & 2.86\tiny$\pm$0.62 & 83.64\tiny$\pm$2.86 & 10.67\tiny$\pm$1.87 \\[3pt]
    
    9 & 5.63\tiny$\pm$1.61 & \cellcolor[gray]{0.95}\textbf{50.90\tiny$\pm$5.67} & \cellcolor[gray]{0.95}\textbf{21.83\tiny$\pm$4.31} & 5.96\tiny$\pm$1.98 & 59.53\tiny$\pm$5.53 & 23.29\tiny$\pm$3.96 & \cellcolor[gray]{0.95}\textbf{3.77\tiny$\pm$2.08} & 60.69\tiny$\pm$3.90 & 26.55\tiny$\pm$3.21 \\[3pt]
    
    10 & 4.60\tiny$\pm$1.31 & \cellcolor[gray]{0.95}\textbf{50.67\tiny$\pm$3.65} & \cellcolor[gray]{0.95}\textbf{24.37\tiny$\pm$3.03} & 5.60\tiny$\pm$4.01 & 58.56\tiny$\pm$6.49 & 25.02\tiny$\pm$4.83 & \cellcolor[gray]{0.95}\textbf{3.66\tiny$\pm$0.54} & 60.09\tiny$\pm$2.47 & 27.23\tiny$\pm$1.50 \\[3pt]
    
    11 & \cellcolor[gray]{0.85}\textbf{1.44\tiny$\pm$0.33} & 65.92\tiny$\pm$1.70 & \cellcolor[gray]{0.95}\textbf{9.03\tiny$\pm$0.38} & 3.56\tiny$\pm$1.41 & \cellcolor[gray]{0.95}\textbf{56.76\tiny$\pm$3.60} & 10.96\tiny$\pm$1.96 & 3.97\tiny$\pm$1.48 & 66.96\tiny$\pm$7.52 & 9.91\tiny$\pm$3.44 \\[3pt]
    
    12 & \cellcolor[gray]{0.95}\textbf{2.63\tiny$\pm$0.40} & 71.76\tiny$\pm$1.61 & \cellcolor[gray]{0.95}\textbf{18.97\tiny$\pm$1.05} & 5.34\tiny$\pm$1.63 & \cellcolor[gray]{0.95}\textbf{66.72\tiny$\pm$4.76} & 21.70\tiny$\pm$4.08 & 3.05\tiny$\pm$0.36 & 70.49\tiny$\pm$2.58 & 19.90\tiny$\pm$1.71 \\[3pt]
    
    13 & \cellcolor[gray]{0.95}\textbf{4.51\tiny$\pm$0.75} & \cellcolor[gray]{0.95}\textbf{49.43\tiny$\pm$2.11} & \cellcolor[gray]{0.95}\textbf{36.43\tiny$\pm$2.12} & 6.29\tiny$\pm$2.27 & 51.18\tiny$\pm$5.76 & 37.77\tiny$\pm$5.41 & 4.51\tiny$\pm$1.15 & 54.64\tiny$\pm$3.47 & 37.09\tiny$\pm$3.68 \\[3pt]
    
    14 & \cellcolor[gray]{0.95}\textbf{3.75\tiny$\pm$0.81} & \cellcolor[gray]{0.95}\textbf{50.60\tiny$\pm$2.22} & \cellcolor[gray]{0.95}\textbf{39.56\tiny$\pm$2.06} & 4.26\tiny$\pm$1.22 & 52.29\tiny$\pm$3.74 & 41.47\tiny$\pm$3.68 & 5.42\tiny$\pm$1.12 & 52.00\tiny$\pm$3.93 & 42.60\tiny$\pm$4.47 \\[3pt]
    
    15 & \cellcolor[gray]{0.95}\textbf{3.67\tiny$\pm$0.92} & 46.28\tiny$\pm$1.19 & \cellcolor[gray]{0.95}\textbf{36.83\tiny$\pm$1.55} & 4.52\tiny$\pm$0.97 & \cellcolor[gray]{0.95}\textbf{45.05\tiny$\pm$1.60} & 37.45\tiny$\pm$1.67 & 4.54\tiny$\pm$1.00 & 48.14\tiny$\pm$2.66 & 37.85\tiny$\pm$2.71 \\[3pt]
    
    16 & 1.73\tiny$\pm$0.34 & 89.64\tiny$\pm$0.80 & 7.21\tiny$\pm$0.42 & \cellcolor[gray]{0.95}\textbf{1.41\tiny$\pm$0.13} & \cellcolor[gray]{0.95}\textbf{88.91\tiny$\pm$0.80} & \cellcolor[gray]{0.95}\textbf{6.84\tiny$\pm$0.74} & 1.58\tiny$\pm$0.32 & 89.01\tiny$\pm$2.17 & 6.91\tiny$\pm$0.99 \\[3pt]
    
    17 & 3.86\tiny$\pm$1.29 & \cellcolor[gray]{0.95}\textbf{64.76\tiny$\pm$4.25} & \cellcolor[gray]{0.95}\textbf{17.59\tiny$\pm$2.93} & 5.09\tiny$\pm$2.14 & 69.27\tiny$\pm$5.22 & 18.16\tiny$\pm$3.98 & \cellcolor[gray]{0.85}\textbf{2.79\tiny$\pm$0.72} & 72.40\tiny$\pm$1.46 & 18.41\tiny$\pm$1.27 \\[3pt]
    
    18 & 3.08\tiny$\pm$0.99 & \cellcolor[gray]{0.95}\textbf{72.07\tiny$\pm$2.49} & 15.49\tiny$\pm$0.95 & \cellcolor[gray]{0.95}\textbf{2.85\tiny$\pm$0.60} & 75.23\tiny$\pm$2.04 & 15.81\tiny$\pm$1.36 & 3.03\tiny$\pm$0.75 & 79.54\tiny$\pm$2.29 & \cellcolor[gray]{0.95}\textbf{14.56\tiny$\pm$1.57} \\[3pt]
    
    19 & 2.11\tiny$\pm$0.15 & 86.71\tiny$\pm$0.43 & \cellcolor[gray]{0.95}\textbf{11.15\tiny$\pm$0.47} & \cellcolor[gray]{0.95}\textbf{2.09\tiny$\pm$0.34} & 86.99\tiny$\pm$1.08 & 11.54\tiny$\pm$1.14 & 2.38\tiny$\pm$0.61 & \cellcolor[gray]{0.95}\textbf{86.28\tiny$\pm$0.82} & 12.37\tiny$\pm$0.55 \\[3pt]
    
    20 & \cellcolor[gray]{0.95}\textbf{1.34\tiny$\pm$0.28} & 92.06\tiny$\pm$0.39 & 5.53\tiny$\pm$0.33 & 1.36\tiny$\pm$0.27 & 91.18\tiny$\pm$0.85 & \cellcolor[gray]{0.95}\textbf{4.63\tiny$\pm$0.81} & 1.77\tiny$\pm$0.78 & \cellcolor[gray]{0.95}\textbf{91.13\tiny$\pm$2.14} & 5.37\tiny$\pm$1.20 \\[3pt]
    
    21 & 1.81\tiny$\pm$0.17 & \cellcolor[gray]{0.95}\textbf{87.93\tiny$\pm$0.48} & 8.56\tiny$\pm$0.60 & \cellcolor[gray]{0.95}\textbf{1.79\tiny$\pm$0.57} & 89.23\tiny$\pm$0.93 & \cellcolor[gray]{0.95}\textbf{8.20\tiny$\pm$0.81} & 2.69\tiny$\pm$0.33 & 88.34\tiny$\pm$1.42 & 9.54\tiny$\pm$1.33 \\[3pt]
    
    22 & \cellcolor[gray]{0.95}\textbf{2.22\tiny$\pm$0.65} & 71.68\tiny$\pm$1.20 & 14.94\tiny$\pm$1.05 & 2.70\tiny$\pm$0.92 & \cellcolor[gray]{0.95}\textbf{71.49\tiny$\pm$2.65} & \cellcolor[gray]{0.95}\textbf{14.08\tiny$\pm$1.39} & 2.60\tiny$\pm$0.98 & 72.13\tiny$\pm$2.90 & 15.42\tiny$\pm$1.77 \\[3pt]
    
    23 & \cellcolor[gray]{0.95}\textbf{2.56\tiny$\pm$0.92} & 53.61\tiny$\pm$1.61 & \cellcolor[gray]{0.95}\textbf{18.67\tiny$\pm$1.00} & 3.39\tiny$\pm$0.88 & 51.35\tiny$\pm$2.24 & 19.06\tiny$\pm$1.41 & 3.45\tiny$\pm$1.10 & \cellcolor[gray]{0.95}\textbf{50.94\tiny$\pm$4.49} & 20.52\tiny$\pm$2.38 \\[3pt]
    
    24 & \cellcolor[gray]{0.95}\textbf{3.01\tiny$\pm$0.48} & 62.07\tiny$\pm$1.01 & 17.19\tiny$\pm$1.02 & 3.14\tiny$\pm$1.03 & \cellcolor[gray]{0.95}\textbf{58.51\tiny$\pm$3.24} & 17.23\tiny$\pm$1.22 & 3.76\tiny$\pm$0.83 & 63.73\tiny$\pm$4.49 & \cellcolor[gray]{0.95}\textbf{16.19\tiny$\pm$1.87} \\[3pt]
    
    25 & \cellcolor[gray]{0.95}\textbf{2.67\tiny$\pm$0.29} & 59.90\tiny$\pm$1.37 & 12.83\tiny$\pm$0.63 & 3.39\tiny$\pm$0.75 & \cellcolor[gray]{0.85}\textbf{54.13\tiny$\pm$2.21} & 12.96\tiny$\pm$0.46 & 3.28\tiny$\pm$0.86 & 61.72\tiny$\pm$4.80 & \cellcolor[gray]{0.95}\textbf{12.07\tiny$\pm$2.19} \\[3pt]
    
    \midrule
    
    \textbf{Average} & 2.90\tiny$\pm$0.71 & 66.06\tiny$\pm$2.03 & 17.87\tiny$\pm$1.43 & 3.57\tiny$\pm$1.20 & 67.66\tiny$\pm$3.19 & 18.53\tiny$\pm$2.31 & 3.22\tiny$\pm$0.87 & 69.40\tiny$\pm$3.04 & 19.03\tiny$\pm$2.01 \\[3pt]

    \bottomrule
    \end{tabular}
    }
    \label{table:aece_multimodals}
\end{table*}

%% file: tables/table_aece_upscaled_multimodals.tex
\begin{table*}[h!]
\caption{
    \textbf{Adaptive ECE for Multimodal Models under Loss-Upweighting.}
    Aggregate and class-stratified Adaptive Expected Calibration Error (AECE) are reported for multimodal architectures trained with loss upweighting.
    Loss upweighting reduces positive-class AECE relative to standard multimodal models, but it redistributes calibration error toward the negative class, indicating that this mitigation does not fully resolve class-dependent miscalibration.
    \footnotesize{(Dark-bold: $p<0.05$, Wilcoxon signed-rank test, 5 seeds; Light-bold: highest mean, not significant)}
    }
    \centering
    \resizebox{\linewidth}{!}{
    \begin{tabular}{c|ccc|ccc|cccc}
    \multirow{2}{*}[-1.75em]{\textbf{\makecell{Clinical \\ Condition}}} 

    & \multicolumn{3}{c}{\textbf{MedFuse (Loss-Upweighting)}}
    & \multicolumn{3}{c}{\textbf{DrFuse (Loss-Upweighting)}}
    & \multicolumn{3}{c}{\textbf{MeTra (Loss-Upweighting)}}\\
    \midrule
    \addlinespace[10pt]
    & \textbf{$\widehat{\mathbf{AECE}} \downarrow$}
    & \textbf{$\widehat{\mathbf{AECE}}_{c=1} \downarrow$}
    & \textbf{$\widehat{\mathbf{AECE}}_{c=0} \downarrow$}
    
     & \textbf{$\widehat{\mathbf{AECE}} \downarrow$}
    & \textbf{$\widehat{\mathbf{AECE}}_{c=1} \downarrow$}
    & \textbf{$\widehat{\mathbf{AECE}}_{c=0} \downarrow$}
    
     & \textbf{$\widehat{\mathbf{AECE}} \downarrow$}
    & \textbf{$\widehat{\mathbf{AECE}}_{c=1} \downarrow$}
    & \textbf{$\widehat{\mathbf{AECE}}_{c=0} \downarrow$}
    \\
    \midrule

    1 & 14.47\tiny$\pm$1.33 & \cellcolor[gray]{0.95}\textbf{38.49\tiny$\pm$1.01} & \cellcolor[gray]{0.95}\textbf{39.54\tiny$\pm$1.49} & 14.91\tiny$\pm$1.51 & 40.19\tiny$\pm$1.19 & 40.42\tiny$\pm$1.72 & \cellcolor[gray]{0.95}\textbf{13.67\tiny$\pm$3.13} & 43.47\tiny$\pm$3.99 & 40.70\tiny$\pm$2.76 \\[3pt]

    2 & 15.85\tiny$\pm$1.07 & 28.20\tiny$\pm$1.19 & 19.61\tiny$\pm$1.08 & \cellcolor[gray]{0.95}\textbf{15.46\tiny$\pm$3.11} & 26.76\tiny$\pm$3.93 & \cellcolor[gray]{0.95}\textbf{18.51\tiny$\pm$3.20} & 21.47\tiny$\pm$3.66 & \cellcolor[gray]{0.95}\textbf{22.96\tiny$\pm$1.76} & 25.26\tiny$\pm$3.93 \\[3pt]
    
    3 & \cellcolor[gray]{0.95}\textbf{30.73\tiny$\pm$2.15} & 43.34\tiny$\pm$2.27 & \cellcolor[gray]{0.95}\textbf{38.31\tiny$\pm$2.15} & 35.58\tiny$\pm$5.53 & \cellcolor[gray]{0.95}\textbf{41.28\tiny$\pm$5.99} & 43.24\tiny$\pm$5.51 & 31.64\tiny$\pm$7.15 & 46.78\tiny$\pm$8.88 & 39.66\tiny$\pm$6.98 \\[3pt]
    
    4 & 11.02\tiny$\pm$2.99 & \cellcolor[gray]{0.95}\textbf{43.65\tiny$\pm$2.86} & \cellcolor[gray]{0.95}\textbf{44.38\tiny$\pm$3.15} & 12.31\tiny$\pm$1.80 & 46.40\tiny$\pm$1.75 & 48.18\tiny$\pm$1.84 & \cellcolor[gray]{0.95}\textbf{9.66\tiny$\pm$2.90} & 48.86\tiny$\pm$3.10 & 45.35\tiny$\pm$2.94 \\[3pt]
    
    5 & \cellcolor[gray]{0.95}\textbf{19.10\tiny$\pm$2.90} & \cellcolor[gray]{0.95}\textbf{39.06\tiny$\pm$2.45} & \cellcolor[gray]{0.85}\textbf{37.57\tiny$\pm$3.07} & 22.74\tiny$\pm$3.28 & 41.06\tiny$\pm$3.14 & 42.39\tiny$\pm$3.37 & 21.17\tiny$\pm$5.49 & 45.53\tiny$\pm$6.87 & 42.35\tiny$\pm$5.08 \\[3pt]
    
    6 & \cellcolor[gray]{0.85}\textbf{22.38\tiny$\pm$5.69} & \cellcolor[gray]{0.95}\textbf{43.41\tiny$\pm$6.81} & \cellcolor[gray]{0.85}\textbf{33.87\tiny$\pm$5.52} & 32.35\tiny$\pm$1.77 & 45.29\tiny$\pm$1.73 & 46.18\tiny$\pm$1.78 & 29.49\tiny$\pm$1.48 & 50.92\tiny$\pm$1.53 & 43.53\tiny$\pm$1.50 \\[3pt]
    
    7 & 26.14\tiny$\pm$0.83 & 43.50\tiny$\pm$0.79 & 46.45\tiny$\pm$0.84 & 26.42\tiny$\pm$1.30 & \cellcolor[gray]{0.95}\textbf{43.32\tiny$\pm$1.62} & 45.61\tiny$\pm$1.26 & \cellcolor[gray]{0.95}\textbf{22.74\tiny$\pm$3.04} & 45.18\tiny$\pm$3.38 & \cellcolor[gray]{0.95}\textbf{42.55\tiny$\pm$3.00} \\[3pt]
    
    8 & \cellcolor[gray]{0.85}\textbf{19.58\tiny$\pm$4.81} & \cellcolor[gray]{0.85}\textbf{32.96\tiny$\pm$4.22} & \cellcolor[gray]{0.85}\textbf{26.28\tiny$\pm$5.02} & 32.57\tiny$\pm$3.77 & 44.21\tiny$\pm$4.29 & 42.06\tiny$\pm$3.71 & 29.97\tiny$\pm$5.85 & 48.38\tiny$\pm$6.71 & 40.15\tiny$\pm$5.74 \\[3pt]
    
    9 & \cellcolor[gray]{0.95}\textbf{14.09\tiny$\pm$4.56} & \cellcolor[gray]{0.95}\textbf{36.02\tiny$\pm$4.54} & \cellcolor[gray]{0.95}\textbf{34.69\tiny$\pm$4.87} & 19.13\tiny$\pm$2.87 & 38.66\tiny$\pm$2.72 & 43.25\tiny$\pm$2.94 & 15.60\tiny$\pm$4.46 & 46.09\tiny$\pm$5.79 & 41.32\tiny$\pm$4.08 \\[3pt]
    
    10 & 12.76\tiny$\pm$3.80 & \cellcolor[gray]{0.95}\textbf{36.72\tiny$\pm$4.26} & \cellcolor[gray]{0.95}\textbf{36.87\tiny$\pm$4.31} & 16.44\tiny$\pm$4.15 & 39.35\tiny$\pm$3.87 & 44.52\tiny$\pm$4.30 & \cellcolor[gray]{0.95}\textbf{11.47\tiny$\pm$5.38} & 47.35\tiny$\pm$6.92 & 40.48\tiny$\pm$5.98 \\[3pt]
    
    11 & \cellcolor[gray]{0.95}\textbf{19.46\tiny$\pm$1.80} & 30.43\tiny$\pm$1.15 & \cellcolor[gray]{0.95}\textbf{26.26\tiny$\pm$1.90} & 22.79\tiny$\pm$4.17 & \cellcolor[gray]{0.95}\textbf{26.38\tiny$\pm$3.88} & 29.41\tiny$\pm$4.23 & 26.63\tiny$\pm$6.84 & 32.92\tiny$\pm$6.52 & 34.76\tiny$\pm$7.06 \\[3pt]
    
    12 & \cellcolor[gray]{0.95}\textbf{23.30\tiny$\pm$1.27} & 42.68\tiny$\pm$1.36 & \cellcolor[gray]{0.95}\textbf{40.99\tiny$\pm$1.29} & 26.10\tiny$\pm$4.15 & \cellcolor[gray]{0.95}\textbf{40.14\tiny$\pm$4.34} & 43.63\tiny$\pm$4.11 & 26.53\tiny$\pm$5.90 & 42.61\tiny$\pm$4.93 & 45.07\tiny$\pm$6.25 \\[3pt]
    
    13 & 9.12\tiny$\pm$2.37 & 43.34\tiny$\pm$3.42 & \cellcolor[gray]{0.95}\textbf{42.34\tiny$\pm$3.63} & 10.02\tiny$\pm$4.06 & \cellcolor[gray]{0.95}\textbf{43.05\tiny$\pm$4.03} & 47.20\tiny$\pm$4.09 & \cellcolor[gray]{0.95}\textbf{8.85\tiny$\pm$4.11} & 46.86\tiny$\pm$5.58 & 45.61\tiny$\pm$5.10 \\[3pt]
    
    14 & \cellcolor[gray]{0.95}\textbf{6.64\tiny$\pm$2.80} & 46.92\tiny$\pm$3.72 & \cellcolor[gray]{0.95}\textbf{44.29\tiny$\pm$3.80} & 8.55\tiny$\pm$1.99 & \cellcolor[gray]{0.95}\textbf{45.22\tiny$\pm$2.00} & 49.87\tiny$\pm$2.09 & 8.30\tiny$\pm$1.89 & 46.71\tiny$\pm$2.69 & 49.15\tiny$\pm$2.71 \\[3pt]
    
    15 & 4.72\tiny$\pm$0.70 & 42.31\tiny$\pm$1.07 & \cellcolor[gray]{0.95}\textbf{41.49\tiny$\pm$1.21} & 5.10\tiny$\pm$1.56 & \cellcolor[gray]{0.95}\textbf{42.00\tiny$\pm$2.10} & 41.85\tiny$\pm$2.15 & \cellcolor[gray]{0.95}\textbf{4.61\tiny$\pm$0.89} & 43.61\tiny$\pm$1.45 & 42.75\tiny$\pm$0.80 \\[3pt]
    
    16 & \cellcolor[gray]{0.95}\textbf{33.39\tiny$\pm$2.00} & 48.08\tiny$\pm$2.12 & 39.63\tiny$\pm$2.01 & 33.64\tiny$\pm$2.55 & \cellcolor[gray]{0.85}\textbf{41.54\tiny$\pm$3.60} & \cellcolor[gray]{0.95}\textbf{39.19\tiny$\pm$2.49} & 37.05\tiny$\pm$4.10 & 44.48\tiny$\pm$3.11 & 43.29\tiny$\pm$4.18 \\[3pt]
    
    17 & \cellcolor[gray]{0.95}\textbf{20.11\tiny$\pm$3.19} & \cellcolor[gray]{0.95}\textbf{39.77\tiny$\pm$2.72} & \cellcolor[gray]{0.85}\textbf{37.21\tiny$\pm$3.34} & 24.27\tiny$\pm$3.37 & 41.72\tiny$\pm$3.31 & 42.53\tiny$\pm$3.41 & 23.11\tiny$\pm$5.01 & 45.61\tiny$\pm$6.25 & 42.73\tiny$\pm$4.70 \\[3pt]
    
    18 & \cellcolor[gray]{0.95}\textbf{27.01\tiny$\pm$3.55} & \cellcolor[gray]{0.95}\textbf{40.99\tiny$\pm$3.09} & \cellcolor[gray]{0.95}\textbf{40.88\tiny$\pm$3.69} & 27.99\tiny$\pm$2.45 & 44.11\tiny$\pm$2.58 & 42.46\tiny$\pm$2.43 & 30.46\tiny$\pm$1.86 & 44.18\tiny$\pm$2.04 & 45.69\tiny$\pm$1.86 \\[3pt]
    
    19 & \cellcolor[gray]{0.95}\textbf{34.37\tiny$\pm$1.04} & \cellcolor[gray]{0.95}\textbf{48.07\tiny$\pm$0.92} & \cellcolor[gray]{0.95}\textbf{46.22\tiny$\pm$1.07} & 35.13\tiny$\pm$1.06 & 48.70\tiny$\pm$1.27 & 47.21\tiny$\pm$1.04 & 35.67\tiny$\pm$3.84 & 49.40\tiny$\pm$4.29 & 47.88\tiny$\pm$3.78 \\[3pt]
    
    20 & 31.26\tiny$\pm$1.58 & \cellcolor[gray]{0.95}\textbf{46.86\tiny$\pm$1.13} & 35.72\tiny$\pm$1.61 & \cellcolor[gray]{0.85}\textbf{28.76\tiny$\pm$2.18} & 50.16\tiny$\pm$2.28 & \cellcolor[gray]{0.85}\textbf{33.12\tiny$\pm$2.24} & 32.60\tiny$\pm$3.85 & 52.47\tiny$\pm$4.25 & 37.46\tiny$\pm$3.83 \\[3pt]
    
    21 & 35.91\tiny$\pm$3.05 & \cellcolor[gray]{0.85}\textbf{41.29\tiny$\pm$2.14} & 44.02\tiny$\pm$3.16 & 34.58\tiny$\pm$2.43 & 48.66\tiny$\pm$2.44 & 43.56\tiny$\pm$2.44 & \cellcolor[gray]{0.95}\textbf{32.90\tiny$\pm$2.58} & 53.28\tiny$\pm$2.44 & \cellcolor[gray]{0.95}\textbf{41.96\tiny$\pm$2.60} \\[3pt]
    
    22 & 24.67\tiny$\pm$1.45 & \cellcolor[gray]{0.85}\textbf{38.51\tiny$\pm$1.15} & 38.98\tiny$\pm$1.52 & 23.37\tiny$\pm$2.08 & 41.12\tiny$\pm$2.30 & \cellcolor[gray]{0.95}\textbf{37.91\tiny$\pm$2.11} & \cellcolor[gray]{0.95}\textbf{22.88\tiny$\pm$0.85} & 43.97\tiny$\pm$1.61 & 38.03\tiny$\pm$0.73 \\[3pt]
    
    23 & 13.66\tiny$\pm$1.51 & \cellcolor[gray]{0.95}\textbf{34.59\tiny$\pm$1.06} & 32.58\tiny$\pm$1.73 & 14.17\tiny$\pm$2.79 & 35.16\tiny$\pm$3.00 & 32.85\tiny$\pm$2.84 & \cellcolor[gray]{0.95}\textbf{12.24\tiny$\pm$2.32} & 37.30\tiny$\pm$3.63 & \cellcolor[gray]{0.95}\textbf{31.61\tiny$\pm$1.89} \\[3pt]
    
    24 & 19.28\tiny$\pm$1.54 & 37.39\tiny$\pm$0.93 & 35.92\tiny$\pm$1.78 & 19.27\tiny$\pm$0.98 & \cellcolor[gray]{0.95}\textbf{36.12\tiny$\pm$1.00} & 35.24\tiny$\pm$1.04 & \cellcolor[gray]{0.95}\textbf{17.68\tiny$\pm$1.55} & 40.68\tiny$\pm$1.94 & \cellcolor[gray]{0.95}\textbf{34.84\tiny$\pm$1.54} \\[3pt]
    
    25 & 19.30\tiny$\pm$1.10 & 33.08\tiny$\pm$0.87 & 30.33\tiny$\pm$1.18 & 19.43\tiny$\pm$1.96 & \cellcolor[gray]{0.95}\textbf{30.19\tiny$\pm$2.23} & 29.27\tiny$\pm$2.05 & \cellcolor[gray]{0.95}\textbf{16.98\tiny$\pm$3.07} & 37.24\tiny$\pm$3.80 & \cellcolor[gray]{0.95}\textbf{28.33\tiny$\pm$3.03} \\[3pt]
    
    \midrule
    
    \textbf{Average} & 20.33\tiny$\pm$2.36 & 39.99\tiny$\pm$2.29 & 37.38\tiny$\pm$2.58 & 22.44\tiny$\pm$2.68 & 40.83\tiny$\pm$2.82 & 40.39\tiny$\pm$2.74 & 21.74\tiny$\pm$3.65 & 44.27\tiny$\pm$4.14 & 40.42\tiny$\pm$3.68 \\[3pt]

    \bottomrule
    \end{tabular}
    }
    \label{table:aece_multimodals_upweighted}
\end{table*}

%% file: tables/table_brier_unimodals.tex
\setlength{\tabcolsep}{6pt}
\begin{table*}[h!]
\caption{
    \textbf{Brier Score for Unimodal Models.}
    Aggregate and class-stratified Brier scores are reported for unimodal baselines.
    Positive-class Brier loss remains substantially higher than negative-class Brier loss, showing that class-dependent reliability failures persist under a proper scoring rule that does not rely on binning.
    \footnotesize{(Dark-bold: $p<0.05$, Wilcoxon signed-rank test, 5 seeds; Light-bold: highest mean, not significant)}
    }
    \centering
    \scriptsize
    \begin{adjustbox}{max width=\linewidth}
    \begin{tabular}{c|ccc|ccc|cccc}
    \multirow{2}{*}[-1.5em]{\textbf{\makecell{Clinical \\ Condition}}} 
    & \multicolumn{3}{c}{\textbf{EHR}}
    & \multicolumn{3}{c}{\textbf{CXR}}
    & \multicolumn{3}{c}{\textbf{MedFuse}}\\
    \midrule
    \addlinespace[5pt]
    & \textbf{$\mathbf{BS} \downarrow$}
    & \textbf{$\mathbf{BS}_{c=1} \downarrow $}
    & \textbf{$\mathbf{BS}_{c=0} \downarrow$}
    
     & \textbf{$\mathbf{BS} \downarrow$}
    & \textbf{$\mathbf{BS}_{c=1} \downarrow$}
    & \textbf{$\mathbf{BS}_{c=0} \downarrow$}
    
     & \textbf{$\mathbf{BS} \downarrow$}
    & \textbf{$\mathbf{BS}_{c=1} \downarrow$}
    & \textbf{$\mathbf{BS}_{c=0} \downarrow$}

    \\[2pt]
    \midrule

    1 & 18.75\tiny$\pm$0.07 & 37.06\tiny$\pm$0.30 & \cellcolor[gray]{0.95}\textbf{10.08\tiny$\pm$0.03} & 23.54\tiny$\pm$0.61 & 40.52\tiny$\pm$3.54 & 15.50\tiny$\pm$2.35 & \cellcolor[gray]{0.95}\textbf{17.90\tiny$\pm$0.21} & \cellcolor[gray]{0.95}\textbf{34.00\tiny$\pm$2.46} & 10.28\tiny$\pm$1.20 \\[3pt]
    
    2 & \cellcolor[gray]{0.95}\textbf{5.23\tiny$\pm$0.03} & \cellcolor[gray]{0.85}\textbf{48.19\tiny$\pm$0.43} & 1.54\tiny$\pm$0.05 & 8.04\tiny$\pm$0.30 & 78.77\tiny$\pm$4.03 & 1.98\tiny$\pm$0.64 & 5.30\tiny$\pm$0.12 & 50.55\tiny$\pm$1.85 & \cellcolor[gray]{0.95}\textbf{1.43\tiny$\pm$0.06} \\[3pt]
    
    3 & 8.10\tiny$\pm$0.05 & 78.15\tiny$\pm$0.58 & \cellcolor[gray]{0.85}\textbf{0.93\tiny$\pm$0.02} & 9.35\tiny$\pm$0.59 & 76.08\tiny$\pm$4.96 & 2.52\tiny$\pm$1.13 & \cellcolor[gray]{0.95}\textbf{7.99\tiny$\pm$0.02} & \cellcolor[gray]{0.95}\textbf{76.07\tiny$\pm$0.21} & 1.02\tiny$\pm$0.03 \\[3pt]
    
    4 & 22.91\tiny$\pm$0.07 & 37.48\tiny$\pm$0.14 & 14.00\tiny$\pm$0.10 & 24.46\tiny$\pm$0.50 & 36.38\tiny$\pm$4.50 & 17.17\tiny$\pm$3.53 & \cellcolor[gray]{0.95}\textbf{21.21\tiny$\pm$0.20} & \cellcolor[gray]{0.95}\textbf{34.26\tiny$\pm$4.39} & \cellcolor[gray]{0.95}\textbf{13.23\tiny$\pm$2.42} \\[3pt]
    
    5 & 16.75\tiny$\pm$0.11 & 51.69\tiny$\pm$0.41 & \cellcolor[gray]{0.95}\textbf{5.66\tiny$\pm$0.02} & 21.83\tiny$\pm$0.63 & 48.04\tiny$\pm$2.19 & 13.51\tiny$\pm$1.21 & \cellcolor[gray]{0.95}\textbf{15.29\tiny$\pm$0.18} & \cellcolor[gray]{0.95}\textbf{43.13\tiny$\pm$5.36} & 6.45\tiny$\pm$1.57 \\[3pt]
    
    6 & 12.48\tiny$\pm$0.04 & 66.83\tiny$\pm$0.33 & \cellcolor[gray]{0.95}\textbf{2.98\tiny$\pm$0.04} & 13.29\tiny$\pm$0.67 & 63.54\tiny$\pm$10.93 & 4.51\tiny$\pm$2.60 & \cellcolor[gray]{0.95}\textbf{11.56\tiny$\pm$0.06} & \cellcolor[gray]{0.95}\textbf{56.58\tiny$\pm$2.58} & 3.70\tiny$\pm$0.46 \\[3pt]
    
    7 & 16.47\tiny$\pm$0.10 & 55.29\tiny$\pm$0.52 & \cellcolor[gray]{0.95}\textbf{5.15\tiny$\pm$0.06} & 18.89\tiny$\pm$0.42 & 62.12\tiny$\pm$2.77 & 6.29\tiny$\pm$1.09 & \cellcolor[gray]{0.95}\textbf{16.24\tiny$\pm$0.02} & \cellcolor[gray]{0.95}\textbf{53.80\tiny$\pm$0.66} & 5.29\tiny$\pm$0.18 \\[3pt]
    
    8 & 9.98\tiny$\pm$0.04 & 76.34\tiny$\pm$0.45 & \cellcolor[gray]{0.95}\textbf{1.36\tiny$\pm$0.03} & 7.00\tiny$\pm$0.54 & \cellcolor[gray]{0.95}\textbf{43.27\tiny$\pm$6.68} & 2.29\tiny$\pm$1.31 & \cellcolor[gray]{0.95}\textbf{6.37\tiny$\pm$0.17} & 44.73\tiny$\pm$2.77 & 1.39\tiny$\pm$0.49 \\[3pt]
    
    9 & 19.06\tiny$\pm$0.13 & 40.56\tiny$\pm$0.46 & 10.04\tiny$\pm$0.06 & 20.44\tiny$\pm$1.65 & \cellcolor[gray]{0.95}\textbf{28.45\tiny$\pm$4.99} & 17.08\tiny$\pm$4.28 & \cellcolor[gray]{0.95}\textbf{16.53\tiny$\pm$0.29} & 33.18\tiny$\pm$6.06 & \cellcolor[gray]{0.95}\textbf{9.55\tiny$\pm$2.69} \\[3pt]
    
    10 & 20.00\tiny$\pm$0.10 & 40.70\tiny$\pm$0.18 & \cellcolor[gray]{0.95}\textbf{9.50\tiny$\pm$0.09} & 20.84\tiny$\pm$0.87 & 36.28\tiny$\pm$4.18 & 13.01\tiny$\pm$3.25 & \cellcolor[gray]{0.95}\textbf{17.67\tiny$\pm$0.21} & \cellcolor[gray]{0.95}\textbf{32.40\tiny$\pm$3.60} & 10.20\tiny$\pm$2.02 \\[3pt]
    
    11 & 9.53\tiny$\pm$0.13 & 65.54\tiny$\pm$1.60 & \cellcolor[gray]{0.85}\textbf{1.89\tiny$\pm$0.07} & 12.28\tiny$\pm$0.44 & 71.55\tiny$\pm$6.08 & 4.20\tiny$\pm$1.26 & \cellcolor[gray]{0.95}\textbf{8.03\tiny$\pm$0.05} & \cellcolor[gray]{0.95}\textbf{48.81\tiny$\pm$1.84} & 2.47\tiny$\pm$0.22 \\[3pt]
    
    12 & 16.19\tiny$\pm$0.10 & 60.04\tiny$\pm$0.56 & \cellcolor[gray]{0.85}\textbf{4.43\tiny$\pm$0.04} & 19.32\tiny$\pm$0.89 & 58.21\tiny$\pm$5.90 & 8.90\tiny$\pm$2.52 & \cellcolor[gray]{0.95}\textbf{15.26\tiny$\pm$0.09} & \cellcolor[gray]{0.95}\textbf{52.94\tiny$\pm$2.22} & 5.16\tiny$\pm$0.51 \\[3pt]
    
    13 & 22.47\tiny$\pm$0.06 & 31.82\tiny$\pm$0.17 & \cellcolor[gray]{0.95}\textbf{16.09\tiny$\pm$0.04} & 26.58\tiny$\pm$0.59 & \cellcolor[gray]{0.95}\textbf{25.34\tiny$\pm$6.30} & 27.43\tiny$\pm$5.23 & \cellcolor[gray]{0.95}\textbf{21.50\tiny$\pm$0.13} & 27.77\tiny$\pm$2.30 & 17.21\tiny$\pm$1.54 \\[3pt]
    
    14 & 23.90\tiny$\pm$0.12 & 29.94\tiny$\pm$0.33 & 19.29\tiny$\pm$0.07 & 28.35\tiny$\pm$0.93 & 33.54\tiny$\pm$5.82 & 24.38\tiny$\pm$5.90 & \cellcolor[gray]{0.95}\textbf{22.33\tiny$\pm$0.14} & \cellcolor[gray]{0.95}\textbf{27.85\tiny$\pm$2.13} & \cellcolor[gray]{0.95}\textbf{18.10\tiny$\pm$1.76} \\[3pt]
    
    15 & 21.64\tiny$\pm$0.13 & \cellcolor[gray]{0.95}\textbf{25.92\tiny$\pm$0.09} & 18.08\tiny$\pm$0.18 & 27.66\tiny$\pm$0.68 & 34.63\tiny$\pm$2.42 & 21.88\tiny$\pm$1.41 & \cellcolor[gray]{0.95}\textbf{21.31\tiny$\pm$0.15} & 25.98\tiny$\pm$1.42 & \cellcolor[gray]{0.95}\textbf{17.42\tiny$\pm$0.93} \\[3pt]
    
    16 & 6.52\tiny$\pm$0.01 & 84.09\tiny$\pm$0.26 & \cellcolor[gray]{0.85}\textbf{0.58\tiny$\pm$0.02} & 7.31\tiny$\pm$0.32 & 82.45\tiny$\pm$3.10 & 1.56\tiny$\pm$0.56 & \cellcolor[gray]{0.95}\textbf{6.49\tiny$\pm$0.03} & \cellcolor[gray]{0.95}\textbf{80.77\tiny$\pm$1.32} & 0.81\tiny$\pm$0.13 \\[3pt]
    
    17 & 15.97\tiny$\pm$0.08 & 55.57\tiny$\pm$0.35 & \cellcolor[gray]{0.95}\textbf{4.67\tiny$\pm$0.02} & 20.46\tiny$\pm$0.33 & 52.09\tiny$\pm$0.91 & 11.43\tiny$\pm$0.40 & \cellcolor[gray]{0.95}\textbf{14.74\tiny$\pm$0.16} & \cellcolor[gray]{0.95}\textbf{46.38\tiny$\pm$5.42} & 5.71\tiny$\pm$1.42 \\[3pt]
    
    18 & 13.50\tiny$\pm$0.04 & 66.02\tiny$\pm$0.22 & \cellcolor[gray]{0.85}\textbf{2.79\tiny$\pm$0.05} & 14.47\tiny$\pm$0.58 & 58.61\tiny$\pm$4.66 & 5.47\tiny$\pm$1.64 & \cellcolor[gray]{0.95}\textbf{12.99\tiny$\pm$0.17} & \cellcolor[gray]{0.95}\textbf{55.93\tiny$\pm$2.78} & 4.24\tiny$\pm$0.75 \\[3pt]
    
    19 & 10.85\tiny$\pm$0.02 & 74.26\tiny$\pm$0.43 & 1.74\tiny$\pm$0.04 & 12.98\tiny$\pm$0.55 & \cellcolor[gray]{0.95}\textbf{73.31\tiny$\pm$5.12} & 4.32\tiny$\pm$1.31 & \cellcolor[gray]{0.95}\textbf{10.76\tiny$\pm$0.02} & 75.48\tiny$\pm$0.76 & \cellcolor[gray]{0.95}\textbf{1.47\tiny$\pm$0.10} \\[3pt]
    
    20 & 5.10\tiny$\pm$0.02 & 84.78\tiny$\pm$0.40 & 0.55\tiny$\pm$0.03 & 5.74\tiny$\pm$0.27 & \cellcolor[gray]{0.95}\textbf{82.97\tiny$\pm$1.64} & 1.33\tiny$\pm$0.37 & \cellcolor[gray]{0.95}\textbf{5.05\tiny$\pm$0.01} & 85.03\tiny$\pm$0.72 & \cellcolor[gray]{0.95}\textbf{0.48\tiny$\pm$0.04} \\[3pt]
    
    21 & 8.56\tiny$\pm$0.03 & 78.76\tiny$\pm$0.31 & 1.18\tiny$\pm$0.04 & 9.24\tiny$\pm$0.37 & 78.18\tiny$\pm$4.57 & 1.99\tiny$\pm$0.86 & \cellcolor[gray]{0.95}\textbf{8.37\tiny$\pm$0.02} & \cellcolor[gray]{0.95}\textbf{77.82\tiny$\pm$0.83} & \cellcolor[gray]{0.95}\textbf{1.07\tiny$\pm$0.10} \\[3pt]
    
    22 & 13.59\tiny$\pm$0.05 & 54.71\tiny$\pm$0.34 & 4.28\tiny$\pm$0.05 & 16.18\tiny$\pm$0.45 & 62.20\tiny$\pm$6.11 & 5.75\tiny$\pm$1.90 & \cellcolor[gray]{0.95}\textbf{13.24\tiny$\pm$0.07} & \cellcolor[gray]{0.95}\textbf{54.17\tiny$\pm$1.79} & \cellcolor[gray]{0.95}\textbf{3.97\tiny$\pm$0.34} \\[3pt]
    
    23 & 15.23\tiny$\pm$0.27 & \cellcolor[gray]{0.95}\textbf{34.40\tiny$\pm$0.88} & 7.72\tiny$\pm$0.15 & 19.61\tiny$\pm$0.36 & 47.47\tiny$\pm$5.88 & 8.69\tiny$\pm$2.38 & \cellcolor[gray]{0.95}\textbf{14.87\tiny$\pm$0.18} & 35.14\tiny$\pm$1.96 & \cellcolor[gray]{0.95}\textbf{6.92\tiny$\pm$0.53} \\[3pt]
    
    24 & 14.90\tiny$\pm$0.15 & 46.48\tiny$\pm$0.64 & \cellcolor[gray]{0.95}\textbf{5.62\tiny$\pm$0.06} & 18.92\tiny$\pm$0.35 & 58.83\tiny$\pm$5.22 & 7.19\tiny$\pm$1.87 & \cellcolor[gray]{0.95}\textbf{14.42\tiny$\pm$0.09} & \cellcolor[gray]{0.95}\textbf{44.08\tiny$\pm$1.21} & 5.70\tiny$\pm$0.44 \\[3pt]
    
    25 & 11.27\tiny$\pm$0.21 & 45.32\tiny$\pm$0.81 & \cellcolor[gray]{0.95}\textbf{4.09\tiny$\pm$0.11} & 15.12\tiny$\pm$0.54 & 60.44\tiny$\pm$3.38 & 5.57\tiny$\pm$1.36 & \cellcolor[gray]{0.95}\textbf{10.94\tiny$\pm$0.09} & \cellcolor[gray]{0.95}\textbf{43.20\tiny$\pm$1.32} & 4.14\tiny$\pm$0.28 \\[3pt]
    
    \midrule
    
    \textbf{Average} & 14.36\tiny$\pm$0.09 & 54.80\tiny$\pm$0.45 & 6.17\tiny$\pm$0.06 & 16.88\tiny$\pm$0.58 & 55.73\tiny$\pm$4.64 & 9.36\tiny$\pm$2.01 & 13.45\tiny$\pm$0.12 & 49.60\tiny$\pm$2.32 & 6.30\tiny$\pm$0.81 \\[3pt]

    \bottomrule
    \end{tabular}
    \end{adjustbox}
    \label{table:brier_unimodals}
\end{table*}

%% file: tables/table_brier_new_multimodals.tex
\begin{table*}[h!]
\caption{
    \textbf{Brier Score for Multimodal Models.}
    Aggregate and class-stratified Brier scores are reported for standard multimodal architectures.
    Across MedFuse, DrFuse, and MeTra, the positive class has substantially higher Brier loss than the negative class, indicating that multimodal fusion does not remove minority-class calibration error despite improving aggregate discrimination.
    \footnotesize{(Dark-bold: $p<0.05$, Wilcoxon signed-rank test, 5 seeds; Light-bold: highest mean, not significant)}
    }
    \centering
    \scriptsize
    \begin{adjustbox}{max width=\linewidth}
    \begin{tabular}{c|ccc|ccc|cccc}
    \multirow{2}{*}[-1.5em]{\textbf{\makecell{Clinical \\ Condition}}} 
    & \multicolumn{3}{c}{\textbf{MedFuse}}
    & \multicolumn{3}{c}{\textbf{DrFuse}}
    & \multicolumn{3}{c}{\textbf{MeTra}}\\
    \midrule
    \addlinespace[5pt]
    & \textbf{$\mathbf{BS} \downarrow$}
    & \textbf{$\mathbf{BS}_{c=1} \downarrow $}
    & \textbf{$\mathbf{BS}_{c=0} \downarrow$}
    
     & \textbf{$\mathbf{BS} \downarrow$}
    & \textbf{$\mathbf{BS}_{c=1} \downarrow$}
    & \textbf{$\mathbf{BS}_{c=0} \downarrow$}
    
     & \textbf{$\mathbf{BS} \downarrow$}
    & \textbf{$\mathbf{BS}_{c=1} \downarrow$}
    & \textbf{$\mathbf{BS}_{c=0} \downarrow$}

    \\[2pt]
    \midrule

    1 & \cellcolor[gray]{0.85}\textbf{17.90\tiny$\pm$0.21} & \cellcolor[gray]{0.95}\textbf{34.00\tiny$\pm$2.46} & 10.28\tiny$\pm$1.20 & 18.89\tiny$\pm$0.31 & 35.95\tiny$\pm$3.57 & 10.97\tiny$\pm$1.95 & 18.99\tiny$\pm$0.20 & 38.85\tiny$\pm$0.98 & \cellcolor[gray]{0.95}\textbf{9.58\tiny$\pm$0.36} \\[3pt]
    
    2 & 5.30\tiny$\pm$0.12 & 50.55\tiny$\pm$1.85 & \cellcolor[gray]{0.95}\textbf{1.43\tiny$\pm$0.06} & \cellcolor[gray]{0.95}\textbf{4.99\tiny$\pm$0.11} & 43.53\tiny$\pm$5.07 & 2.04\tiny$\pm$0.49 & 5.10\tiny$\pm$0.04 & \cellcolor[gray]{0.95}\textbf{41.10\tiny$\pm$5.99} & 2.02\tiny$\pm$0.51 \\[3pt]
    
    3 & \cellcolor[gray]{0.95}\textbf{7.99\tiny$\pm$0.02} & 76.07\tiny$\pm$0.21 & 1.02\tiny$\pm$0.03 & 8.02\tiny$\pm$0.05 & 78.18\tiny$\pm$3.84 & \cellcolor[gray]{0.95}\textbf{1.01\tiny$\pm$0.41} & 8.05\tiny$\pm$0.11 & \cellcolor[gray]{0.95}\textbf{72.62\tiny$\pm$4.76} & 1.45\tiny$\pm$0.54 \\[3pt]
    
    4 & \cellcolor[gray]{0.85}\textbf{21.21\tiny$\pm$0.20} & \cellcolor[gray]{0.95}\textbf{34.26\tiny$\pm$4.39} & \cellcolor[gray]{0.95}\textbf{13.23\tiny$\pm$2.42} & 22.28\tiny$\pm$0.14 & 36.31\tiny$\pm$5.85 & 13.72\tiny$\pm$3.69 & 22.72\tiny$\pm$0.28 & 37.69\tiny$\pm$5.09 & 13.56\tiny$\pm$3.43 \\[3pt]
    
    5 & \cellcolor[gray]{0.85}\textbf{15.29\tiny$\pm$0.18} & \cellcolor[gray]{0.95}\textbf{43.13\tiny$\pm$5.36} & 6.45\tiny$\pm$1.57 & 16.14\tiny$\pm$0.31 & 48.24\tiny$\pm$6.35 & 6.25\tiny$\pm$1.66 & 16.65\tiny$\pm$0.23 & 51.22\tiny$\pm$2.59 & \cellcolor[gray]{0.95}\textbf{5.67\tiny$\pm$0.83} \\[3pt]
    
    6 & \cellcolor[gray]{0.85}\textbf{11.56\tiny$\pm$0.06} & \cellcolor[gray]{0.85}\textbf{56.58\tiny$\pm$2.58} & 3.70\tiny$\pm$0.46 & 12.21\tiny$\pm$0.10 & 66.25\tiny$\pm$2.26 & \cellcolor[gray]{0.95}\textbf{2.58\tiny$\pm$0.46} & 12.50\tiny$\pm$0.08 & 66.02\tiny$\pm$1.37 & 3.16\tiny$\pm$0.33 \\[3pt]
    
    7 & 16.24\tiny$\pm$0.02 & 53.80\tiny$\pm$0.66 & 5.29\tiny$\pm$0.18 & \cellcolor[gray]{0.85}\textbf{15.40\tiny$\pm$0.12} & 53.42\tiny$\pm$2.78 & \cellcolor[gray]{0.95}\textbf{4.93\tiny$\pm$0.72} & 15.80\tiny$\pm$0.09 & \cellcolor[gray]{0.95}\textbf{50.01\tiny$\pm$3.24} & 5.83\tiny$\pm$0.97 \\[3pt]
    
    8 & \cellcolor[gray]{0.85}\textbf{6.37\tiny$\pm$0.17} & \cellcolor[gray]{0.85}\textbf{44.73\tiny$\pm$2.77} & \cellcolor[gray]{0.95}\textbf{1.39\tiny$\pm$0.49} & 9.36\tiny$\pm$0.08 & 70.99\tiny$\pm$4.30 & 1.75\tiny$\pm$0.57 & 9.85\tiny$\pm$0.09 & 71.25\tiny$\pm$4.31 & 1.87\tiny$\pm$0.64 \\[3pt]
    
    9 & \cellcolor[gray]{0.85}\textbf{16.53\tiny$\pm$0.29} & \cellcolor[gray]{0.95}\textbf{33.18\tiny$\pm$6.06} & 9.55\tiny$\pm$2.69 & 18.17\tiny$\pm$0.26 & 40.32\tiny$\pm$6.49 & \cellcolor[gray]{0.95}\textbf{8.92\tiny$\pm$2.36} & 18.68\tiny$\pm$0.40 & 40.14\tiny$\pm$4.29 & 9.68\tiny$\pm$2.23 \\[3pt]
    
    10 & \cellcolor[gray]{0.85}\textbf{17.67\tiny$\pm$0.21} & \cellcolor[gray]{0.95}\textbf{32.40\tiny$\pm$3.60} & 10.20\tiny$\pm$2.02 & 19.34\tiny$\pm$0.77 & 38.86\tiny$\pm$7.62 & \cellcolor[gray]{0.95}\textbf{9.52\tiny$\pm$2.76} & 19.79\tiny$\pm$0.20 & 39.57\tiny$\pm$2.56 & 9.76\tiny$\pm$1.19 \\[3pt]
    
    11 & \cellcolor[gray]{0.95}\textbf{8.03\tiny$\pm$0.05} & 48.81\tiny$\pm$1.84 & \cellcolor[gray]{0.95}\textbf{2.47\tiny$\pm$0.22} & 8.16\tiny$\pm$0.28 & \cellcolor[gray]{0.95}\textbf{40.21\tiny$\pm$4.02} & 3.84\tiny$\pm$0.84 & 8.81\tiny$\pm$0.33 & 51.87\tiny$\pm$8.92 & 2.93\tiny$\pm$1.57 \\[3pt]
    
    12 & \cellcolor[gray]{0.95}\textbf{15.26\tiny$\pm$0.09} & 52.94\tiny$\pm$2.22 & \cellcolor[gray]{0.95}\textbf{5.16\tiny$\pm$0.51} & 15.27\tiny$\pm$0.35 & \cellcolor[gray]{0.95}\textbf{46.76\tiny$\pm$6.28} & 6.94\tiny$\pm$2.00 & 15.42\tiny$\pm$0.28 & 51.66\tiny$\pm$3.16 & 5.70\tiny$\pm$0.83 \\[3pt]
    
    13 & \cellcolor[gray]{0.85}\textbf{21.50\tiny$\pm$0.13} & \cellcolor[gray]{0.95}\textbf{27.77\tiny$\pm$2.30} & 17.21\tiny$\pm$1.54 & 22.29\tiny$\pm$0.40 & 29.45\tiny$\pm$6.14 & 17.28\tiny$\pm$4.05 & 22.53\tiny$\pm$0.17 & 32.46\tiny$\pm$4.03 & \cellcolor[gray]{0.95}\textbf{15.75\tiny$\pm$2.49} \\[3pt]
    
    14 & \cellcolor[gray]{0.85}\textbf{22.33\tiny$\pm$0.14} & \cellcolor[gray]{0.95}\textbf{27.85\tiny$\pm$2.13} & \cellcolor[gray]{0.95}\textbf{18.10\tiny$\pm$1.76} & 23.14\tiny$\pm$0.24 & 28.89\tiny$\pm$4.01 & 18.69\tiny$\pm$2.96 & 23.56\tiny$\pm$0.22 & 28.67\tiny$\pm$4.28 & 19.66\tiny$\pm$3.58 \\[3pt]
    
    15 & 21.31\tiny$\pm$0.15 & 25.98\tiny$\pm$1.42 & \cellcolor[gray]{0.95}\textbf{17.42\tiny$\pm$0.93} & \cellcolor[gray]{0.95}\textbf{21.27\tiny$\pm$0.18} & \cellcolor[gray]{0.95}\textbf{25.10\tiny$\pm$1.49} & 18.20\tiny$\pm$1.33 & 21.94\tiny$\pm$0.17 & 27.02\tiny$\pm$2.63 & 17.71\tiny$\pm$2.33 \\[3pt]
    
    16 & 6.49\tiny$\pm$0.03 & 80.77\tiny$\pm$1.32 & 0.81\tiny$\pm$0.13 & \cellcolor[gray]{0.85}\textbf{6.14\tiny$\pm$0.02} & \cellcolor[gray]{0.95}\textbf{79.44\tiny$\pm$1.44} & \cellcolor[gray]{0.95}\textbf{0.72\tiny$\pm$0.12} & 6.39\tiny$\pm$0.04 & 79.79\tiny$\pm$3.52 & 0.78\tiny$\pm$0.26 \\[3pt]
    
    17 & \cellcolor[gray]{0.85}\textbf{14.74\tiny$\pm$0.16} & \cellcolor[gray]{0.95}\textbf{46.38\tiny$\pm$5.42} & 5.71\tiny$\pm$1.42 & 15.55\tiny$\pm$0.29 & 51.52\tiny$\pm$7.32 & 5.60\tiny$\pm$1.71 & 15.87\tiny$\pm$0.21 & 54.52\tiny$\pm$1.97 & \cellcolor[gray]{0.95}\textbf{4.84\tiny$\pm$0.61} \\[3pt]
    
    18 & \cellcolor[gray]{0.95}\textbf{12.99\tiny$\pm$0.17} & \cellcolor[gray]{0.95}\textbf{55.93\tiny$\pm$2.78} & 4.24\tiny$\pm$0.75 & 13.00\tiny$\pm$0.09 & 58.60\tiny$\pm$2.93 & 3.85\tiny$\pm$0.54 & 13.39\tiny$\pm$0.13 & 64.43\tiny$\pm$3.39 & \cellcolor[gray]{0.95}\textbf{2.98\tiny$\pm$0.64} \\[3pt]
    
    19 & \cellcolor[gray]{0.85}\textbf{10.76\tiny$\pm$0.02} & 75.48\tiny$\pm$0.76 & \cellcolor[gray]{0.95}\textbf{1.47\tiny$\pm$0.10} & 10.89\tiny$\pm$0.03 & 75.88\tiny$\pm$1.86 & 1.52\tiny$\pm$0.28 & 10.97\tiny$\pm$0.04 & \cellcolor[gray]{0.95}\textbf{74.75\tiny$\pm$1.39} & 1.81\tiny$\pm$0.16 \\[3pt]
    
    20 & 5.05\tiny$\pm$0.01 & 85.03\tiny$\pm$0.72 & 0.48\tiny$\pm$0.04 & \cellcolor[gray]{0.85}\textbf{4.82\tiny$\pm$0.03} & 83.89\tiny$\pm$1.62 & \cellcolor[gray]{0.95}\textbf{0.45\tiny$\pm$0.09} & 5.07\tiny$\pm$0.08 & \cellcolor[gray]{0.95}\textbf{83.66\tiny$\pm$3.67} & 0.58\tiny$\pm$0.26 \\[3pt]
    
    21 & \cellcolor[gray]{0.85}\textbf{8.37\tiny$\pm$0.02} & \cellcolor[gray]{0.95}\textbf{77.82\tiny$\pm$0.83} & 1.07\tiny$\pm$0.10 & 8.59\tiny$\pm$0.03 & 79.92\tiny$\pm$1.63 & \cellcolor[gray]{0.95}\textbf{0.89\tiny$\pm$0.14} & 8.63\tiny$\pm$0.06 & 78.48\tiny$\pm$2.44 & 1.29\tiny$\pm$0.32 \\[3pt]
    
    22 & 13.24\tiny$\pm$0.07 & 54.17\tiny$\pm$1.79 & 3.97\tiny$\pm$0.34 & \cellcolor[gray]{0.85}\textbf{12.95\tiny$\pm$0.11} & \cellcolor[gray]{0.95}\textbf{54.14\tiny$\pm$3.30} & \cellcolor[gray]{0.95}\textbf{3.67\tiny$\pm$0.65} & 13.40\tiny$\pm$0.20 & 54.80\tiny$\pm$3.82 & 4.03\tiny$\pm$0.78 \\[3pt]
    
    23 & \cellcolor[gray]{0.95}\textbf{14.87\tiny$\pm$0.18} & 35.14\tiny$\pm$1.96 & \cellcolor[gray]{0.95}\textbf{6.92\tiny$\pm$0.53} & 14.87\tiny$\pm$0.17 & 33.10\tiny$\pm$2.15 & 7.96\tiny$\pm$0.82 & 15.05\tiny$\pm$0.22 & \cellcolor[gray]{0.95}\textbf{31.60\tiny$\pm$4.13} & 8.56\tiny$\pm$1.66 \\[3pt]
    
    24 & 14.42\tiny$\pm$0.09 & 44.08\tiny$\pm$1.21 & 5.70\tiny$\pm$0.44 & \cellcolor[gray]{0.85}\textbf{13.93\tiny$\pm$0.11} & \cellcolor[gray]{0.95}\textbf{40.38\tiny$\pm$3.02} & 6.30\tiny$\pm$0.89 & 14.34\tiny$\pm$0.22 & 45.76\tiny$\pm$4.60 & \cellcolor[gray]{0.95}\textbf{5.11\tiny$\pm$1.31} \\[3pt]
    
    25 & 10.94\tiny$\pm$0.09 & 43.20\tiny$\pm$1.32 & 4.14\tiny$\pm$0.28 & \cellcolor[gray]{0.85}\textbf{10.36\tiny$\pm$0.22} & \cellcolor[gray]{0.85}\textbf{37.42\tiny$\pm$1.51} & 4.93\tiny$\pm$0.50 & 10.92\tiny$\pm$0.30 & 44.53\tiny$\pm$5.15 & \cellcolor[gray]{0.95}\textbf{3.84\tiny$\pm$1.19} \\[3pt]
    
    \midrule
    
    \textbf{Average} & 13.45\tiny$\pm$0.12 & 49.60\tiny$\pm$2.32 & 6.30\tiny$\pm$0.81 & 13.84\tiny$\pm$0.19 & 51.07\tiny$\pm$3.87 & 6.50\tiny$\pm$1.28 & 14.18\tiny$\pm$0.18 & 52.50\tiny$\pm$3.69 & 6.33\tiny$\pm$1.16 \\[3pt]
    
    \bottomrule
    \end{tabular}
    \end{adjustbox}
    \label{table:brier_multimodals}
\end{table*}

%% file: tables/table_brier_upscaled_multimodals.tex
\begin{table*}[h!]
\caption{
    \textbf{Brier Score for Multimodal Models under Loss-Upweighting.}
    Aggregate and class-stratified Brier scores are reported for multimodal architectures trained with loss upweighting.
    Loss upweighting reduces positive-class Brier loss but increases negative-class and aggregate Brier loss, suggesting that the intervention shifts calibration error across classes rather than fully resolving the reliability gap.
    \footnotesize{(Dark-bold: $p<0.05$, Wilcoxon signed-rank test, 5 seeds; Light-bold: highest mean, not significant)}
    }
    \centering
    \scriptsize
    \resizebox{\linewidth}{!}{
    \begin{tabular}{c|ccc|ccc|cccc}
    \multirow{2}{*}[-1.5em]{\textbf{\makecell{Clinical \\ Condition}}} 

    & \multicolumn{3}{c}{\textbf{MedFuse (Loss-Upweighting)}}
    & \multicolumn{3}{c}{\textbf{DrFuse (Loss-Upweighting)}}
    & \multicolumn{3}{c}{\textbf{MeTra (Loss-Upweighting)}}\\
    \midrule
    \addlinespace[5pt]
    & \textbf{$\mathbf{BS} \downarrow$}
    & \textbf{$\mathbf{BS}_{c=1} \downarrow$}
    & \textbf{$\mathbf{BS}_{c=0} \downarrow$}
    
     & \textbf{$\mathbf{BS} \downarrow$}
    & \textbf{$\mathbf{BS}_{c=1} \downarrow$}
    & \textbf{$\mathbf{BS}_{c=0} \downarrow$}
    
     & \textbf{$\mathbf{BS} \downarrow$}
    & \textbf{$\mathbf{BS}_{c=1} \downarrow$}
    & \textbf{$\mathbf{BS}_{c=0} \downarrow$}
    \\[2pt]
    \midrule

    1 & \cellcolor[gray]{0.85}\textbf{20.22\tiny$\pm$0.47} & \cellcolor[gray]{0.95}\textbf{19.19\tiny$\pm$1.02} & \cellcolor[gray]{0.95}\textbf{20.71\tiny$\pm$1.16} & 21.30\tiny$\pm$0.55 & 20.82\tiny$\pm$1.21 & 21.52\tiny$\pm$1.34 & 21.35\tiny$\pm$0.81 & 22.66\tiny$\pm$3.41 & 20.73\tiny$\pm$2.78 \\[3pt]
    
    2 & 10.68\tiny$\pm$0.51 & 17.41\tiny$\pm$0.75 & 10.10\tiny$\pm$0.61 & \cellcolor[gray]{0.95}\textbf{10.46\tiny$\pm$1.44} & 16.47\tiny$\pm$2.92 & \cellcolor[gray]{0.95}\textbf{10.00\tiny$\pm$1.77} & 12.87\tiny$\pm$1.95 & \cellcolor[gray]{0.95}\textbf{11.96\tiny$\pm$1.37} & 12.94\tiny$\pm$2.23 \\[3pt]
    
    3 & 20.79\tiny$\pm$1.45 & 23.15\tiny$\pm$2.02 & 20.55\tiny$\pm$1.80 & 24.24\tiny$\pm$4.12 & \cellcolor[gray]{0.95}\textbf{21.09\tiny$\pm$5.20} & 24.56\tiny$\pm$5.03 & \cellcolor[gray]{0.95}\textbf{20.63\tiny$\pm$4.35} & 25.50\tiny$\pm$8.98 & \cellcolor[gray]{0.95}\textbf{20.14\tiny$\pm$5.71} \\[3pt]
    
    4 & \cellcolor[gray]{0.95}\textbf{22.49\tiny$\pm$0.73} & \cellcolor[gray]{0.95}\textbf{21.98\tiny$\pm$2.67} & 22.81\tiny$\pm$2.78 & 24.15\tiny$\pm$0.56 & 22.95\tiny$\pm$1.58 & 24.88\tiny$\pm$1.85 & 23.48\tiny$\pm$0.51 & 25.52\tiny$\pm$3.00 & \cellcolor[gray]{0.95}\textbf{22.23\tiny$\pm$2.64} \\[3pt]
    
    5 & \cellcolor[gray]{0.95}\textbf{20.05\tiny$\pm$1.22} & \cellcolor[gray]{0.95}\textbf{20.14\tiny$\pm$2.46} & \cellcolor[gray]{0.95}\textbf{20.02\tiny$\pm$2.35} & 22.08\tiny$\pm$1.49 & 20.60\tiny$\pm$2.84 & 22.54\tiny$\pm$2.81 & 21.79\tiny$\pm$2.29 & 23.77\tiny$\pm$6.22 & 21.17\tiny$\pm$4.94 \\[3pt]
    
    6 & \cellcolor[gray]{0.95}\textbf{18.68\tiny$\pm$3.02} & 25.18\tiny$\pm$5.83 & \cellcolor[gray]{0.95}\textbf{17.55\tiny$\pm$4.56} & 23.65\tiny$\pm$1.18 & \cellcolor[gray]{0.95}\textbf{23.04\tiny$\pm$1.62} & 23.76\tiny$\pm$1.67 & 21.94\tiny$\pm$0.97 & 27.74\tiny$\pm$1.44 & 20.92\tiny$\pm$1.39 \\[3pt]
    
    7 & 23.33\tiny$\pm$0.38 & 21.81\tiny$\pm$0.78 & 23.78\tiny$\pm$0.71 & 22.71\tiny$\pm$0.78 & \cellcolor[gray]{0.95}\textbf{21.33\tiny$\pm$1.23} & 23.09\tiny$\pm$1.33 & \cellcolor[gray]{0.95}\textbf{21.41\tiny$\pm$1.24} & 23.51\tiny$\pm$3.18 & \cellcolor[gray]{0.95}\textbf{20.80\tiny$\pm$2.50} \\[3pt]
    
    8 & \cellcolor[gray]{0.85}\textbf{13.27\tiny$\pm$2.93} & \cellcolor[gray]{0.95}\textbf{22.82\tiny$\pm$3.49} & \cellcolor[gray]{0.85}\textbf{12.03\tiny$\pm$3.76} & 22.72\tiny$\pm$2.70 & 24.31\tiny$\pm$3.81 & 22.53\tiny$\pm$3.49 & 20.68\tiny$\pm$3.72 & 27.08\tiny$\pm$6.30 & 19.85\tiny$\pm$5.01 \\[3pt]
    
    9 & \cellcolor[gray]{0.95}\textbf{19.19\tiny$\pm$1.74} & 19.15\tiny$\pm$3.82 & \cellcolor[gray]{0.95}\textbf{19.21\tiny$\pm$4.04} & 22.59\tiny$\pm$1.14 & \cellcolor[gray]{0.95}\textbf{19.14\tiny$\pm$2.39} & 24.03\tiny$\pm$2.59 & 21.79\tiny$\pm$1.54 & 24.53\tiny$\pm$4.91 & 20.64\tiny$\pm$4.17 \\[3pt]
    
    10 & \cellcolor[gray]{0.95}\textbf{19.96\tiny$\pm$1.16} & 19.70\tiny$\pm$3.53 & 20.09\tiny$\pm$3.49 & 22.19\tiny$\pm$1.43 & \cellcolor[gray]{0.95}\textbf{19.26\tiny$\pm$3.37} & 23.66\tiny$\pm$3.80 & 21.59\tiny$\pm$0.98 & 25.94\tiny$\pm$6.96 & \cellcolor[gray]{0.95}\textbf{19.38\tiny$\pm$4.93} \\[3pt]
    
    11 & \cellcolor[gray]{0.95}\textbf{14.69\tiny$\pm$0.96} & 16.26\tiny$\pm$0.92 & \cellcolor[gray]{0.95}\textbf{14.48\tiny$\pm$1.20} & 16.54\tiny$\pm$2.52 & \cellcolor[gray]{0.95}\textbf{13.18\tiny$\pm$3.00} & 17.00\tiny$\pm$3.27 & 18.50\tiny$\pm$4.04 & 16.60\tiny$\pm$5.45 & 18.76\tiny$\pm$5.28 \\[3pt]
    
    12 & \cellcolor[gray]{0.95}\textbf{21.60\tiny$\pm$0.45} & 21.43\tiny$\pm$1.30 & \cellcolor[gray]{0.95}\textbf{21.65\tiny$\pm$0.87} & 22.71\tiny$\pm$2.39 & \cellcolor[gray]{0.95}\textbf{19.03\tiny$\pm$3.36} & 23.69\tiny$\pm$3.90 & 23.51\tiny$\pm$3.10 & 20.98\tiny$\pm$5.14 & 24.19\tiny$\pm$5.27 \\[3pt]
    
    13 & \cellcolor[gray]{0.95}\textbf{22.50\tiny$\pm$0.44} & 22.32\tiny$\pm$3.24 & \cellcolor[gray]{0.95}\textbf{22.63\tiny$\pm$2.94} & 23.21\tiny$\pm$0.93 & \cellcolor[gray]{0.95}\textbf{21.06\tiny$\pm$3.53} & 24.72\tiny$\pm$4.00 & 23.34\tiny$\pm$0.78 & 24.26\tiny$\pm$5.28 & 22.71\tiny$\pm$4.80 \\[3pt]
    
    14 & \cellcolor[gray]{0.85}\textbf{22.91\tiny$\pm$0.42} & 24.05\tiny$\pm$3.45 & \cellcolor[gray]{0.95}\textbf{22.04\tiny$\pm$3.30} & 23.92\tiny$\pm$0.47 & \cellcolor[gray]{0.95}\textbf{21.48\tiny$\pm$1.75} & 25.81\tiny$\pm$2.17 & 24.09\tiny$\pm$0.46 & 22.84\tiny$\pm$2.49 & 25.05\tiny$\pm$2.69 \\[3pt]
    
    15 & 21.58\tiny$\pm$0.10 & 21.95\tiny$\pm$1.02 & 21.27\tiny$\pm$0.96 & \cellcolor[gray]{0.95}\textbf{21.36\tiny$\pm$0.25} & \cellcolor[gray]{0.95}\textbf{21.58\tiny$\pm$1.93} & \cellcolor[gray]{0.95}\textbf{21.18\tiny$\pm$1.82} & 22.09\tiny$\pm$0.24 & 22.54\tiny$\pm$1.02 & 21.71\tiny$\pm$1.01 \\[3pt]
    
    16 & \cellcolor[gray]{0.95}\textbf{21.06\tiny$\pm$1.59} & 27.14\tiny$\pm$2.09 & \cellcolor[gray]{0.95}\textbf{20.59\tiny$\pm$1.87} & 21.18\tiny$\pm$2.09 & \cellcolor[gray]{0.95}\textbf{21.66\tiny$\pm$2.93} & 21.14\tiny$\pm$2.46 & 23.25\tiny$\pm$2.74 & 22.94\tiny$\pm$3.54 & 23.28\tiny$\pm$3.21 \\[3pt]
    
    17 & \cellcolor[gray]{0.95}\textbf{20.22\tiny$\pm$1.38} & \cellcolor[gray]{0.95}\textbf{20.99\tiny$\pm$2.71} & \cellcolor[gray]{0.95}\textbf{20.00\tiny$\pm$2.53} & 22.39\tiny$\pm$1.61 & 21.33\tiny$\pm$3.03 & 22.68\tiny$\pm$2.88 & 21.99\tiny$\pm$2.30 & 23.99\tiny$\pm$5.58 & 21.41\tiny$\pm$4.52 \\[3pt]
    
    18 & \cellcolor[gray]{0.95}\textbf{21.62\tiny$\pm$2.15} & 22.05\tiny$\pm$2.89 & \cellcolor[gray]{0.95}\textbf{21.53\tiny$\pm$3.18} & 22.31\tiny$\pm$1.42 & 23.29\tiny$\pm$2.37 & 22.12\tiny$\pm$2.16 & 23.48\tiny$\pm$1.04 & \cellcolor[gray]{0.95}\textbf{21.65\tiny$\pm$1.95} & 23.86\tiny$\pm$1.64 \\[3pt]
    
    19 & \cellcolor[gray]{0.95}\textbf{23.46\tiny$\pm$0.72} & 25.00\tiny$\pm$0.88 & \cellcolor[gray]{0.95}\textbf{23.24\tiny$\pm$0.95} & 23.69\tiny$\pm$0.84 & \cellcolor[gray]{0.95}\textbf{24.90\tiny$\pm$1.11} & 23.52\tiny$\pm$1.12 & 24.68\tiny$\pm$2.72 & 25.95\tiny$\pm$4.21 & 24.50\tiny$\pm$3.71 \\[3pt]
    
    20 & 17.87\tiny$\pm$0.90 & \cellcolor[gray]{0.95}\textbf{28.31\tiny$\pm$1.16} & 17.27\tiny$\pm$1.00 & \cellcolor[gray]{0.95}\textbf{16.35\tiny$\pm$1.54} & 31.72\tiny$\pm$1.72 & \cellcolor[gray]{0.95}\textbf{15.50\tiny$\pm$1.72} & 18.25\tiny$\pm$2.74 & 31.37\tiny$\pm$4.72 & 17.50\tiny$\pm$3.11 \\[3pt]
    
    21 & 23.58\tiny$\pm$2.10 & \cellcolor[gray]{0.95}\textbf{20.74\tiny$\pm$2.31} & 23.87\tiny$\pm$2.56 & 22.06\tiny$\pm$1.62 & 26.16\tiny$\pm$2.61 & 21.61\tiny$\pm$2.05 & \cellcolor[gray]{0.95}\textbf{20.65\tiny$\pm$1.95} & 30.12\tiny$\pm$2.48 & \cellcolor[gray]{0.95}\textbf{19.66\tiny$\pm$2.40} \\[3pt]
    
    22 & 20.27\tiny$\pm$0.67 & \cellcolor[gray]{0.85}\textbf{18.95\tiny$\pm$1.18} & 20.57\tiny$\pm$1.08 & \cellcolor[gray]{0.95}\textbf{19.66\tiny$\pm$1.06} & 21.47\tiny$\pm$1.84 & 19.26\tiny$\pm$1.71 & 19.73\tiny$\pm$0.57 & 23.25\tiny$\pm$1.21 & \cellcolor[gray]{0.95}\textbf{18.93\tiny$\pm$0.97} \\[3pt]
    
    23 & \cellcolor[gray]{0.95}\textbf{16.97\tiny$\pm$0.61} & 18.47\tiny$\pm$0.78 & 16.38\tiny$\pm$1.14 & 17.26\tiny$\pm$0.90 & \cellcolor[gray]{0.95}\textbf{18.08\tiny$\pm$2.21} & 16.94\tiny$\pm$2.07 & 17.22\tiny$\pm$0.56 & 19.39\tiny$\pm$2.84 & \cellcolor[gray]{0.95}\textbf{16.37\tiny$\pm$1.83} \\[3pt]
    
    24 & 19.49\tiny$\pm$0.74 & 19.72\tiny$\pm$1.00 & 19.42\tiny$\pm$1.24 & 18.83\tiny$\pm$0.45 & \cellcolor[gray]{0.95}\textbf{18.54\tiny$\pm$0.80} & 18.91\tiny$\pm$0.79 & \cellcolor[gray]{0.95}\textbf{18.21\tiny$\pm$0.56} & 21.43\tiny$\pm$1.53 & \cellcolor[gray]{0.95}\textbf{17.26\tiny$\pm$1.12} \\[3pt]
    
    25 & 16.82\tiny$\pm$0.53 & 18.23\tiny$\pm$0.97 & 16.53\tiny$\pm$0.81 & 16.31\tiny$\pm$0.98 & \cellcolor[gray]{0.95}\textbf{16.05\tiny$\pm$1.63} & 16.36\tiny$\pm$1.51 & \cellcolor[gray]{0.95}\textbf{15.45\tiny$\pm$1.38} & 20.60\tiny$\pm$2.97 & \cellcolor[gray]{0.95}\textbf{14.36\tiny$\pm$2.30} \\[3pt]
    
    \midrule
    
    \textbf{Average} & 19.73\tiny$\pm$1.09 & 21.45\tiny$\pm$2.09 & 19.53\tiny$\pm$2.04 & 20.96\tiny$\pm$1.38 & 21.14\tiny$\pm$2.40 & 21.24\tiny$\pm$2.37 & 20.88\tiny$\pm$1.74 & 23.45\tiny$\pm$3.85 & 20.33\tiny$\pm$3.21 \\[3pt]

    \bottomrule
    \end{tabular}
    }
    \label{table:brier_multimodals_upweighted}
\end{table*}